\documentclass{article}

\PassOptionsToPackage{numbers, compress}{natbib}

\usepackage[main,final]{neurips_2026}

\usepackage[utf8]{inputenc}
\usepackage[T1]{fontenc}
\usepackage{hyperref}
\usepackage{url}
\usepackage{microtype}
\usepackage{graphicx}
\usepackage{subcaption}
\usepackage{booktabs}
\usepackage{rotating}
\usepackage{amsmath}
\usepackage{amsfonts}
\usepackage{amssymb}
\usepackage{mathtools}
\usepackage{amsthm}
\usepackage[capitalize,noabbrev]{cleveref}
\usepackage{xspace}
\usepackage{bm}
\usepackage{enumitem}
\usepackage{changepage}
\usepackage{multibib}
\usepackage{multirow}
\usepackage{makecell}
\usepackage{nicefrac}
\usepackage{xcolor}
\usepackage{algorithm}
\usepackage{algorithmic}
\usepackage{prompt_style}
\usepackage[disable,textsize=tiny]{todonotes}
\usepackage{wrapfig}
\usepackage{needspace}
\usepackage{marvosym}

\newcites{app}{Appendix References}

\theoremstyle{plain}
\newtheorem{theorem}{Theorem}[section]

\newtheorem{lemma}[theorem]{Lemma}
\newtheorem{corollary}[theorem]{Corollary}
\theoremstyle{definition}

\newtheorem{assumption}[theorem]{Assumption}
\theoremstyle{remark}

\newcommand{\sys}{LB-MCTS\xspace}

\title{Tree-Structured Synergy of Large Language Models and Bayesian Optimization for Efficient CASH}

\author{%
  Beicheng Xu$^{\dagger}$ \quad Weitong Qian$^{\dagger}$ \quad Lingching Tung$^{\dagger}$ \quad Yupeng Lu$^{\dagger}$ \quad Bin Cui$^{\dagger}$\\
  $^{\dagger}$School of CS \& Key Lab of High Confidence Software Technologies (MOE) \&\\
  Beijing Key Laboratory of Software and Hardware Cooperative Artificial Intelligence Systems,\\
  Peking University, Beijing, China.\\
  \texttt{\{beichengxu,weitong.qian,lingchingtung\}@stu.pku.edu.cn}\\
  \texttt{\{xinkelyp,cui.bin\}@pku.edu.cn}
}

\begin{document}

\begingroup
\renewcommand{\fnsymbol}[1]{\text{\Letter}}
\maketitle
\endgroup

\begin{abstract}
To lower the expertise barrier in machine learning, the AutoML community has focused on the CASH problem, which jointly automates algorithm selection and hyperparameter tuning.
While traditional methods like Bayesian Optimization (BO) struggle with cold-start issues, Large Language Models (LLMs) can mitigate these through semantic priors.
However, existing LLM-based optimizers generalize poorly to high-dimensional, structured CASH spaces. 
CASH requires coordinated exploration and exploitation both across candidate algorithms and within their algorithm-specific hyperparameter spaces. 
Yet existing approaches often rely on flat optimization histories that mix observations across algorithms and obscure configuration-refinement trajectories.
In this paper, we propose \sys, a trajectory-structured optimization framework that uses a Monte Carlo Tree Search tree as a shared state for algorithm selection, hyperparameter refinement, and BO--LLM proposer synergy.
Within this shared state, BO provides algorithm-specific surrogate modeling for quantitative search, while the LLM exploits path-aware selective memory to generate semantic proposals and reflections.
As the surrogate model improves, a reliability-aware proposer policy adaptively shifts from LLM-driven to BO-driven proposals within a unified search trajectory.
Experiments on 104 AMLB datasets demonstrate that \sys consistently outperforms BO-based, LLM-based, and hybrid baselines.
Code: \href{https://github.com/PKU-DAIR/open-box}{PKU-DAIR/open-box}.
\end{abstract}

\section{Introduction}
\label{sec:introduction}

In recent years, machine learning has advanced in diverse domains, such as computer vision~\cite{he2016deep,lecun2015deep}, and recommendation systems~\cite{sun2019bert4rec}.
Despite these advancements, developing tailored solutions with strong performance remains a knowledge-intensive task, requiring careful selection of suitable ML algorithms~\cite{zoppi2025strategy} and hyperparameter tuning~\cite{Ren2026}.
To lower barriers and streamline deployment, the AutoML community has introduced the Combined Algorithm Selection and Hyperparameter Optimization (CASH) problem~\cite{autoweka_thornton2013} and proposed several methods~\cite {he2021automl,hutter2019automated,xu2026pseo} to automate this optimization process.
Among them, the dominant approach is Bayesian Optimization (BO), which builds surrogate models of the performance landscape to guide iterative evaluations.
However, by treating configurations as abstract numeric vectors, pure BO fails to incorporate prior knowledge, necessitating extensive evaluations to explore high-dimensional and conditional search spaces~\cite{bohb_falkner2018bohb}.

Recently, the reasoning and in-context learning (ICL) capabilities of large language models~\cite{song2024communication} have made them attractive for black-box optimization~\cite{kroeger2023context,opro_yang2023large}. 
LLMs can serve as context-aware ``experts'' with domain knowledge~\cite{sun2025multimodal,zhou2024db}, alleviating the limitations of BO.
Early works treat the LLM as a standalone optimizer, iteratively prompting it to obtain candidates~\cite{opro_yang2023large}. 
However, pure LLM optimizer suffers from inherent limitations and is unreliable in many practical scenarios~\cite{huang2024exploring}. Because it does not fit a surrogate model (nor quantify uncertainty) and struggles to utilize the full optimization history with a limited context window. 
Therefore, attempts have been made to integrate LLMs with BO~\cite{bopro_agarwal2025searching,bora_cisse2025language}, e.g., using LLMs to generate warm-starts or candidates for BO, or interleaving them with BO for alternating optimization. Such synergy has yielded improved performance.

However, existing LLM-based optimizers remain confined to unstructured and low-dimensional problems, leaving the highly structured and high-dimensional CASH problem unexplored. Tackling CASH with LLMs presents three critical challenges that current approaches fail to address:
\textbf{C1}: \textbf{Efficient context utilization}.
CASH features a hierarchical search space (i.e., specific hyperparameters active only for specific algorithms) where the history of one algorithm (e.g., SVM) provides negligible guidance for another (e.g., XGBoost). Yet, existing methods indiscriminately populate prompts with the best, recent, or entire optimization history. This approach introduces cross-algorithm noise, leading to low context utilization and high token costs. 
\textbf{C2}: \textbf{Exploration-exploitation trade-off}. 
Success in CASH demands a dual-level exploration-exploitation balance: selecting between distinct algorithms, and navigating between promising vs. unexplored regions within each algorithm’s subspace. 
However, existing LLM-based methods lack an explicit trade-off mechanism, leading to premature convergence and local optima.
\textbf{C3}: \textbf{Effective synergy with BO}.
Existing hybrids often neglect BO's structured optimization trajectories, relying instead on a mere exchange of unstructured history.
Furthermore, they either use static integration strategies without dynamic adaptation or rely on rigid thresholds to switch between LLM and BO, which generalizes poorly.
These challenges share a common root: existing methods lack a structured state that models hierarchical CASH optimization trajectories and coherently integrates BO- and LLM-produced trajectories.
Consequently, LLMs receive flat and noisy histories, and BO--LLM interaction reduces to heuristic alternation.
To address this limitation, we propose \sys, a trajectory-structured optimization framework for CASH.
The key idea is to use an MCTS tree as a structured search state, where each path records a hierarchical optimization trajectory and supports BO--LLM proposer synergy through shared trajectory updates.
The contributions are summarized as follows: 
(i) We formulate CASH as tree-structured search over optimization trajectories, enabling root-level algorithm selection and subtree-level exploration-exploitation within each algorithm's HPO space (\textbf{C2}).
(ii) We realize BO--LLM proposer synergy by unifying BO- and LLM-produced optimization trajectories in the shared tree state and dynamically selecting the proposer for configuration proposal, achieving more sample-efficient optimization than either component alone (\textbf{C3}).
(iii) We introduce Selective Tuning Memory (STM), which retrieves algorithm-specific and trajectory-relevant trials from the shared tree state, allowing the LLM to reason from focused, low-noise context (\textbf{C1}).
(iv) Experiments on 104 AMLB datasets show that \sys achieves the best average validation rank of 2.16, outperforming the strongest baseline RB (3.49), as well as BO-based, LLM-based, and hybrid baselines.

\section{Preliminary and Related Work}
\label{sec:preliminary_and_related_work}
\noindent \textbf{CASH Problem and Traditional Methods.} 
Let $\mathcal{A} = \{A^1, \dots, A^K\}$ be a set of $K$ candidate algorithms, where each algorithm $A^i$ is associated with a specific hyperparameter space $\Lambda^i$. 
Given a dataset partitioned into training and validation sets ($\mathcal{D}_{train}, \mathcal{D}_{val}$), CASH aims to jointly identify the optimal algorithm $A^*$ and its configuration $\lambda^*$ to maximize the validation metric $\mathcal{F}$:
\begin{equation}
    (A^*,\ \boldsymbol{\lambda}^*) = \operatorname*{argmax}_{A^i \in \mathcal{A}, \boldsymbol{\lambda} \in \Lambda^i} \mathcal{F}(A^i(\boldsymbol{\lambda}, \mathcal{D}_{train}), \mathcal{D}_{val}).
\end{equation}
The CASH problem was first introduced by Auto-WEKA~\cite{autoweka_thornton2013} and solved by Bayesian Optimization (BO)~\cite{gp_snoek2012practical,smac_hutter2011sequential,tpe_bergstra2011algorithms,MFTune_10.14778/3836663.3836715}, 
which iteratively fits a surrogate on evaluated configurations and maximizes an acquisition function to select the next trial.
Many leading AutoML systems like Auto-sklearn~\cite{auto-sklearn_feurer2022auto}, LightAutoML~\cite{lightautoml_vakhrushev2021lightautoml}, and H2O~\cite{h2o_ledell2020h2o} also adopt BO as the core solver for CASH.
Another line of work improves BO by diversifying evaluated models~\cite{divbo_shen2022divbo,optdivbo_poduval2024cash}.
In addition, genetic programming~\cite{tpot_olson2016tpot} and the multi-armed bandits (MAB) based methods~\cite{rb_li2020efficient, admm_liu2020admm} are also used to address the CASH problem. 
\noindent \textbf{LLM-based Optimizer.} 
Recently, researchers have explored LLMs as standalone alternatives to traditional optimizers with their emergent reasoning capabilities. These approaches have been applied to diverse tasks, including chemistry~\cite{ChemCrow_m2024augmenting}, hyperparameter optimization~\cite{llambo_liularge}, feature engineering~\cite{10.1145/3770855.3817982,CoFEH_10.1145/3770855.3817664}, and molecular optimization~\cite{lico_nguyenlico}.
Typical methods like LLAMBO~\cite{llambo_liularge} and OPRO~\cite{opro_yang2023large} use ICL to iteratively propose solutions.
However, these methods are mainly validated on low-dimensional black-box tasks and often require frequent LLM calls, limiting scalability and reliability.
Another recent line integrates BO principles into pre-trained foundation models.
FIBO~\cite{de2025simplifying} directly samples candidates from the posterior over the optimum, while ToSFiT~\cite{menet2025thompson} fine-tunes LLM with a Gaussian Process rewarding model.
These methods remain largely generative-model-driven, approximating BO-style candidate generation rather than preserving an explicit numerical BO module, and thus still inherit limitations in numerical fidelity and uncertainty quantification.

In response to the limitations of pure LLM optimizers, hybrid approaches have emerged that combine LLMs and BO~\cite{Bo_Chemian_rankovic2023bochemian,llamea_llm_van2024loop}.
BOPRO~\cite{bopro_agarwal2025searching} optimizes prompt instructions with BO, whereas ADO-LLM~\cite{ado-llm_yin2024ado} uses the LLM to generate BO candidates. 
SLLMBO~\cite{sllmbo_mahammadli2024sequential} alternates between LLM-based exploitation and TPE~\cite{tpe_bergstra2011algorithms} in a fixed 50:50 ratio.
BORA~\cite{bora_cisse2025language} switches between LLM-based search and GP-based BO using a heuristic variance-threshold policy. 
Although these hybrid approaches improve upon pure LLM optimizers, they are still primarily restricted to low-dimensional settings~($D<20$) with flat, unstructured parameter spaces.
Naively extending them to the highly structured, high-dimensional CASH setting is ineffective due to the challenges \textbf{C1}--\textbf{C3} discussed in \cref{sec:introduction}.
Overall, effectively synergizing LLMs and BO for the CASH problem remains an open challenge.

\noindent \textbf{Monte-Carlo Tree Search (MCTS).} 
MCTS builds search trees for structured spaces by balancing exploration (trying new actions) and exploitation (refining promising actions)~\cite{kocsis2006bandit,chaslot2008monte,MoGo_gelly2011monte}.
Each MCTS iteration consists of four phases:
1) \textbf{Selection:} Starting from the root, it traverses the tree to select the node for expansion. 
Modern variants (e.g., AlphaGo Zero~\cite{alphago_zero_silver2017mastering}) employ PUCT~\cite{puct_rosin2011multi} for selection. 
For a parent node $s$, the next child node $s_i$ is selected from the set of children $\mathcal{C}(s)$ to maximize:
\begin{equation}
    \label{eq:puct}
    \arg\max_{s_i \in \mathcal{C}(s)}\left\{ Q(s_i) + c_{puct} \cdot P(s_i) \cdot \frac{\sqrt{N_s}}{1 + N_{s_i}} \right\},
\end{equation}
where $Q(s_i)$ is the mean value of $s_i$, $P(s_i)$ is the prior probability of visiting $s_i$, and $N_s$($N_{s_i}$) denotes the visit count of the parent(child). The constant $c_{puct}$ balances exploitation and exploration.
2) \textbf{Expansion:} Add a new child node with an untried action. 
3) \textbf{Playout:} Conduct a simulation to yield a reward value.
4) \textbf{Backpropagation:} Propagate the reward back to update ancestral node statistics.
Motivated by this, AlphaD3M~\cite{AlphaD3M_drori2021alphad3m} and MOSAIC~\cite{mosaic_rakotoarison2019automated} build upon MCTS to explore the pipeline search space.
Recent work also combines LLMs with MCTS in sequential decision-making tasks, such as long-horizon robot task planning~\cite{zhao2023large}, text-based games~\cite{shi2025monte}, and code generation~\cite{ml-master_liu2025ml}.
In contrast, \sys uses an MCTS tree as a shared CASH state, integrating algorithm selection, LLM-generated semantic proposals, and BO-based optimization into unified optimization trajectories.

\section{Method}

In this section, we present \sys, a trajectory-structured optimization framework that uses the \textbf{MCTS} search tree to model hierarchical CASH trajectories and synergize \textbf{L}LM and \textbf{B}O proposers.
We first formulate the tree-structured search process, then describe algorithm selection, BO/LLM-based configuration proposal, and the reliability-aware proposer selection policy.

\subsection{Tree Structured Search for CASH}
\label{sec:tree_structured_search_for_cash}
\begin{figure}[t!]
	\centering
	\includegraphics[width=0.95\linewidth]{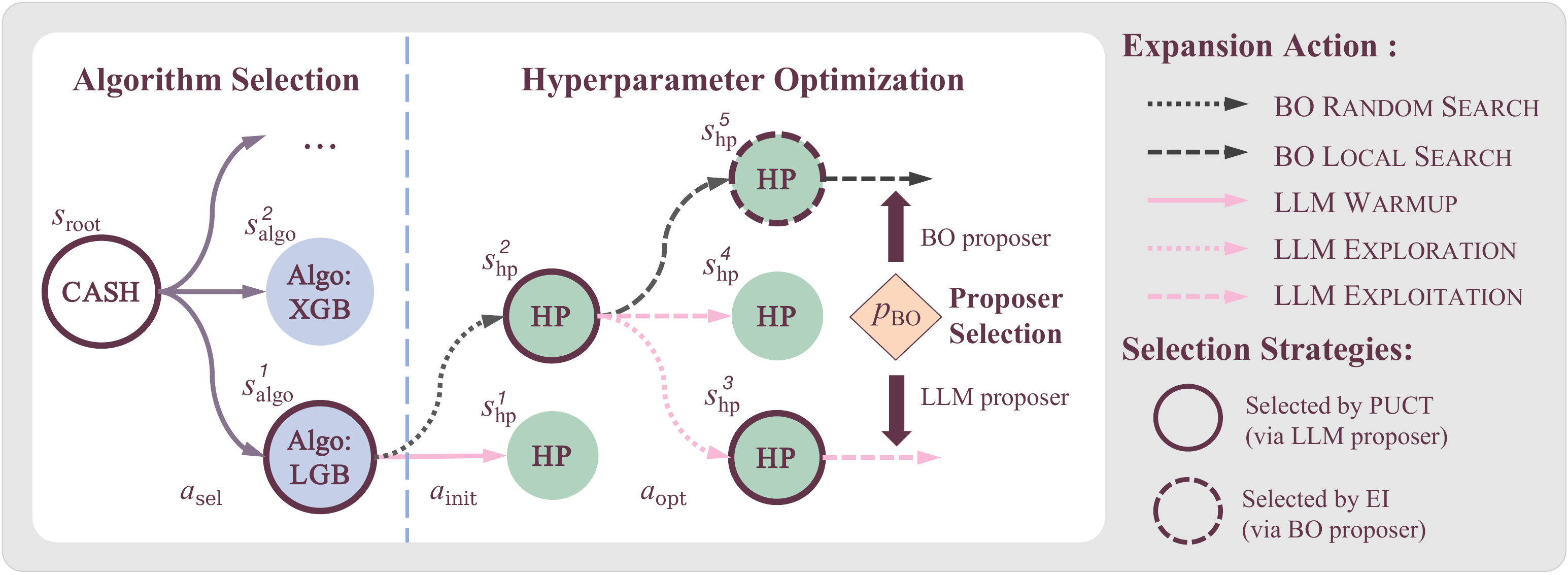}
	\caption{Tree-structured search of \sys; see Appendix~\ref{app:visualization_of_search_process} for a concrete case.}
    \label{fig:lb-mcts_structure}
\end{figure}

We model CASH problem as a sequential decision-making process. As shown in \cref{fig:lb-mcts_structure}, the search space is organized into a tree structure containing three types of nodes:
\begin{itemize}[leftmargin=1.6em, labelsep=0.5em, topsep=2pt, partopsep=0pt, itemsep=3pt, parsep=0pt]
    \item \textbf{CASH node}~(root $s_{\text{root}}$): represents the initial state of the CASH problem. It encompasses all candidate ML algorithms and their respective hyperparameter spaces.
    \item \textbf{Algo nodes}~($s_{\text{algo}}$): immediate children of the root. Each represents the hyperparameter optimization problem for a specific algorithm  $A^i$ in the subspace $\Lambda^i$.
    \item \textbf{HP nodes}~($s_{\text{hp}}$): all descendants of an Algo node. Each HP node represents a concrete configuration $\boldsymbol{\lambda}_{s_{\text{hp}}}$ of the algorithm and its validation metric $y_{s_{\text{hp}}}$.
\end{itemize}
Each node $s$ maintains three key statistics: the visit count $N_s$, the cumulative reward $R_s$, and the subtree best performance $y^{\max}_s$, i.e., the best validation metric observed so far in the subtree rooted at $s$.
After that, we define actions as transitions between nodes, yielding three action types:
\begin{itemize}[leftmargin=1.6em, labelsep=0.5em, topsep=2pt, partopsep=0pt, itemsep=3pt, parsep=0pt]
    \item \textbf{Algorithm Selection Action} ($s_{\text{root}} \xrightarrow{a_{\text{sel}}} s_{algo}$): select an algorithm $A^i \in \mathcal{A}$ to be optimized, constraining the search to its specific hyperparameter subspace.
    
    \item \textbf{Initialization Action} ($s_{\text{algo}} \xrightarrow{a_{\text{init}}} s_{\text{hp}}$): 
    generates the initial configuration $\boldsymbol{\lambda}_0$ for the selected algorithm, serving as an initialization of an optimization trajectory.

    \item \textbf{Optimization Action} ($s_{\text{hp}} \xrightarrow{a_{\text{opt}}} s_{\text{hp}}'$): 
    proposes a new configuration $\boldsymbol{\lambda}_{t+1}$ based on a basic configuration, aiming to refine the performance.
\end{itemize}

We define an \textbf{optimization trajectory} as a root-originated sequence of nodes and actions.
\textit{The trajectory structurally captures the process of choosing an algorithm, initializing a configuration, and iteratively refining its hyperparameter values.} 
CASH is thus reformulated as a \textbf{sequential decision-making process} to find the optimal trajectory that maximizes the validation performance of the final configuration.
The process can be iteratively solved by MCTS introduced in \cref{sec:preliminary_and_related_work}.
In our setting, 
Expansion represents executing an action~($a_{\text{init}}$ or $a_{\text{opt}}$) to propose a configuration via LLM/BO. 
Playout evaluates the proposed configuration to obtain its validation metric $y_{new}$.
Finally, Backpropagation propagates this value up the trajectory: for each ancestral node $s$, it computes a local relative-improvement reward $r_s = [y_{new}-y^{\max}_s]_+/|y^{\max}_s|$ before the subtree-best update, where $[x]_+=\max(x,0)$.
It then updates the visit count ($N_s \leftarrow N_s + 1$), the cumulative reward ($R_s \leftarrow R_s + r_s$), and the subtree best ($y^{\max}_s \leftarrow \max(y^{\max}_s, y_{new})$) to guide future iterations.

\textbf{Surrogate model for each algorithm.}
We maintain a separate Gaussian Process (GP)~\cite{gp_williams2006gaussian} surrogate model $f_i$ for each algorithm $A^i$. 
Importantly, $f_i$ is trained on the local dataset $\mathcal{D}_i = \{(\boldsymbol{\lambda}_{s^k_{\text{hp}}}, y_{s^k_{\text{hp}}})\}_{k=1}^{N_i}$, comprising all $N_i$ configurations and their performance collected within the subtree rooted at $s_{\text{algo}}^{i}$.
This per-algorithm modeling exploits disjoint hyperparameter spaces to reduce dimensionality, enhancing precision compared to a single high-dimensional global surrogate~\cite{rb_li2020efficient}.
We define the GP covariance function $k(\boldsymbol{\lambda}, \boldsymbol{\lambda}')$ as a mixed kernel, with a Matérn-$5/2$ kernel ($k_{\nu}$) on continuous dimensions and a Hamming kernel ($k_{cat}$) on categorical dimensions:
\begin{equation}
\label{eq:gp_kernel}
    k(\boldsymbol{\lambda}, \boldsymbol{\lambda}') = k_{\nu}(\boldsymbol{\lambda}_{cont}, \boldsymbol{\lambda}_{cont}') \times k_{cat}(\boldsymbol{\lambda}_{cat}, \boldsymbol{\lambda}_{cat}'),
\end{equation}

\subsection{Algorithm Selection}
\label{sec:algorithm_selection}
In each MCTS iteration, the Selection phase starts at $s_{\text{root}}$ and chooses one of its $K$ child Algo node, which directly corresponds to selecting an algorithm $A^i \in \mathcal{A}$ for CASH. Concretely, we utilize PUCT in \cref{eq:puct} and the estimated mean value of each Algo node $s_{\text{algo}}^{i}$ is
\begin{equation}
\label{eq:puct_qvalue}
    Q(s_{\text{algo}}^{i}) =
    \tilde{y}^{\max}_{s_{\text{algo}}^{i}}
    \left(
    1+
    {R_{s_{\text{algo}}^{i}}}/{N_{s_{\text{algo}}^{i}}}
    \right),
\end{equation}
where $\tilde{y}^{\max}_{s_{\text{algo}}^{i}}$ is the normalized best validation performance in its subtree (min–max normalized using the global
range of observed metric scores), and ${R_{s_{\text{algo}}^{i}}}/{N_{s_{\text{algo}}^{i}}}$ is the average local relative-improvement rate.
This multiplicative form treats the mean value $Q$ as an estimated performance potential, favoring algorithms with both high incumbent performance and continuous improvement tendency.

\textbf{Algorithm Prior.} 
We compute a PUCT prior score for each algorithm based on surrogate predictions. Specifically, for each algorithm $A^i$, we uniformly sample $n_s$ configurations from $\Lambda^i$, predict their performance using the corresponding surrogate $f_i$, and compute the normalized mean predicted score $\bar{v}_i$. The prior distribution is then given by a softmax over these scores:
$
    P(s_{\text{algo}}^{i}) = \frac{\exp(\bar{v}_i)}{\sum_{k=1}^{K} \exp(\bar{v}_k)}.
$

\subsection{Hyperparameter Optimization}

Once the root selects an algorithm $A^i$, the task reduces to an algorithm-specific HPO subproblem: finding $\boldsymbol{\lambda}\in\Lambda^i$ with best performance. 
During subsequent phases, 
\sys integrates two complementary configuration proposers: BO for quantitative search, and LLM for semantic proposals.

\subsubsection{BO proposer} 
\label{sec:bo_proposer}
The BO proposer directly generates a new node by selecting the best candidate from a sampled pool. 

\textbf{Candidate Generation.} We employ two sampling strategies to generate a pool of candidate configurations $\mathcal{P}$. 
(i) Random sampling draws candidates randomly from the search space $\Lambda^i$. 
(ii) Local sampling treats every HP node $s_{\text{hp}}$ in the subtree of $s_{\text{algo}}^{i}$ as a basic configuration and samples candidates from its neighborhood (i.e., via small perturbations of configuration in $s_{\text{hp}}$).

\textbf{Selection \& Expansion.} We then rank all candidates in $\mathcal{P}$ using the Expected Improvement (EI)~\cite{ei_jones1998efficient} acquisition function based on the surrogate $f_i$, and select the candidate $\boldsymbol{\lambda}_{new}$ with the highest score.
The expansion type depends on the source of $\boldsymbol{\lambda}_{new}$:
If it originates from random sampling, it is treated as an initialization action~(e.g., $s^1_{\text{algo}} \xrightarrow[\text{\tiny BO \textsc{Random Search}}]{a_{\text{init}}} s^2_{\text{hp}}$), expanding directly from the Algo node $s_{\text{algo}}^{i}$;
If it originates from local sampling, it is treated as an optimization action~(e.g., $s^2_{\text{hp}} \xrightarrow[\text{\tiny BO \textsc{Local Search}}]{a_{\text{opt}}} s^5_{\text{hp}}$), expanding as a child of the basic configuration node.

\subsubsection{LLM proposer} 
\label{sec:llm_proposer}
Following the algorithm selection, the LLM proposer continues the Selection phase using PUCT until a leaf HP node is reached.
It then expands by querying the LLM to generate a new configuration.

\noindent \textbf{Selection.} 
The LLM proposer extends the Selection by recursively applying PUCT (\cref{eq:puct} and \cref{eq:puct_qvalue}) within the algorithm subtree. The traversal proceeds until it reaches a node that LLM has not fully expanded, e.g., $s^3_{\text{hp}}$ in \cref{fig:lb-mcts_structure}. We assign a uniform prior $P(s_{\text{hp}})$ to all child nodes here. %

\noindent \textbf{Expansion via ICL.}
Given a selected leaf node $s_\text{base}$ with a basic configuration $\boldsymbol{\lambda}_{\text{base}}$, the LLM proposer executes Expansion by querying the LLM to propose a new configuration through adjusting $\boldsymbol{\lambda}_{\text{base}}$. We construct a structured prompt comprising four key components:
(i) the \emph{task description} $\psi$ (dataset description and optimization metric),
(ii) the \emph{Selective Tuning Memory} $\mathcal{M}_i(s_\text{base})$, containing informative historical trials from algorithm $A^i$,
(iii) the basic configuration $\boldsymbol{\lambda}_{\text{base}}$, and
(iv) an optimization directive $d \in \{\textsc{Warmup}, \textsc{Exploration}, \textsc{Exploitation}\}$.
Note that (ii) and (iii) are empty for an initialization expansion when $s$ is an Algo node.
Formally, the LLM generates a reasoning chain $\mathcal{R}$ followed by a new configuration $\boldsymbol{\lambda}_{\text{new}}$:
\begin{equation}
(\mathcal{R}, \boldsymbol{\lambda}_{\text{new}}) =
\mathrm{LLM}\big(\psi, \mathcal{M}_i(s_\text{base}), \boldsymbol{\lambda}_{\text{base}}, d\big).
\end{equation}
The output $\boldsymbol{\lambda}_{\text{new}}$ is added as a new child node $s^\text{new}_\text{hp}$ of $s_\text{base}$.
Crucially, the directive $d$ explicitly governs the search mode: $\textsc{Warmup}$ provides a high-quality initial point when expanding an Algo node~(e.g., $s^1_{\text{algo}} \xrightarrow[\text{\tiny LLM \textsc{Warmup}}]{\tiny a_{\text{init}}} s^1_{\text{hp}}$). 
$\textsc{Exploitation}$ instructs the LLM to locally refine $\boldsymbol{\lambda}_{\text{base}}$ for immediate gains~(e.g., $s^2_{\text{hp}} \xrightarrow[\text{\tiny LLM \textsc{Exploitation}}]{\tiny a_{\text{opt}}} s^4_{\text{hp}}$); and $\textsc{Exploration}$ encourages bold adjustments to probe unknown regions~(e.g., $s^2_{\text{hp}} \xrightarrow[\text{\tiny LLM \textsc{Exploration}}]{\tiny a_{\text{opt}}} s^3_{\text{hp}}$). 
An Algo node is regarded as a leaf before it has generated $n_{\text{w}}=3$ initial configurations via \textsc{Warmup}.
An HP node is considered fully expanded by the LLM only after both $\textsc{Exploration}$ and $\textsc{Exploitation}$ have been executed once; otherwise, it remains a leaf node.
Detailed prompt template and example are shown in Appendix~\ref{app:prompt_tuning} and~\ref{app:example}.

\begin{figure}[t]
    \centering
    \includegraphics[width=0.95\linewidth]{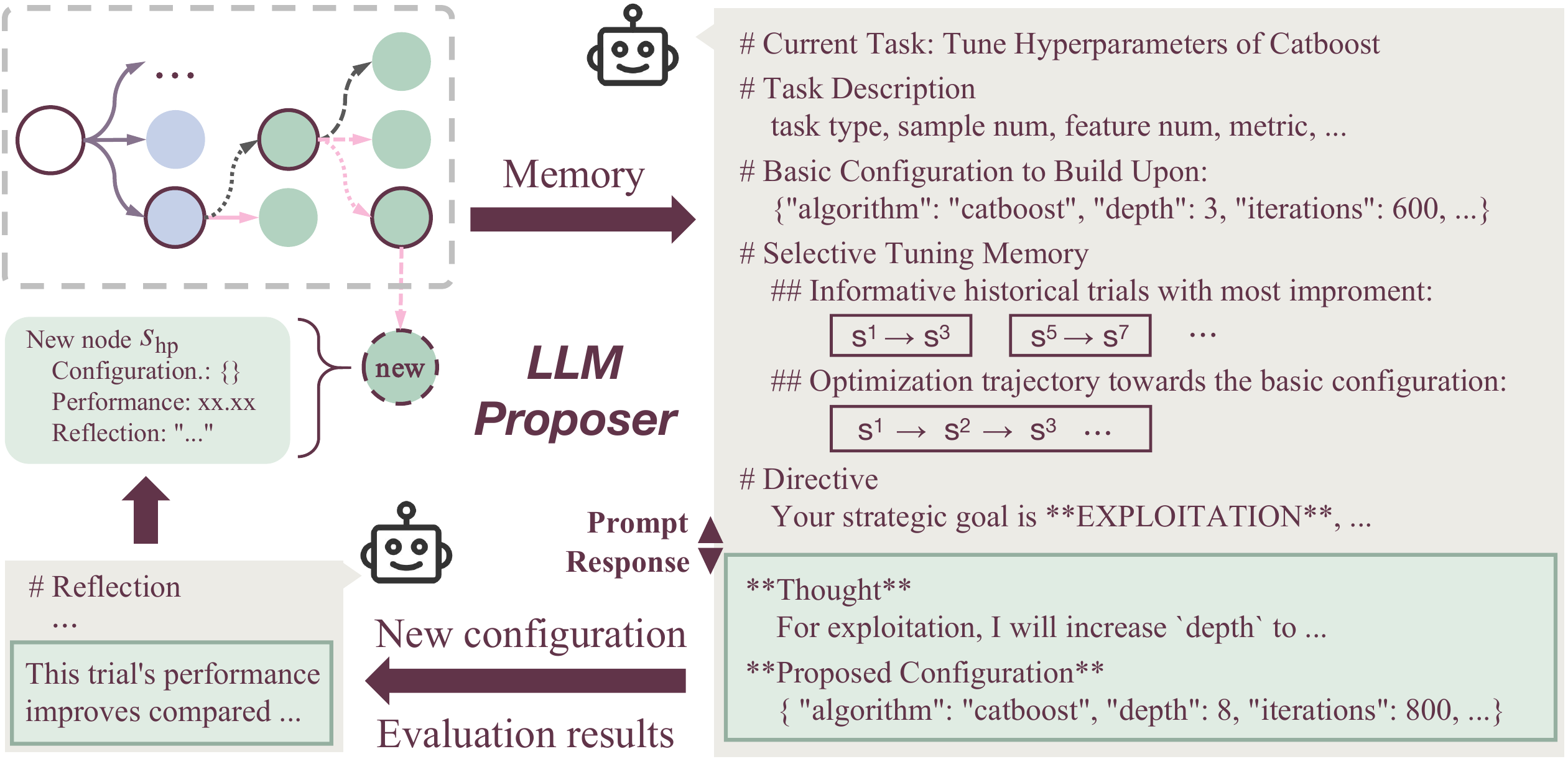}
    \caption{MCTS search of the LLM proposer.}
    \label{fig:llm_proposer}
\end{figure}

\noindent \textbf{Reflection after playout.}
Upon generation, the new configuration $\boldsymbol{\lambda}_{\text{new}}$ is evaluated on the training/validation split to obtain the performance metric $y_{\text{new}}$. To bridge the gap between numerical rewards and the LLM's semantic reasoning, we subsequently incorporate a reflection mechanism inspired by Reflexion~\cite{reflection_shinn2023reflexion}. 
Excluding the \textsc{Warmup} expansion of Algo nodes, we prompt LLM to analyze the evaluation result by comparing it against $\boldsymbol{\lambda}_{\text{base}}$ and configurations in memory $\mathcal{M}_i(s_\text{base})$:
\begin{equation}
\rho_{\text{new}}
= \mathrm{LLM}\big(\psi, \mathcal{M}_i(s_\text{base}), \boldsymbol{\lambda}_{\text{base}}, y_{\text{base}}, \boldsymbol{\lambda}_{\text{new}}, y_{\text{new}}\big).
\end{equation}
$\rho_{\text{new}}$ summarizes what worked or failed and provides instructions for future expansion, which is stored in the new HP node $s^\text{new}_\text{hp}$.
Prompt template and example are detailed in Appendix~\ref{app:prompt_reflection} and~\ref{app:example}.

\noindent \textbf{Selective Tuning Memory.}
Efficiently utilizing optimization history is crucial for ICL. 
Therefore, we propose the Selective Tuning Memory (STM) mechanism.
Given their disjoint search spaces, histories from different algorithms provide no mutual benefit.
Thus, we first isolate the optimization history so that the memory $\mathcal{M}_i(s_\text{base})$ only contains trials from the same algorithm subtree as $s_\text{base}$.
After that, we view each transition in the subtree (from a parent node $s_{p}$ to a child $s_{c}$) as an ``optimization attempt" $e = (\boldsymbol{\lambda}_{p}, y_{p}, \boldsymbol{\lambda}_{c}, y_{c}, \rho_{c})$, capturing the modification from the parent configuration $\boldsymbol{\lambda}_{p}$ to the child $\boldsymbol{\lambda}_{c}$, their performances, and reflection $\rho_{c}$. 
For nodes expanded through BO and \textsc{Warmup} lacking LLM reflection, reflections are programmatically generated via fixed rules (Appendix~\ref{app:reflection_for_bo_and_warmup}).

Given an HP node with configuration $\boldsymbol{\lambda}_{\text{base}}$ to expand, our goal is to retrieve historical attempts that are both relevant (similar starting point) and instructive (high resulting performance $y_{c}$). 
The similarity between a historical attempt's starting configuration $\boldsymbol{\lambda}_{p}$ and $\boldsymbol{\lambda}_{\text{base}}$ is measured through the algorithm's GP covariance kernel $k(\boldsymbol{\lambda}_{p}, \boldsymbol{\lambda}_{\text{base}})$ in \cref{eq:gp_kernel}, which jointly captures continuous-categorical similarity.
For initialization attempts with an Algo node parent, we assign a similarity of $-\infty$.
To balance relevance and performance, we compute the Pareto frontier over all historical attempts in the algorithm's subtree based on $(k(\cdot, \boldsymbol{\lambda}_{\text{base}}), y_{c})$, selecting the non-dominated attempts as \textbf{Global Memory} $\mathcal{M}_{\text{global}}$.
Complementing this, we include \textbf{Local Memory} $\mathcal{M}_{\text{local}}$, comprising the sequence of attempts along the ancestral path from the algorithm root to $s_\text{base}$. This preserves the optimization trajectory context. 
The final memory is the union 
$\mathcal{M}_i(s_{\text{base}}) = \mathcal{M}_{\text{global}} \cup \mathcal{M}_{\text{local}}$, providing both high-quality, relevant references and coherent trajectory.

\noindent \textbf{LLM proposer workflow.}
Overall, the LLM proposer follows PUCT-based base-node selection, ICL-based expansion, evaluation, reflection, and backpropagation, as illustrated in \cref{fig:llm_proposer}.

\subsubsection{Dynamic Proposer Selection}
\label{sec:dynamic_proposer_selection}
To synergize the complementary strengths of semantic reasoning and quantitative modeling, \sys dynamically selects either the LLM or BO proposer at each HPO step. 
Our insight is that BO becomes increasingly reliable as data accumulates and the surrogate model matures, whereas LLMs may excel in the early stages with prior knowledge.
Thus, for each algorithm $A^i$, we maintain a BO-selection probability $p^i_{\text{BO}}$ based on the generalization capability of its surrogate $f_i$.
Specifically, given the algorithm-specific dataset, we perform $k$-fold cross-validation, training $f_i$ on training folds and computing the Kendall's rank correlation coefficient ($\tau^i$) between predicted and true rankings on validation folds to quantify how well $f_i$ can capture the relative order of configurations. 
This correlation is rescaled to $[0, 1]$ as a probability: $p^i_{\text{BO}} = \text{max}(\epsilon, \frac{\tau^i + 1}{2})$, where a small constant $\epsilon$ (e.g., $0.05$) ensures the BO proposer remains active.
We initialize the BO probability $p^i_{\text{BO}}$ to $\epsilon$ and update it every $k$ observations (e.g., $k=5$).
In each iteration, the BO proposer is selected with probability $p^i_\text{BO}$ and the LLM proposer with probability $1 - p^i_\text{BO}$. 
This mechanism ensures a smooth transition from early LLM-driven search to BO-driven search as the surrogate model becomes accurate.

\subsection{Algorithm Summary}
The procedure of \sys is summarized in Appendix~\ref{app:full_algo}.
We run MCTS iteratively:
1) \textbf{Algorithm Selection}: The root selects an Algo node with PUCT.
2) \textbf{Proposer Selection}: For the chosen algorithm, we estimate the GP surrogate’s quality to select the BO or LLM proposer.
3) \textbf{Proposal}: The BO proposer samples candidates and selects with EI; the LLM proposer extends PUCT to select a base node and prompts LLM to generate a new configuration.
4) \textbf{Playout \& Reflection}: The configuration is evaluated, followed by a reflection, to add a new HP node.
5) \textbf{Backpropagation}: The result is propagated to update ancestral nodes' statistics.
After the budget is exhausted, \sys returns the configuration with the best validation performance.

\textbf{Difference with Previous Methods.}  
Unlike prior LLM optimizers for flat black-box problems, \sys targets structured CASH.
Unlike existing LLM--BO hybrids that pass flat histories to LLMs, \sys unifies BO- and LLM-produced trajectories in a shared tree state, enabling richer path-aware reasoning.
Unlike prior LLM--MCTS methods, \sys uses the MCTS tree as a shared CASH state, integrating algorithm selection and LLM--BO synergistic HPO into unified trajectories.

\begin{theorem}[\textbf{Global Convergence}]
\label{thm:convergence}
If the BO-selection probability is lower bounded by $\epsilon>0$, \sys converges to the global optimum almost surely as the number of iterations $T \to \infty$:
\begin{equation}
    \lim_{T \to \infty} \left( \mathcal{F}(\boldsymbol{\lambda}^*) - \max_{t=1}^T \mathcal{F}(\boldsymbol{\lambda}_t) \right) = 0,
\end{equation}
\end{theorem}
Appendix~\ref{app:proof} provides the detailed proof, including the finite-time regret bound in Theorem~\ref{thm:finite_time_regret} and the conditional evaluation-budget comparison in Corollary~\ref{cor:finite_time_acceleration}, 
showing that \sys inherits BO's convergence guarantee while leveraging LLMs for acceleration.

\section{Experiment}
\label{sec:experiment}
\subsection{Experiment Setup}

\textbf{Baselines.}
We compare \sys with 8 baselines: 
--- \textit{Four BO-based CASH methods}: 
1) \textbf{SMAC}~\cite{smac_hutter2011sequential}: standard BO;
2) \textbf{OptDivBO}~\cite{optdivbo_poduval2024cash}: a BO variant encouraging diversity;
3) Rising Bandit (\textbf{RB})~\cite{rb_li2020efficient}: a multi-armed-bandit extension of BO for CASH;
4) \textbf{MOSAIC}: a BO variant with PUCT for algorithm selection.
--- \textit{Two pure LLM Optimizer}:
5) \textbf{OPRO}, 
6) \textbf{LLAMBO}, 
--- \textit{Two hybrid LLM-BO Methods}:
7) \textbf{BOPRO}, which optimizes LLM's prompt instructions with BO; 
and 8) \textbf{BORA}, 
which switches between LLM and BO via a variance heuristic.

\textbf{CASH space, datasets, and metrics}.
We construct a hierarchical CASH search space comprising 8 machine learning algorithms (e.g., XGBoost) for both classification and regression tasks, resulting in high-dimensional search spaces with 45 and 46 hyperparameters, respectively (see Appendix~\ref{app:search_space}).
For datasets, we utilize the AutoML Benchmark (AMLB)~\cite{amlb_gijsbers2019open}, which contains 71 classification tasks and 33 regression tasks across varying scales and domains. Following~\cite{auto-sklearn_feurer2022auto}, we report Balanced Accuracy for classification and Mean Squared Error (MSE) for regression as performance metrics.

\textbf{Basic settings}.
Each dataset is split into training, validation, and test sets with a 60/20/20 ratio.
We report the best-observed validation metric during optimization and the final test metric.
While it takes a different amount of time to evaluate the same configuration on different datasets, we use the evaluation iterations as the unit of budget, where each baseline evaluates 300 configurations.
We implement the BO surrogate, the random and local sampling mechanism using OpenBox 0.8.1~\cite{openbox_jiang2024openbox}, an open-source toolkit.
For the algorithm prior, the number of sampled configurations $n_s$ is set to 100.
The exploration constant in PUCT is set to $c_{puct} = \sqrt{2}$.
We employ \texttt{gpt-4o-mini-2024-07-18}~\cite{gpt-4o-mini}, OpenAI's cost-efficient model, as the backbone for all LLM-based optimizers (ablation in Appendix~\ref{app:llm_backbone}).
More implementation details for other baselines are provided in Appendix~\ref{app:implementation_details}.

\begin{figure*}[tb]
    \centering

    \begin{subfigure}[b]{0.52\linewidth} %
        \centering
        \includegraphics[width=\linewidth]{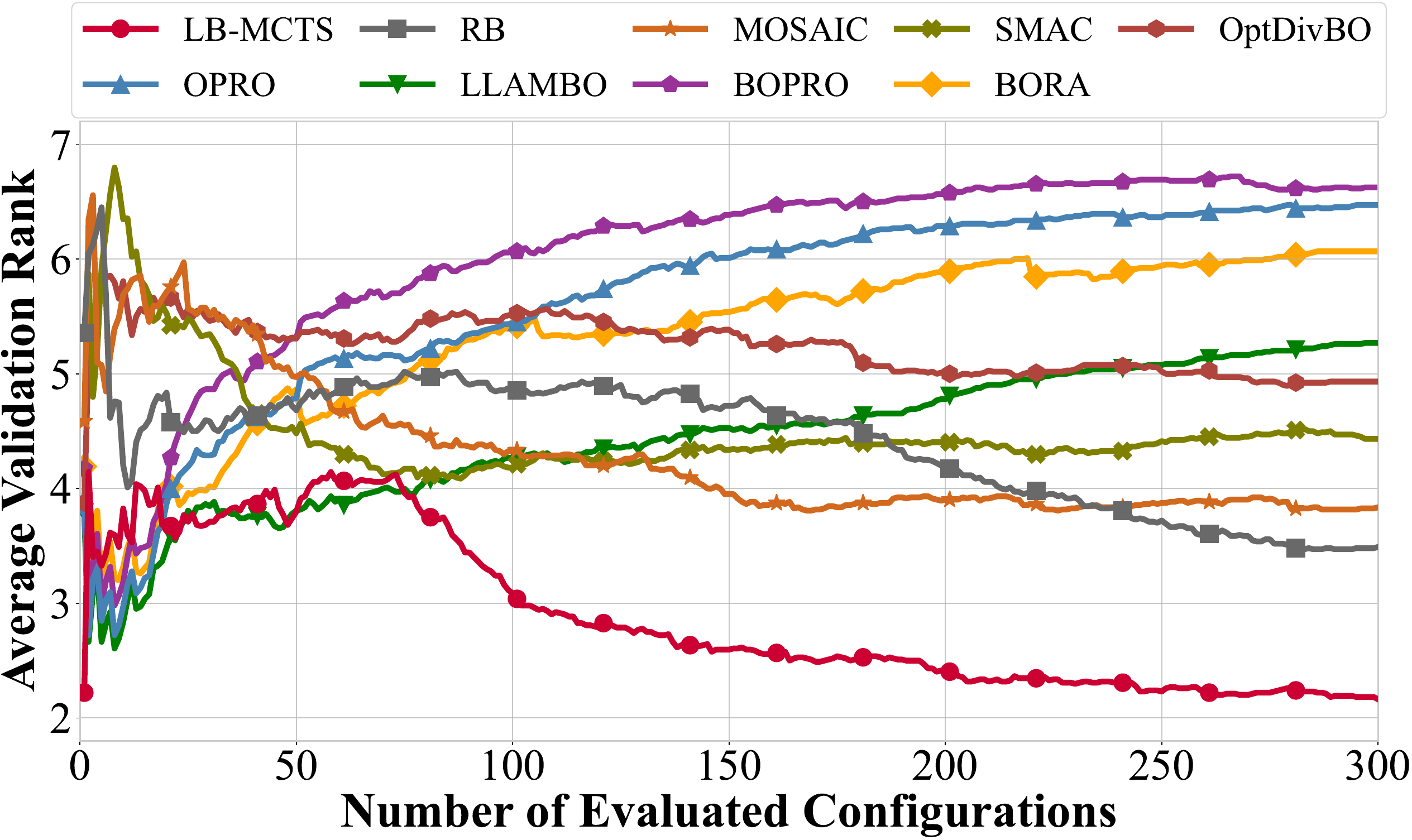}
        \caption{Average validation rank during optimization.}
        \label{fig:validation_rank}
    \end{subfigure}
    \hfill %
    \begin{subfigure}[b]{0.47\linewidth} %
        \centering
        \includegraphics[width=\linewidth]{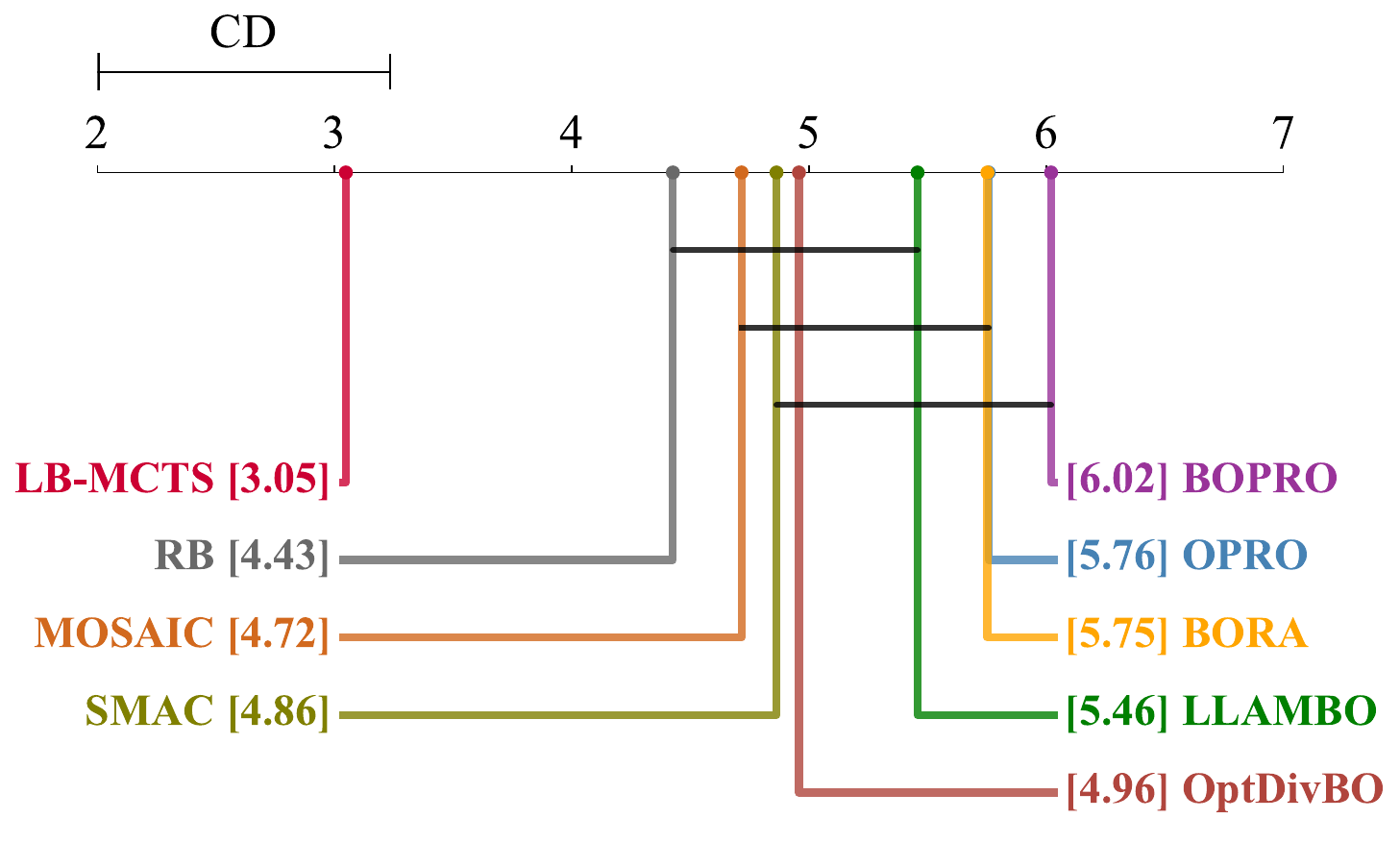}
        \caption{CD plot of test rank with Nemenyi post-hoc test.}
        \label{fig:test_cd}
    \end{subfigure}
    
    \caption{Average Performance of 9 methods across 104 datasets.}
    \label{fig:main_figure} %
\end{figure*}

\subsection{Comparison with Baselines}
This section evaluates \sys on 104 AMLB datasets. 
\cref{fig:validation_rank} displays the average rank of the \textit{best-achieved validation performance} at each iteration, from which we draw four observations:
(i)  Current LLM-based baselines (BORA, LLAMBO, BOPRO, and OPRO) perform poorly on CASH.
They are competitive only in the first $\sim$50 iterations; afterwards their average ranks steadily deteriorate, ending between 5.27 and 6.63—worse than other methods.
This is because they treat CASH as a flat joint space, mix histories across algorithms, and lack explicit exploration-exploitation control, getting stuck in local optima in the structured, high-dimensional search space.
(ii) Among LLM-based optimizers, OPRO and BOPRO perform worst, as they rely on simplistic iterative prompting where BO (in BOPRO’s case) is limited to superficial example selection. 
BORA (rank 6.07) gains slightly via LLM--BO switching, but its behavior is governed by fixed uncertainty thresholds that do not transfer well to CASH.
LLAMBO attains the highest rank (5.27) but incurs prohibitive computational overhead.
(iii) Among BO-based baselines, OptDivBO performs worst (rank 4.93) as it prioritizes ensemble diversity over individual model performance. MOSAIC improves on SMAC by decoupling algorithm selection from HPO. RB leads BO baselines by using local surrogates after algorithm selection instead of a global surrogate over the joint space.
(iv) Among all methods, \sys significantly dominates the others.  
It consistently outperforms all baselines beyond around 70 iterations. While the second-best baseline (RB) ranks 3.49 finally, \textbf{the rank of \sys is 2.16}. 
Appendix~\ref{app:task_characteristics} further analyzes these results by task characteristics, showing that \sys remains robust across diverse tasks.
To further assess generality beyond conventional ML algorithms, Appendix~\ref{app:dl_fe_experiment} reports supplementary experiments on a broader search space including deep learning models and feature engineering, where \sys continues to achieve the best average rank.

\textbf{Test set performance}. We further \textit{evaluate the validation-best configurations on the test set} and illustrate the results in the Critical Difference (CD) diagram (\cref{fig:test_cd}).
We observe that the ranks of methods on the test set may deviate from their validation ranks. The reason is that the distributions of the validation and test set are not exactly the same \cite{hutter2019automated}. 
But overall, \sys remains statistically significantly superior to all baselines, attaining a \textbf{leading average rank of 3.05}.
To further assess the practical utility, we evaluate the post-hoc ensemble performance of all methods in Appendix~\ref{app:ensemble}.

The success of \sys stems from two main factors.
First, \sys remedies key limitations of prior LLM-based optimizers:
(i) The STM mechanism isolates algorithm-specific histories and combines global and local memory to enhance reasoning (validated in Appendix~\ref{app:ablation_memory}).
(ii) As shown in \cref{sec:exploration-exploitation_trade-off}, MCTS enables a principled exploration–exploitation trade-off both across algorithms and within each algorithm’s subtree.
(iii) The reflection mechanism further strengthens STM by turning numeric feedback into reusable verbal guidance (Appendix~\ref{app:ablation_reflection}).
Second, \sys achieves optimal synergy between LLM and BO proposers, outperforming either component alone (\cref{sec:bo_llm_analysis}).
Appendix~\ref{app:cost} provides analyses of both API cost and wall-clock overhead, showing that \sys achieves these gains with an economical cost of about $\$0.127$ per task.

\subsection{MCTS for Exploration-Exploitation Trade-off}
\label{sec:exploration-exploitation_trade-off}
In this section, we validate the ability of \sys to trade off exploration and exploitation at two levels: (1) Can it allocate optimization resources to superior algorithms? (2) Within each algorithm's subspace, can it effectively navigate the trade-off between promising and unexplored regions?

\begin{figure}[t!]
    \centering
    \begin{minipage}[t]{0.47\linewidth}
        \centering
        \includegraphics[width=\linewidth]{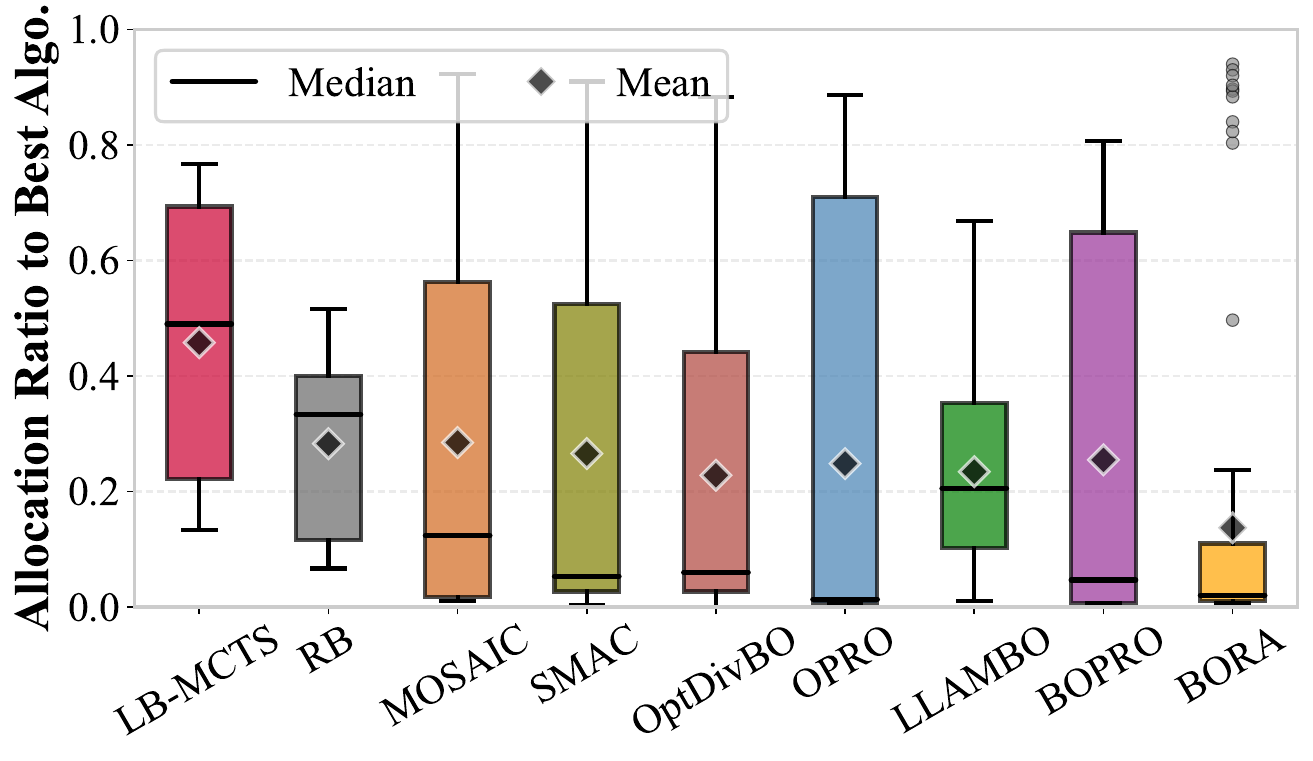}
        \caption{Resource allocation.}
        \label{fig:ee_algorithm}
    \end{minipage}
    \hfill
    \begin{minipage}[t]{0.49\linewidth}
        \centering
        \includegraphics[width=\linewidth]{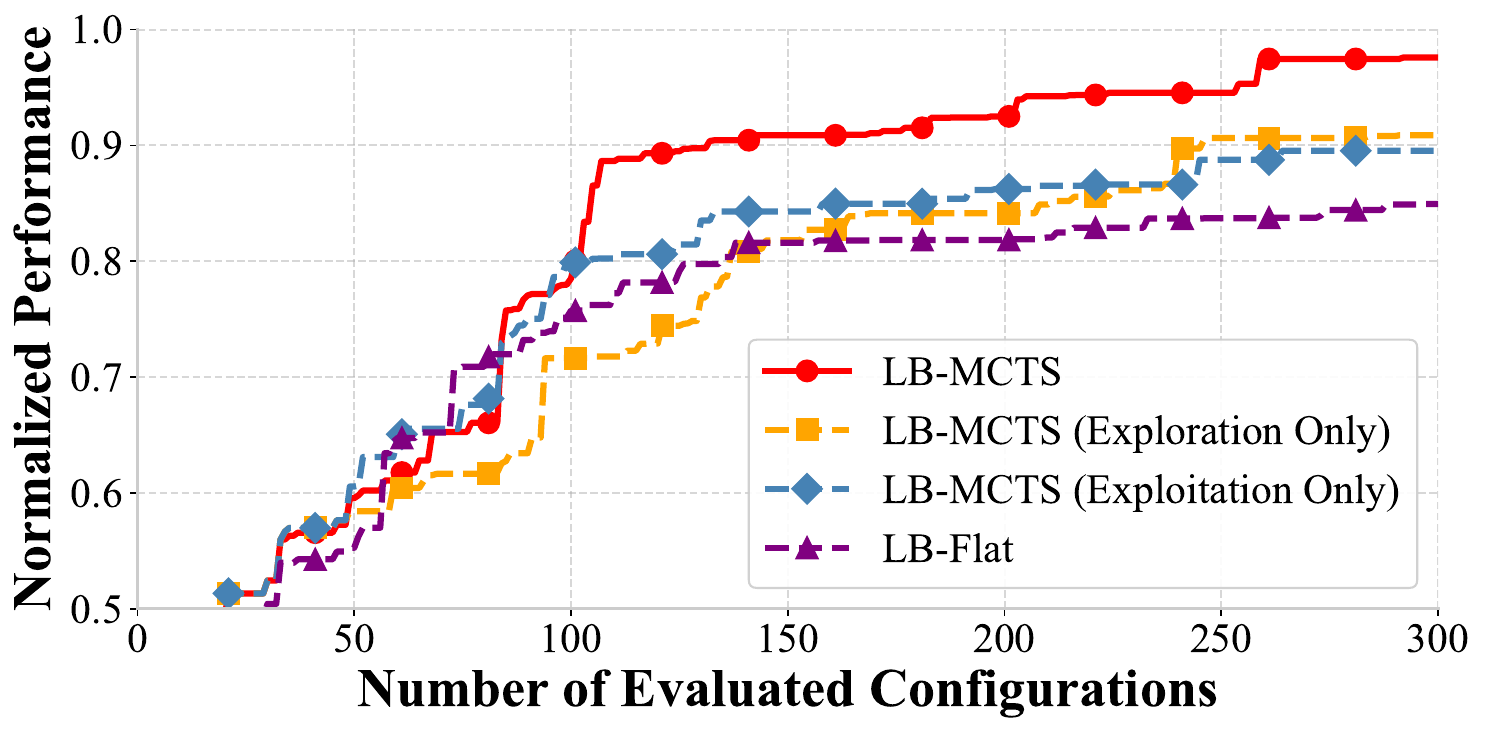}
        \caption{Impact of the tree-structured search.}
        \label{fig:ee_hpo}
    \end{minipage}
\end{figure}

\textbf{Resource allocation across algorithms}. 
For each dataset, we first identified the optimal algorithm as the algorithm associated with the global best configuration found across all 9 methods and 300 iterations. 
Subsequently, for each CASH method, we compute the \textit{ratio of its evaluated configurations that belong to the ``optimal algorithm''}.
Figure~\ref{fig:ee_algorithm} visualizes the distribution of these allocation ratios for each method across the 104 datasets, from which we get two observations:
(i) Superiority: \sys achieves the highest mean allocation ratios (0.43), whereas other baselines allocate 0.14 to 0.28 on average. This demonstrates its superior ability to correctly identify the most promising algorithm and concentrate optimization budget on it.
(ii) Robustness: \sys exhibits the highest median (0.49), along with highest lower quartile (box bottom) and lower whisker, ensuring the optimal algorithm is rarely overlooked. In contrast, aggressive baselines (e.g., SMAC, OPRO, MOSAIC) exhibit near-zero medians and bottom whiskers, indicating frequent premature convergence to suboptimal algorithms; methods like RB and LLAMBO yield consistently low ratios due to conservative uniformity.
In summary, \sys employs root-level PUCT to balance inter-algorithm trade-off, ensuring stable resource allocation to optimal algorithms across datasets.

\textbf{Navigating within each algorithm's subspace}.
Effective HPO requires balancing exploiting promising regions and exploring unknown areas to avoid local optima. To achieve this, \sys explicitly prompts the LLM proposer to \textit{expand the tree structure using two directives}: \textsc{Exploitation} for local refinement and \textsc{Exploration} for global search. 
We compare this design against two single directive variants—\textsc{Exploration} only and \textsc{Exploitation} only—expanded twice per node, and LB-Flat, which retains only the root and Algo Nodes, substituting subsequent tree search with unstructured history sharing between BO and LLM (an OPRO-style proposer).
To conserve API costs, all ablation studies are conducted on 10 representative datasets from AMLB (details in Appendix~\ref{app:dataset_selection}). 
\cref{fig:ee_hpo} plots the average \textit{best-so-far validation performance} across the 10 datasets, where scores on each dataset are min-max normalized globally.
The results show that relying solely on \textsc{Exploitation} yields rapid initial gains but leads to premature stagnation, trapped in local optima. Conversely, using only \textsc{Exploration} results in slower convergence due to insufficient refinement of promising regions. 
\sys outperforms both variants and LB-Flat, finally achieving \textbf{an average rank of 1.5}, compared to 2.3 and 2.4 for the two variants.
Most importantly, the poor performance of LB-Flat (rank 3.4) underscores that unstructured history sharing—a common trait in existing methods—is insufficient for complex CASH problems without the hierarchical guidance of MCTS.
Details of LB-Flat and further analysis are provided in Appendix~\ref{app:ablation_exploration_exploitation}. 
In summary, the synergy of dual directives and tree-structured search enables \sys to effectively navigate the search space to avoid the local optima.
Additional ablations in Appendix~\ref{app:ablation_puct} validate the choice of PUCT over alternative tree-search strategies and show that $c_{puct}=\sqrt{2}$ provides robust exploration.

\subsection{Analysis of BO--LLM Collaboration}
\label{sec:bo_llm_analysis}
\begin{figure}[t!]
    \centering
    \begin{minipage}[t]{0.48\linewidth}
        \centering
        \includegraphics[width=\linewidth]{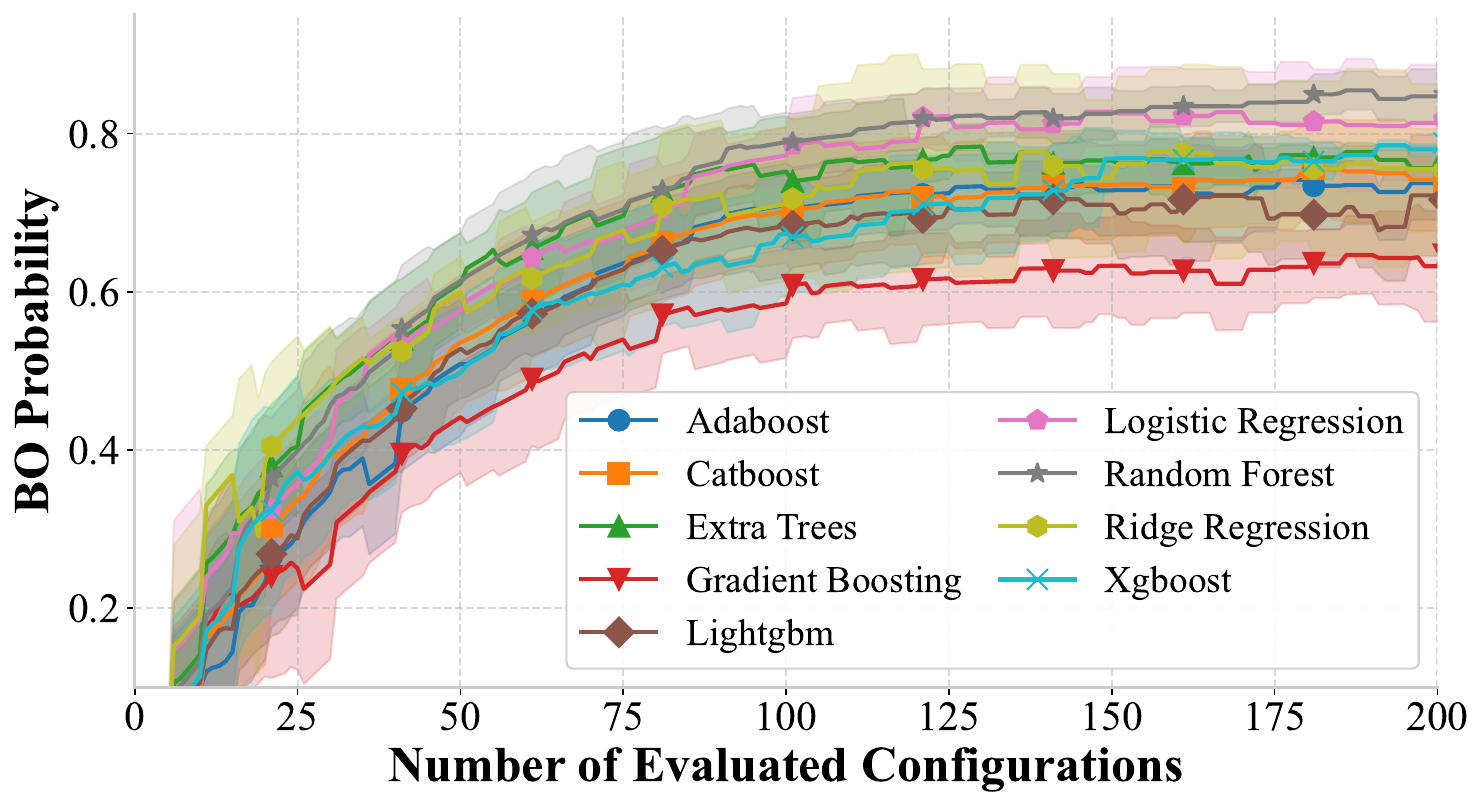}
        \caption{Average $P_{\text{BO}}$ for each algorithm.}
        \label{fig:bo_prob}
    \end{minipage}
    \hfill
    \begin{minipage}[t]{0.48\linewidth}
        \centering
        \includegraphics[width=\linewidth]{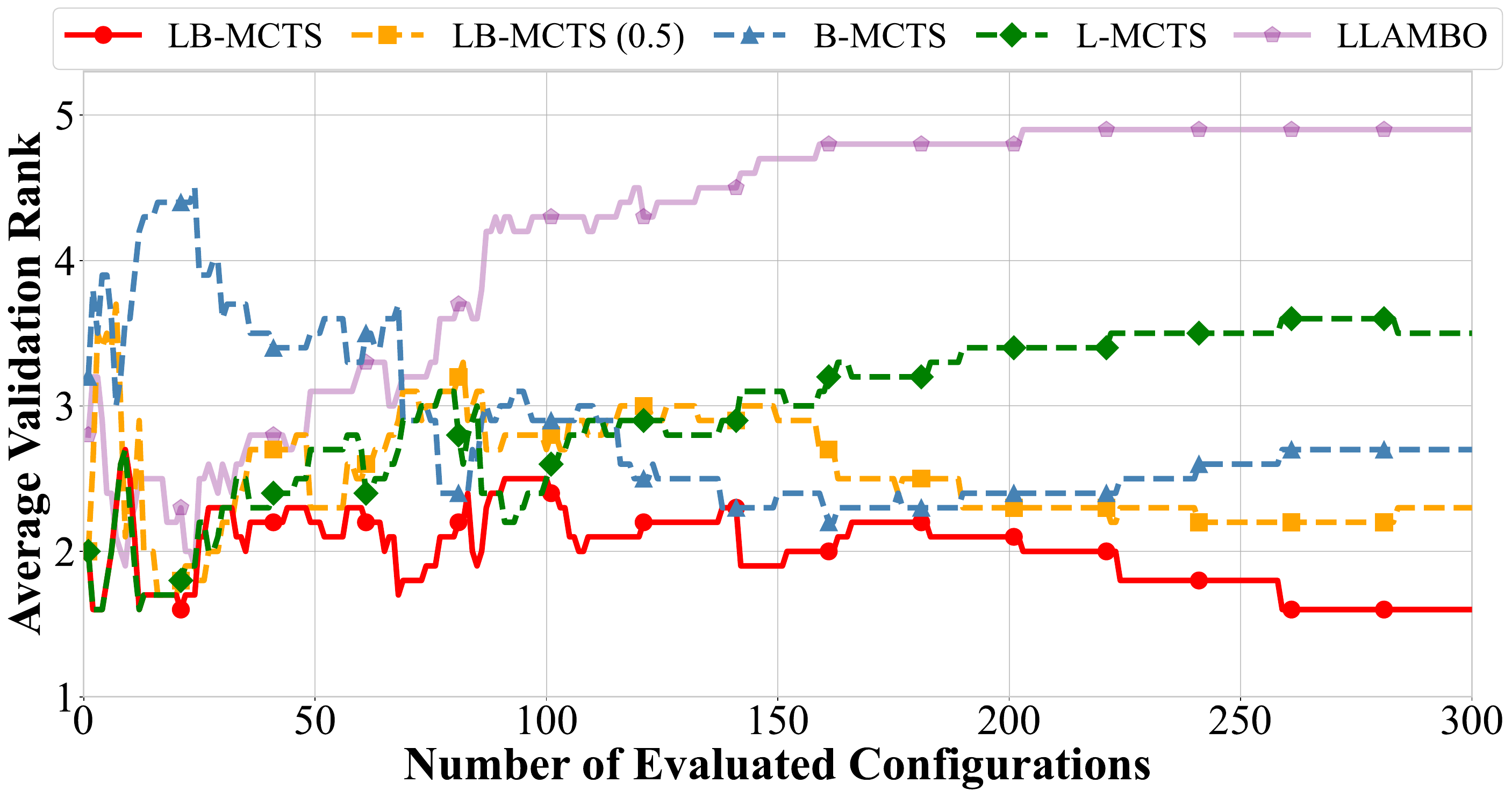}
        \caption{Average rank during optimization.}
        \label{fig:validation_rank_bollm}
    \end{minipage}
\end{figure}

\sys maintains a BO-selection probability ($P_{\text{BO}}$) for each algorithm and samples the proposer at every HPO step. \cref{fig:bo_prob} plots the average $P_{\text{BO}}$  for each algorithm versus the number of evaluations over 104 datasets. 
Initially ($N < 25$), $P_{\text{BO}}$ is low ($< 0.3$), reflecting reliance on semantic priors when data are scarce.
As the number of evaluations grows, the BO surrogate becomes more reliable:
around $\sim 50$ evaluations, $P_{\text{BO}}$ rises to $\approx 0.5$, marking a transition. 
After $\sim 100$ evaluations, it gradually stabilizes between $0.6$ and $0.8$, with modest variation across algorithms (likely reflects differences in landscape smoothness). 
On average, the LLM proposer is selected in 107 out of 300 iterations.

To quantify the benefit of this dynamic synergy, we perform an ablation study on 10 datasets against three variants: L-MCTS (pure LLM), B-MCTS (pure BO), and LB-MCTS(0.5) (fixed 50:50 probability). 
We also include LLAMBO, the strongest LLM-based baseline. \cref{fig:validation_rank_bollm} tracks the average rank of the best validation performance. 
(i) L-MCTS dominates early ($<70$ iterations), while B-MCTS prevails later ($>110$ iterations), confirming LLMs excel at warm-starting while BO excels with sufficient data.
(ii) Benefit of synergy: both LB-MCTS (0.5) and LB-MCTS outperform single-proposer variants.
(iii) \sys achieves the best performance over the entire optimization, achieving a final \textbf{rank of 1.6} versus 2.3 for LB-MCTS (0.5), confirming the value of dynamic synergy.
Notably, \sys outperforms L-MCTS even in the early phase, suggesting that occasional injections from the non-dominant proposer provide necessary diversity for performance gains.
(iv) Finally, L-MCTS (rank 3.5) significantly outperforms LLAMBO (rank 4.9), reaffirming that our trajectory-aware MCTS enables more effective LLM reasoning than unstructured prompting.

\section{Limitation and Future Work}
\label{sec:limitation_and_futurework}

Our method inherits the general challenge that validation feedback can be noisy in some scenarios, such as when the validation set is small or the data itself is highly noisy. In these cases, the resulting feedback signal may be less reliable and can occasionally mislead MCTS and the LLM proposer. While BO has built-in mechanisms to better tolerate noisy evaluations, the current LLM proposer does not explicitly model such noise. A promising direction for future work is therefore to develop noise-aware LLM tuning strategies, such as prompt designs or feedback processing schemes that are more robust to uncertain validation signals.

In addition, the effectiveness of our framework still depends to some extent on the capability of the underlying LLM. Although our experiments show that several relatively economical models, such as GPT-4o-mini and DeepSeek-V3, already perform well in our setting, weaker models may nevertheless provide less informative proposals and thus limit the overall gain. Future improvements could come from lightweight domain adaptation, fine-tuning, or the incorporation of external tools to further strengthen the tuning ability of smaller or less capable models.

From a broader-impact perspective, \sys may lower the expertise and cost barriers for applying AutoML by improving the efficiency of CASH optimization. However, as with other AutoML systems, the selected models may still inherit biases, privacy risks, or reliability issues from the underlying data and downstream application context. Therefore, when applying \sys to sensitive domains, users should conduct appropriate data governance, fairness checks, and domain-specific validation.

\section{Conclusion}

In this paper, we propose \sys, which synergizes LLMs and BO via tree search for the CASH problem. By organizing algorithm selection and hyperparameter refinement in a shared tree, \sys unifies the optimization trajectories produced by both proposers and enables them to build on accumulated search experience. The framework combines hierarchical exploration and exploitation, Selective Tuning Memory for algorithm-specific context retrieval, and a reliability-aware policy for adaptive proposer selection.
Experiments on 104 AMLB datasets demonstrate that \sys outperforms BO-based, LLM-based, and hybrid baselines. Ablation studies further support the contributions of the search structure, memory mechanism, and BO--LLM collaboration.

\begin{ack}
This work is supported by the National Natural Science Foundation of China (U23B2048, U22B2037), the Beijing Municipal Natural Science Foundation (Grant No. L2603008), and the ``Fundamental Research Funds for the Central Universities, Peking University''.
Bin Cui is the corresponding author.
\end{ack}

\bibliographystyle{plainnat}
\bibliography{LB-MCTS}

\clearpage
\appendix
\section{Method Details}

\subsection{Pseudocode of \sys}
\label{app:full_algo}

\begin{algorithm}[!ht]
  \caption{\sys: LB-MCTS for CASH}
  \label{alg:lbmcts}
  \begin{algorithmic}[1]
    \STATE {\bfseries Input:} dataset $\mathcal{D}$, algorithms $\mathcal{A}=\{A^1,\dots,A^K\}$ with spaces $\{\Lambda^i\}$, evaluation budget $T$
    \STATE Initialize root node $s_{\text{root}}$ with Algo children $\{s_{\text{algo}}^{i}\}_{i=1}^K$
    \STATE Initialize per-algorithm data $\mathcal{D}_i \leftarrow \emptyset$, surrogate $f_i$ (GP), episode set $\mathcal{E}_i \leftarrow \emptyset$, $p^i_{\text{BO}} \leftarrow \epsilon$
    \STATE Initialize node stats $N_s \leftarrow 0$, $R_s \leftarrow 0$, $y^{\max}_s \leftarrow -\infty$ for all $s$
    \STATE $t \leftarrow 0$
    \WHILE{$t < T$}
      \vspace{0.5em}
      \STATE \textbf{// Algorithm selection (root PUCT)}
      \STATE Compute $Q(s_{\text{algo}}^{i})$ and prior $P(s_{\text{algo}}^{i})$ for all $i$ (Sec.~\ref{sec:algorithm_selection})
      \STATE Select algorithm node $s_{\text{algo}}^{i}$ by PUCT (Eq.~\ref{eq:puct}); let $A^i$ be the chosen algorithm

      \vspace{0.5em}
    \STATE \textbf{// Dynamic proposer selection}
    \IF{$|\mathcal{D}_i| > 0$ \AND $|\mathcal{D}_i| \pmod 5 = 0$}
        \STATE Estimate surrogate quality $\tau^i$ via $k$-fold CV on $\mathcal{D}_i$
        \STATE $p^i_{\text{BO}} \leftarrow \text{max}(\epsilon, \frac{\tau^i+1}{2})$
    \ENDIF
    \STATE Sample $z \sim \mathrm{Bernoulli}(p^i_{\text{BO}})$

      \vspace{0.5em}
      \STATE \textbf{// HPO with LLM/BO}
      \IF{$z = 1$} 
        \STATE \COMMENT{use BO proposer}
        \STATE $(\boldsymbol{\lambda}_{\text{new}}, s_{\text{parent}}) \leftarrow \textsc{BOPropose}(A^i, s_{\text{algo}}^{i}, f_i, \mathcal{D}_i)$
      \ELSE 
        \STATE \COMMENT{use LLM proposer}
        \STATE $(\boldsymbol{\lambda}_{\text{base}}, s_{\text{base}}) \leftarrow \textsc{PUCTSelectHP}(s_{\text{algo}}^{i})$  \COMMENT{HP-level PUCT with uniform prior}
        \STATE Build STM $\mathcal{M}_i(s_{\text{base}})$ from global episodes $\mathcal{E}_i$ and ancestors of $s_{\text{base}}$
        \STATE Choose directive $d \in \{\textsc{Warmup}, \textsc{Exploration}, \textsc{Exploitation}\}$ for $s_{\text{base}}$
        \STATE $(\mathcal{R},\,\boldsymbol{\lambda}_{\text{new}}) \leftarrow \mathrm{LLM}\big(\psi, \mathcal{M}_i(s_{\text{base}}), \boldsymbol{\lambda}_{\text{base}}, d\big)$
        \STATE $s_{\text{parent}} \leftarrow s_{\text{base}}$
      \ENDIF

      \vspace{0.5em}
      \STATE \textbf{// Playout and reflection}
      \STATE Evaluate $A^i$ with $\boldsymbol{\lambda}_{\text{new}}$ on $\mathcal{D}$ to obtain $y_{\text{new}}$
      \STATE Let $(\boldsymbol{\lambda}_{\text{base}}, y_{\text{base}})$ be the config/performance of $s_{\text{parent}}$ (if any)

      \IF{$z = 1$}  
        \STATE \COMMENT{BO: heuristic reflection}
        \STATE $\rho_{\text{new}} \leftarrow \textsc{SynthesizeSummary}(\Delta\boldsymbol{\lambda}, \Delta y)$ 
      \ELSE 
        \STATE \COMMENT{LLM proposer: verbal reflection}
        \STATE $\rho_{\text{new}} \leftarrow \mathrm{LLM}\big(\psi, \mathcal{M}_i(s_{\text{parent}}), \boldsymbol{\lambda}_{\text{base}}, y_{\text{base}}, \boldsymbol{\lambda}_{\text{new}}, y_{\text{new}}\big)$
      \ENDIF
      \STATE Create new HP node $s_{\text{new}}$ with $(\boldsymbol{\lambda}_{\text{new}}, y_{\text{new}}, \rho_{\text{new}})$ as child of $s_{\text{parent}}$
      \STATE Add $(\boldsymbol{\lambda}_{\text{base}}, y_{\text{base}}, \boldsymbol{\lambda}_{\text{new}}, y_{\text{new}}, \rho_{\text{new}})$ to episode set $\mathcal{E}_i$
      \STATE Update dataset $\mathcal{D}_i \leftarrow \mathcal{D}_i \cup \{(\boldsymbol{\lambda}_{\text{new}}, y_{\text{new}})\}$ and refit GP $f_i$ (periodically)

      \vspace{0.5em}
      \STATE \textbf{// Backpropagation}
      \FOR{each node $s$ on path from $s_{\text{root}}$ to $s_{\text{new}}$}
        \STATE $r_s \leftarrow [y_{\text{new}} - y^{\max}_s]_+ / |y^{\max}_s|$
        \STATE $N_s \leftarrow N_s + 1$;\quad $R_s \leftarrow R_s + r_s$;\quad $y^{\max}_s \leftarrow \max(y^{\max}_s, y_{\text{new}})$
      \ENDFOR

      \STATE $t \leftarrow t + 1$
    \ENDWHILE
    \STATE {\bfseries Output:} best $(A^*, \boldsymbol{\lambda}^*)$ observed so far (maximizing validation performance)
  \end{algorithmic}
\end{algorithm}

Algorithm~\ref{alg:lbmcts} summarizes the procedure of \sys. The detailed workflow is as follows.

\textbf{Initialization} (Lines~2--5).
Given a dataset $\mathcal{D}$ and a set of candidate algorithms $\mathcal{A}=\{A^1,\dots,A^K\}$ with corresponding hyperparameter spaces $\{\Lambda^i\}_{i=1}^K$, we construct the search tree with a root CASH node $s_{\text{root}}$ and $K$ Algo children $\{s_{\text{algo}}^{i}\}_{i=1}^K$ (Lines~2--3). For each algorithm $A^i$, we maintain:
(i) a local dataset $\mathcal{D}_i$ of evaluated configurations,
(ii) a GP surrogate $f_i$ over $\Lambda^i$,
(iii) an episode set $\mathcal{E}_i$ storing optimization attempts (parent/child configurations, performances, and reflections).

While the budget $T$ is not exhausted, \sys execute an MCTS loop over the CASH tree:

\textbf{Algorithm selection at the root} (Lines~7--9);~\cref{sec:algorithm_selection}. At each iteration, \sys first decides which algorithm to optimize by applying the PUCT rule (\cref{eq:puct}) to select an Algo node $s_{\text{algo}}^{i}$. 

\textbf{Dynamic proposer selection} (Lines~10--15);~\cref{sec:dynamic_proposer_selection}. Given the chosen algorithm $A^i$, we decide whether the next configuration should be proposed by BO or LLM. We estimate the generalization quality of the surrogate $f_i$ on the current local dataset via cross-validation, compute the BO selection probability, and sample $z\sim\mathrm{Bernoulli}(p^i_{\text{BO}})$. When the surrogate is still inaccurate (small $\tau^i$), the LLM proposer is chosen more often; as more data accumulate and the surrogate improves, BO is selected with higher probability.
Specifically, $p^i_{\text{BO}}$ is initialized to $\epsilon$ and updated independently for each algorithm $A^i$ \textbf{whenever five new configurations have been added to its local history}.

\textbf{Proposing a new configuration.} (Lines~16--27)
The configuration proposal depends on whether $z=1$ (BO) or $0$ (LLM).
\begin{itemize}[leftmargin=1.6em, topsep=0pt, partopsep=0pt, itemsep=2pt, parsep=0pt]
    \item \textbf{BO proposer} (Line~17--19);~\cref{sec:bo_proposer}. 
    When $z=1$, we call \textsc{BOPropose} with the current algorithm $A^i$, its Algo node $s_{\text{algo}}^{i}$, the surrogate $f_i$, and data $\mathcal{D}_i$. This routine: 1) constructs a candidate pool by mixing \emph{random search} candidates drawn uniformly from $\Lambda^i$ (initialization actions), and \emph{local search} candidates sampled in the neighborhoods of previously evaluated configurations; 2) scores all candidates with an acquisition function (we use EI under the GP $f_i$); 3) selects the candidate $\boldsymbol{\lambda}_{\text{new}}$ with the highest EI and identifies its parent node $s_{\text{parent}}$ in the tree (the Algo node for random samples or the corresponding HP node for local-search samples).

    \item \textbf{LLM proposer} (Lines~20--26);~\cref{sec:llm_proposer}. When $z=0$, we use the LLM proposer. 
    1) First, we extend the MCTS Selection phase into the algorithm subtree by applying HP-level PUCT (Line~22). This yields a base HP node $s_{\text{base}}$ and its configuration $\boldsymbol{\lambda}_{\text{base}}$. 
    2) Next, we build the Selective Tuning Memory $\mathcal{M}_i(s_{\text{base}})$ (Line~23), which combines: a \emph{global memory} of Pareto-selected episodes and a \emph{local memory} consisting of all HP nodes along the ancestral path from $s_{\text{algo}}^i$ to $s_{\text{base}}$. 
    3) Then we choose a directive $d \in \{\textsc{Warmup}, \textsc{Exploration}, \textsc{Exploitation}\}$ according to the expansion state of $s_{\text{base}}$ (Line~24).
    4) We then query the LLM to obtain a reasoning chain $\mathcal{R}$ and a proposed configuration $\boldsymbol{\lambda}_{\text{new}}$ (Line~25). 
    In this case, the parent node is $s_{\text{parent}} = s_{\text{base}}$ (Line~26).
\end{itemize}

\textbf{Playout and reflection} (Lines~28--37).
As described in \cref{sec:llm_proposer}, the newly proposed configuration $\boldsymbol{\lambda}_{\text{new}}$ is evaluated on $\mathcal{D}$ to obtain the validation performance $y_{\text{new}}$ (Line~29). 
We then attach a reflection $\rho_{\text{new}}$ to this trial, with different generation mechanisms depending on the proposer:

\begin{itemize}[leftmargin=1.6em, topsep=0pt, partopsep=0pt, itemsep=2pt, parsep=0pt]
    \item \textbf{BO proposer} (Lines~32-33). We call a \textsc{SynthesizeSummary} to generate a heuristic textual reflection from the parameter change $\Delta \boldsymbol{\lambda}$ and performance change $\Delta y$ (Line~33). This yields short descriptions such as ``\emph{Increasing depth from 4 to 6 improved accuracy by 1.2\%}''.
    \item \textbf{LLM proposer} (Lines~35-36). We call the LLM to produce a semantic reflection, which explains why the modification helped or hurt and suggests how to adjust future configurations.
\end{itemize}

We then create a new HP node $s_{\text{new}}$ as a child of $s_{\text{parent}}$, storing $(\boldsymbol{\lambda}_{\text{new}}, y_{\text{new}}, \rho_{\text{new}})$. The episode $(\boldsymbol{\lambda}_{\text{base}}, y_{\text{base}}, \boldsymbol{\lambda}_{\text{new}}, y_{\text{new}}, \rho_{\text{new}})$ is added to the episode set $\mathcal{E}_i$, and the local dataset $\mathcal{D}_i$ is augmented with the new pair $(\boldsymbol{\lambda}_{\text{new}}, y_{\text{new}})$; the surrogate $f_i$ is periodically refit on $\mathcal{D}_i$ (Lines~38--40).

\textbf{Backpropagation} (Lines~41--46).
Finally, we propagate the result back through the tree to update node statistics. 

\textbf{Output}  (Line~49)
After the evaluation budget $T$ is exhausted, \sys returns the best combination $(A^i, \boldsymbol{\lambda})$ observed so far in terms of validation performance, providing a CASH solution obtained through the coordinated action of MCTS-guided selection, BO-based quantitative search, and LLM-based qualitative reasoning.

\subsection{Finite-Time Regret Analysis}
\label{app:proof}

We provide a finite-time version of Theorem~\ref{thm:convergence}.
Let $A^*$ denote an algorithm branch that contains a global maximizer, and let
$N_*(T)$ be the number of visits to this branch within a total budget $T$.
Among these visits, let $M_*(T)$ be the number of times the BO proposer is selected.
The global simple regret is
\[
    S_T^{\mathrm{Global}}
    =
    \mathcal{F}(\boldsymbol{\lambda}^*) -
    \max_{1 \le t \le T} \mathcal{F}(\boldsymbol{\lambda}_t).
\]

\begin{assumption}[Finite-time root allocation]
\label{ass:root_allocation}
Let $*$ be the index of the optimal branch $A^*$.
Let $N_i(t)$ be the number of visits to algorithm branch $A^i$ after $t$ root selections, and denote by $Q_t(A^i)$ and $P_i(t)$ its root value and prior used at round $t$.
For fixed gaps $\Delta_i>0$, $i\ne *$, define the number of gap violations by
\[
    Z_i(T)=\sum_{t=1}^{T}
    \mathbf 1\{Q_t(A^*)-Q_t(A^i)<\Delta_i\}.
\]
For each budget $T$ and $\rho\in(0,1)$, assume deterministic nonnegative budgets $B_i(T,\rho)$ such that
\[
    \Pr\!\left(Z_i(T)\le B_i(T,\rho),\ \forall i\ne *\right)
    \ge 1-\rho.
\]
This condition allows early errors and misrankings caused by limited or noisy validation feedback, instead of requiring a fixed gap at every round.
Without a concentration result for the root-value estimator, we do not specify a closed-form dependence of $B_i(T,\rho)$ on the noise level.

Under the PUCT selection rule in Eq.~\eqref{eq:puct}, if a suboptimal branch $A^i$ is selected at round $t$ while its gap condition holds, then
\[
    Q_t(A^i)+c_{puct}P_i(t)\frac{\sqrt{t-1}}{1+N_i(t-1)}
    \ge
    Q_t(A^*)+c_{puct}P_*(t)\frac{\sqrt{t-1}}{1+N_*(t-1)}
    \ge
    Q_t(A^*).
\]
Here the visit counts are taken before the current selection.
Combining the gap condition with this inequality and using $P_i(t)\le1$ gives
\[
    1+N_i(t-1)
    \le
    \frac{c_{puct}P_i(t)\sqrt{t-1}}{\Delta_i}
    \le
    \frac{c_{puct}\sqrt{T}}{\Delta_i}.
\]
Since the count increases by one at each selection, at most $c_{puct}\sqrt T/\Delta_i$ selections of $A^i$ can satisfy this inequality.
All remaining selections of $A^i$ occur on gap-violation rounds and number at most $Z_i(T)$.
Consequently, on the assumed event, $N_i(T)\le c_{puct}\sqrt T/\Delta_i+B_i(T,\rho)$, and
\[
    N_*(T)=T-\sum_{i\ne *}N_i(T)
    \ge T-\sum_{i\ne *}\left[
        \frac{c_{puct}\sqrt T}{\Delta_i}+B_i(T,\rho)
    \right].
\]
Thus larger violation budgets weaken the allocation guarantee; $B_i(T,\rho)=0$ recovers the original gap-based bound.

\textbf{Controlled misranking.}
Let $B_\Sigma(T,\rho)=\sum_{i\ne *}B_i(T,\rho)$.
For fixed $K$, gaps, and $\rho$, if $B_\Sigma(T,\rho)=\mathcal O(\sqrt T)$, then $N_*(T)=T-\mathcal O(\sqrt T)$ on the above event, retaining the near-full allocation order with probability at least $1-\rho$ at each horizon.
If $B_\Sigma(T,\rho)=o(\sqrt T)$, violations contribute only a lower-order penalty.

\textbf{Persistent misranking.}
If $B_\Sigma(T,\rho)=\Theta(T)$, the gap-based bound can become vacuous.
Nevertheless, for fixed $K$ and $c_{puct}>0$, if root values are uniformly bounded and root priors have a uniform positive lower bound, PUCT still guarantees $N_*(T)=\Omega(\sqrt T)$, independently of the gap-violation budget.
To see this, suppose otherwise: along some sequence of horizons $T\to\infty$, $N_*(T)=o(\sqrt T)$.
Throughout the last half of each such horizon, the exploration bonus of $A^*$ would then diverge uniformly, since its visit count never exceeds $N_*(T)$.
With bounded root values, any selected branch $A^j$ would eventually need an exploration bonus greater than one, implying $1+N_j(t-1)<c_{puct}\sqrt T$ because $P_j(t)\le1$.
Each branch can be selected only $\mathcal O(\sqrt T)$ times under this count restriction.
For finite $K$, this permits only $\mathcal O(K\sqrt T)$ selections, contradicting the $\Theta(T)$ selections in the last half of the horizon.
Hence persistent misranking can weaken the guaranteed allocation from near-full to square-root order, while the optimal branch still receives an unbounded number of visits under these root conditions.
\end{assumption}

\begin{assumption}[Finite-time BO regret]
\label{ass:bo_regret}
Conditioned on $m$ BO evaluations and $q$ evaluated configurations proposed by the LLM in the optimal branch, denoted by $\mathcal{D}^{\mathrm{LLM}}_q$, the EI proposer is warm-started with this dataset and satisfies the branch-wise simple-regret rate of expected improvement~\citeapp{bull2011convergence}:
\[
    S^{\mathrm{BO}\mid\mathrm{LLM}}_m
    \le
    C_{\mathrm{EI}}(\mathcal{D}^{\mathrm{LLM}}_q) r_{\mathrm{EI}}(m),
\]
where $C_{\mathrm{EI}}(\mathcal{D}^{\mathrm{LLM}}_q)>0$ is a data-dependent constant induced by the warm-start observations and
\begin{equation}
    r_{\mathrm{EI}}(m)
    =
    \begin{cases}
    m^{-\nu/d_*}(\log m)^{\eta}, & 0<\nu \le 1,\\
    m^{-1/d_*}, & \nu>1.
    \end{cases}
    \label{eq:ei_rate}
\end{equation}
Here $d_*$ is the dimension of the optimal branch, $\nu$ is the RKHS smoothness parameter of the GP kernel, and $\eta\ge 0$ denotes the logarithmic factor in the EI convergence rate.
High-quality LLM proposals can reduce the finite-time constant by lowering the initial simple regret and providing informative posterior observations, while the EI exponent $r_{\mathrm{EI}}(m)$ remains unchanged.
\end{assumption}

\begin{assumption}[LLM-assisted proposal quality]
\label{ass:llm_quality}
Motivated by empirical in-context scaling laws for LLMs~\citeapp{anil2024many,liu2024llms}, we model the LLM-assisted regret after $n$ observations in the optimal branch as a power-law decay with an irreducible floor:
\begin{equation}
    S^{\mathrm{LLM}}_n
    \le
    \frac{C_{\mathrm{LLM}}}{n^\alpha} + \Delta_{\inf},
    \label{eq:llm_finite_bound}
\end{equation}
where $C_{\mathrm{LLM}}>0$, $\alpha>0$, and $\Delta_{\inf}\ge 0$ is the irreducible floor of the LLM proposer.
This assumption formalizes the finite-time benefit of semantic priors.
\end{assumption}

\begin{lemma}[BO evaluations in the optimal branch]
\label{lem:bo_count}
Suppose that whenever $A^*$ is selected, the BO proposer is selected with conditional probability at least $\epsilon>0$.
Then
\[
    \Pr\!\left(M_*(T) \ge \frac{\epsilon N_*(T)}{2} \mid N_*(T)\right)
    \ge
    1-\exp\!\left(-\frac{\epsilon N_*(T)}{8}\right).
\]
\end{lemma}

\begin{proof}
Index the visits to $A^*$ by $j=1,\dots,N_*(T)$.
The proposer-selection variable can be written as
$Z_j=\mathbb{I}\{U_j\le p_j\}$, where $U_j$ are independent uniform random variables and the adaptive probability $p_j$ satisfies $p_j\ge\epsilon$.
Thus $Z_j$ stochastically dominates $\mathbb{I}\{U_j\le\epsilon\}$, and $M_*(T)$ dominates a binomial random variable with parameters $N_*(T)$ and $\epsilon$.
The multiplicative Chernoff bound gives
\[
    \Pr\!\left(M_*(T) < \frac{\epsilon N_*(T)}{2}\right)
    \le
    \exp\!\left(-\frac{\epsilon N_*(T)}{8}\right),
\]
which proves the claim.
\end{proof}

\begin{theorem}[Finite-time global simple regret]
\label{thm:finite_time_regret}
Under Assumptions~\ref{ass:root_allocation}--\ref{ass:llm_quality}, with probability at least
\[
    1-\exp\!\left(-\frac{\epsilon N_*(T)}{8}\right),
\]
the global simple regret of \sys satisfies
\begin{equation}
    S_T^{\mathrm{Global}}
    \le
    \min\left\{
    \frac{C_{\mathrm{LLM}}}{N_*(T)^\alpha}+\Delta_{\inf},
    \;
    C_{\mathrm{EI}}(\mathcal{D}^{\mathrm{LLM}}_q) r_{\mathrm{EI}}(M_*(T))
    \right\}.
    \label{eq:finite_time_global_regret}
\end{equation}
\end{theorem}

\begin{proof}
By Assumption~\ref{ass:root_allocation}, $N_*(T)=\Omega(T)$.
Conditioned on this allocation, Lemma~\ref{lem:bo_count} implies that the event
$\mathcal{E}_{\mathrm{count}}=\{M_*(T)\ge \epsilon N_*(T)/2\}$ holds with probability at least
$1-\exp(-\epsilon N_*(T)/8)$.

On $\mathcal{E}_{\mathrm{count}}$, Assumption~\ref{ass:bo_regret} gives
\[
    S^{\mathrm{BO}\mid\mathrm{LLM}}_{M_*(T)}
    \le
    C_{\mathrm{EI}}(\mathcal{D}^{\mathrm{LLM}}_q) r_{\mathrm{EI}}(M_*(T)).
\]
Assumption~\ref{ass:llm_quality} gives
\[
    S^{\mathrm{LLM}}_{N_*(T)}
    \le
    \frac{C_{\mathrm{LLM}}}{N_*(T)^\alpha}+\Delta_{\inf}.
\]

The best configuration returned by \sys is the best point observed over the entire global history.
Therefore, its global simple regret is no larger than the regret achieved by either the BO-generated points or the LLM-assisted points inside the optimal branch:
\[
    S_T^{\mathrm{Global}}
    \le
    \min\left\{S^{\mathrm{LLM}}_{N_*(T)}, S^{\mathrm{BO}\mid\mathrm{LLM}}_{M_*(T)}\right\}.
\]
Combining the two branch-wise bounds with the root-allocation and proposer-count events yields Eq.~\eqref{eq:finite_time_global_regret}.
\end{proof}

\begin{proof}[Proof of Theorem~\ref{thm:convergence}]
By Assumption~\ref{ass:root_allocation}, $N_*(T)=\Omega(T)$ as $T\to\infty$.
Since $N_*(T)=\Omega(T)$, the failure probabilities $\exp(-\epsilon N_*(T)/8)$ are summable.
By Lemma~\ref{lem:bo_count} and the Borel--Cantelli lemma, the event $M_*(T)\ge \epsilon N_*(T)/2$ fails only finitely often almost surely.
Hence $M_*(T)\to\infty$ almost surely.
The EI rate satisfies $r_{\mathrm{EI}}(M_*(T))\to 0$.
Assumption~\ref{ass:bo_regret} gives the warm-started BO bound
\[
    S^{\mathrm{BO}\mid\mathrm{LLM}}_{M_*(T)}
    \le
    C_{\mathrm{EI}}(\mathcal{D}^{\mathrm{LLM}}_q) r_{\mathrm{EI}}(M_*(T)).
\]
The warm-start set affects only the data-dependent constant; along the optimization trajectory this constant is finite and remains bounded, so the BO term converges to zero.
Therefore, the BO term in Eq.~\eqref{eq:finite_time_global_regret} converges to zero almost surely, and because the global simple regret is upper-bounded by the minimum in Eq.~\eqref{eq:finite_time_global_regret}, the global simple regret also converges to zero almost surely.
\end{proof}

\begin{corollary}[Finite-time budgets under linear root allocation]
\label{cor:finite_time_acceleration}
Under Assumptions~\ref{ass:bo_regret} and~\ref{ass:llm_quality}, consider the linear-allocation regime $N_*(T)=\Omega(T)$ for both \sys and the BO-only B-MCTS baseline, on their respective root-allocation events.
Under Assumption~\ref{ass:root_allocation}, controlled violations with $B_\Sigma(T,\rho)=\mathcal O(\sqrt T)$ provide such a regime; the allocation condition must hold for each method separately.
Assume the BO regret constants admit fixed uniform bounds $C_{\mathrm{EI}}$ for the baseline and $C_{\mathrm{EI}}^{\mathrm{LLM}}$ for the interleaved search, where $C_{\mathrm{EI}}(\mathcal D_q^{\mathrm{LLM}})\le C_{\mathrm{EI}}^{\mathrm{LLM}}$ along the latter trajectory.
For the assumed EI rate in Eq.~\eqref{eq:ei_rate} with $\nu>1$,
\[
    r_{\mathrm{EI}}(m)=m^{-1/d_*}.
\]
For a target simple regret $\delta>0$, a sufficient total evaluation budget for the BO-only baseline is, up to logarithmic factors,
\[
    T_{\mathrm{Base}}(\delta)
    =
    \widetilde{\mathcal{O}}
    \left(
    \left(\frac{C_{\mathrm{EI}}}{\delta}\right)^{d_*}
    \right).
\]
Here the EI search in the optimal branch has dimension $d_*$, and linear root allocation converts its branch budget to a total budget of the same order.
By Eq.~\eqref{eq:finite_time_global_regret}, either the LLM term or the BO term falling below $\delta$ suffices to certify the target for \sys.
For $\delta>\Delta_{\inf}$, this gives
\[
    T_{\mathrm{LB}}(\delta)
    =
    \widetilde{\mathcal{O}}
    \left(
    \min\left\{
    \left(\frac{C_{\mathrm{LLM}}}{\delta-\Delta_{\inf}}\right)^{1/\alpha},
    \frac{1}{\epsilon}
    \left(\frac{C_{\mathrm{EI}}^{\mathrm{LLM}}}{\delta}\right)^{d_*}
    \right\}
    \right),
\]
where root-allocation constants and fixed confidence factors are suppressed.
These are sufficient high-probability budget bounds under the stated assumptions.
For the BO count, coupling the first deterministic $n$ branch visits with independent Bernoulli variables of mean $\epsilon$ bounds the joint event $\{M_*(T)<\epsilon n/2,\ N_*(T)\ge n\}$ by $\exp(-\epsilon n/8)$.
This failure probability and those of the root-allocation events are combined by a union bound.
When $\delta\le\Delta_{\inf}$, the LLM floor prevents the first term from certifying the target, but the BO term still gives
\[
    T_{\mathrm{LB}}(\delta)
    =
    \widetilde{\mathcal{O}}
    \left(
    \frac{1}{\epsilon}
    \left(\frac{C_{\mathrm{EI}}^{\mathrm{LLM}}}{\delta}\right)^{d_*}
    \right).
\]
For an algebraic comparison, define the nominal ratio of the displayed leading budget expressions, setting aside the suppressed allocation and confidence factors:
\begin{equation}
    R_{\mathrm{budget}}(\delta)
    :=
    \begin{cases}
    \displaystyle
    \max\left\{
    \left(\frac{C_{\mathrm{EI}}}{\delta}\right)^{d_*}
    \left(\frac{\delta-\Delta_{\inf}}{C_{\mathrm{LLM}}}\right)^{1/\alpha},
    \;
    \epsilon
    \left(
    \frac{C_{\mathrm{EI}}}{C_{\mathrm{EI}}^{\mathrm{LLM}}}
    \right)^{d_*}
    \right\},
    & \delta>\Delta_{\inf},\\[1.2em]
    \displaystyle
    \epsilon
    \left(
    \frac{C_{\mathrm{EI}}}{C_{\mathrm{EI}}^{\mathrm{LLM}}}
    \right)^{d_*},
    & \delta\le\Delta_{\inf}.
    \end{cases}
    \label{eq:acceleration_ratio}
\end{equation}
This ratio compares budget expressions; it is not a bound on the ratio of actual hitting times, which cannot be inferred by dividing two complexity upper bounds.
Informative LLM observations may lower the BO regret constant, although $C_{\mathrm{EI}}^{\mathrm{LLM}}<C_{\mathrm{EI}}$ is not assumed.
Under Assumption~\ref{ass:llm_quality}, directly evaluated LLM proposals also provide a separate route to the target through Eq.~\eqref{eq:llm_finite_bound}.
\end{corollary}

\paragraph{Other allocation regimes.}
The above total-budget orders apply whenever the corresponding linear-allocation conditions hold, including some cases with $B_\Sigma(T,\rho)=\Theta(T)$, such as $B_\Sigma(T,\rho)=0.2T$.
More generally, if both methods satisfy $N_*(T)=\Omega(T^\beta)$ for a common $0<\beta\le1$, converting their branch budgets to total budgets raises the leading expressions to the power $1/\beta$.
With allocation constants, logarithmic factors, and confidence factors suppressed as above, the corresponding nominal budget ratio is $[R_{\mathrm{budget}}(\delta)]^{1/\beta}$.
Square-root allocation corresponds to $\beta=1/2$, giving the nominal ratio $[R_{\mathrm{budget}}(\delta)]^2$.
Thus, whenever $R_{\mathrm{budget}}(\delta)>1$, a favorable nominal budget comparison is preserved across these allocation regimes.

\paragraph{Theoretical-Empirical Alignment.}
Empirically, \sys achieves an average $2.37\times$ acceleration over B-MCTS (which retains only the BO proposer) across 10 datasets in the BO--LLM collaboration ablation (\cref{fig:validation_rank_bollm}).
This is an observed speedup, distinct from the nominal budget comparison above.

\subsection{A Case of MCTS Result}
\label{app:visualization_of_search_process}

\begin{figure*}[p]  %
    \centering
    \includegraphics[width=0.93\textwidth]{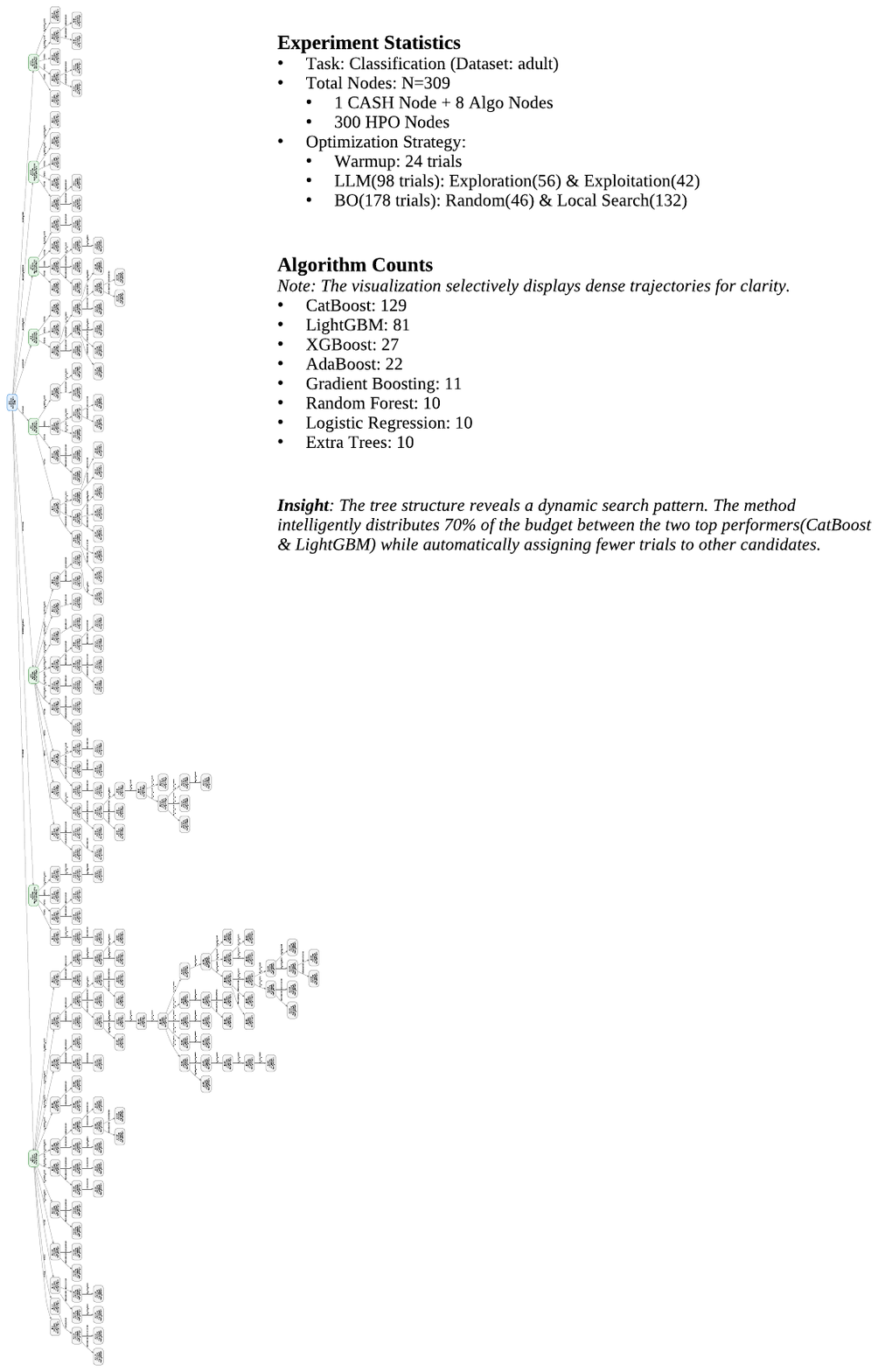}
    \caption{MCTS tree generated by LB-MCTS on the Adult dataset.}
    \label{fig:adult-mcts-tree}
\end{figure*}

While \cref{fig:lb-mcts_structure} in the main paper shows an abstract schematic of the MCTS search structure, \cref{fig:adult-mcts-tree} presents a concrete example of an actual search tree generated by \sys on the ``Adult'' dataset. The visualization reveals how BO and the LLM proposer collaboratively expand the tree: The first leaf under each Algo node serves as the initial configuration for that path, established via either an LLM warm-start or BO random search. Along a typical optimization path, BO-driven local search and LLM-driven exploratory/exploitative proposals alternate in refining configurations.

\clearpage  %

\section{Prompt Design and Case Studies}

\subsection{Prompt Design For Hyperparameter Tuning Phase}
\label{app:prompt_tuning}

The hyperparameter tuning process follows a dual-prompt mechanism: a static \textbf{System Prompt} defining the agent's persona, and a dynamic \textbf{User Prompt} supplying the optimization context. The User Prompt is constructed using a template comprising four key components: (i) task description, 
(ii) dataset context, (iii) selective tuning memory, and (iv) output requirements. Crucially, the strategic directive within the task description is instantiated at runtime based on the MCTS state.
\begin{PromptBox}{System Prompt: }
    \begin{promptsection}{promptGray}
         
        You are an expert AutoML strategist. Your current role is \textbf{Hyperparameter Tuner} in a Combined Algorithm Selection and Hyperparameter optimization (CASH) task.  \\
        An algorithm has been selected. Your goal is to adjust a basic hyperparameter configuration on a monte carlo tree search node to find the best hyperparameter configuration for it. 
    \end{promptsection}
\end{PromptBox}
\begin{PromptBox}{Template: User Prompt For Hyperparameter Tuning Phase}
    \begin{promptsection}{taskblue}
        {\large \textbf{Task Description}}: \\
        Your ultimate goal is to solve the Combined Algorithm Selection and Hyperparameter optimization (CASH) problem: finding the best possible algorithm and its optimal hyperparameters for a classification task on the provided dataset. \\
        \textbf{Algorithm `\texttt{[ALGORITHM]}' has been selected}. Your immediate objective is to propose a hyperparameter configuration for it.
        
        \smallskip
        The proposed configuration must strictly adhere to the defined search space:
        \begin{itemize}[leftmargin=15pt, nosep]
            \footnotesize \ttfamily
            \item parameter\_1: \texttt{[TYPE]}, \texttt{[RANGE/CHOICES]}
            \item parameter\_2: \texttt{[TYPE]}, \texttt{[RANGE/CHOICES]}
            \item \dots
        \end{itemize}
        
        \smallskip
        \textbf{The Basic Configuration to Build Upon}: \\
        This configuration is derived from previous trials and represents a promising starting point: \\
        {\footnotesize \ttfamily \{ "parameter\_1": \texttt{[VALUE]}, "parameter\_2": \texttt{[VALUE]}, \dots \}}
        
        \smallskip
        Your current strategic goal is \textbf{\texttt{[MODE]}}. \textbf{\texttt{[INSTRUCTION\_TEXT]}} \\
        \textit{(Note: This directive is dynamically instantiated based on the MCTS state. See "Strategy Variants" below.)}
    \end{promptsection}

    \vspace{4pt}

    \begin{promptsection}{taskblue}
        {\large \textbf{Dataset Context}}: \\
        The target task is a \textbf{\texttt{[TASK\_TYPE]}} problem evaluated on the provided dataset. 
        The primary optimization metric is \textbf{\texttt{[METRIC]}}.
        The dataset consists of \textbf{\texttt{[N\_SAMPLES]}} samples and \textbf{\texttt{[N\_FEATURES]}} features...
    \end{promptsection}

    \vspace{4pt}

    \begin{promptsection}{taskblue}
        {\large \textbf{Selective Tuning Memory}}: \\
        This component integrates historical trajectories and high-performing trials to inform the current optimization step.
        
        \vspace{4pt}
        \textbf{Optimization Trajectory Context}: \\
        \texttt{[Sequence: Root $\to$ ... $\to$ Ancestral Node $\to$ Current Basic Node]}
        
        \vspace{4pt}
        \textbf{Informative Historical Trials}: \\
        \texttt{[List of Top-k High-Performing \& Similar Historical Trials]}
    \end{promptsection}

    \vspace{4pt}

    \begin{promptsection}{taskblue}
        {\large \textbf{Required Output Format}}: \\
        You must provide your reasoning followed by the final configuration in the following structure.
        
        \smallskip
        \textbf{Thought}: ... \\
        \textbf{Action}: {\footnotesize \ttfamily \{ "parameter\_1": \texttt{[VALUE]}, \dots \}}
    \end{promptsection}
\end{PromptBox}
\begin{PromptBox}{Optimization Directives : Strategy Variants}
    The placeholders \textbf{\texttt{[MODE]}} and \textbf{\texttt{[INSTRUCTION\_TEXT]}} in the template above are dynamically instantiated based on the MCTS search state.

    \vspace{8pt}

    \begin{strategysection}{warmuporange}{warmupborder}
        \textbf{Mode: WARMUP} \\
        \textbf{Instruction}: Propose a promising configuration \textbf{from scratch}. Focus on identifying configurations that are \textbf{distinct from previous attempts} while still demonstrating strong performance. Analyze historical data to \textbf{balance novelty and effectiveness}.
    \end{strategysection}

    \vspace{6pt}

    \begin{strategysection}{exploreyellow}{exploreborder}
        \textbf{Mode: EXPLORATION} \\
        \textbf{Instruction}: Modify the basic configuration to \textbf{explore new areas} of the hyperparameter space. Focus on making changes that \textbf{introduce diversity} and help uncover potentially high-performing configurations. Make \textbf{bold adjustments} to discover new possibilities.
    \end{strategysection}

    \vspace{6pt}

    \begin{strategysection}{exploityellow}{exploitborder}
        \textbf{Mode: EXPLOITATION} \\
        \textbf{Instruction}: \textbf{Refine and optimize} the basic configuration to achieve peak performance. Focus on making \textbf{minor, incremental adjustments} to the basic configuration, guiding it towards a more optimal setup using insights from successful trials.
    \end{strategysection}
\end{PromptBox}

\subsection{Prompt Design For Reflection Phase}
\label{app:prompt_reflection}

Similar to the tuning phase, the reflection phase is governed by a dedicated System Prompt that establishes the `Performance Analyst' persona. The User Prompt is structured to feed execution results back to the LLM, enabling it to perform quantitative attribution analysis and synthesize reusable insights.

\begin{PromptBox}{System Prompt}
    \begin{promptsection}{promptGray}
        You are an expert AutoML strategist. Your current role is \textbf{Performance Analyst} in a Combined Algorithm Selection and Hyperparameter optimization (CASH) task. \\
        Your goal is to analyze the results of a hyperparameter tuning trial. You must provide a quantitative, structured, and globally useful reflection that explains the relationship between parameter changes and performance outcomes to guide future trials.
    \end{promptsection}
\end{PromptBox}

\vspace{10pt}

\begin{PromptBox}{Template: User Prompt For Reflection Phase}

    \begin{promptsection}{taskblue}
        {\large \textbf{Task Description}}: \\
        Your ultimate goal is to solve the Combined Algorithm Selection and Hyperparameter optimization(CASH) problem: finding the best possible algorithm and its optimal hyperparameters for a classification task on the provided dataset. \\
        A hyperparameter configuration for algorithm \textbf{`\texttt{[ALGORITHM]}' } has just been evaluated. Your immediate objective is to \textbf{Reflect on Trial Performance}...
        
        \smallskip
        \textbf{The Trial to Analyze}: \\
        This section details the configuration changes and the resulting performance shift relative to the basic configuration:
        
        \begin{description}[leftmargin=10pt, nosep]

            \item[Basic Configuration] {\ttfamily \{ "parameter": \texttt{[VALUE]}, \dots \}}
            \item[Basic Performance] \texttt{[BASIC\_PERF]}
            \item[Original Rationale] \texttt{[AGENT\_THOUGHT\_FROM\_TUNING\_PHASE]}
            \item[Resulting Configuration] {\ttfamily \{ "parameter": \texttt{[VALUE]}, \dots \}}
            \item[Resulting Performance] \texttt{[NEW\_PERF]} (\textbf{\texttt{[CHANGE\_\%]}} relative to basic)
            \item[Global Ranking] \textbf{\texttt{[RANK]} out of \texttt{[TOTAL\_TRIALS]}} trials
        \end{description}
        
        \smallskip
        \textbf{Analytical Reflection Requirements}: \\
        Your analysis must be ``quantitative, structured, and globally useful":
        \begin{itemize}[leftmargin=12pt, nosep]
            \item In Reflection:
                \begin{enumerate}[leftmargin=10pt, nosep]
                    \item Triage \& Compare: Position this trial among historical data...
                    \item Attribute \& Synthesize: Explain which parameter deviation led to the outcome...
                \end{enumerate}
            \item In Reflection Summary: Distill into a high-confidence summary of global lessons. (Only this summary will be stored in memory).
        \end{itemize}
    \end{promptsection}

    \vspace{4pt}

    \begin{promptsection}{taskblue}
        {\large \textbf{Dataset Context}}: \\
        The target task is a \textbf{\texttt{[TASK\_TYPE]}} problem evaluated on the provided dataset. 
        The primary optimization metric is \textbf{\texttt{[METRIC]}}.
        The dataset consists of \textbf{\texttt{[N\_SAMPLES]}} samples and \textbf{\texttt{[N\_FEATURES]}} features...
    \end{promptsection}

    \vspace{4pt}

    \begin{promptsection}{taskblue}
        {\large \textbf{Selective Tuning Memory}}: \\
        \texttt{[Identical to the Tuning Phase Input: Trajectory Context + Historical Trials]}
    \end{promptsection}

    \vspace{4pt}

    \begin{promptsection}{taskblue}
        {\large \textbf{Required Output Format}}: \\
        
        \textbf{Reflection}: \\
        (Your structured analysis addressing the two points above...)

        \textbf{Reflection Summary}: \\
        (A compact, single paragraph summary of global, reusable and high-confidence conclusions...)
    \end{promptsection}
\end{PromptBox}

\subsection{Reflection Design for BO and LLM \textsc{Warmup}}
\label{app:reflection_for_bo_and_warmup}
While the LLM proposer follows each recommendation with a dedicated reflection stage, the BO proposer lacks an inherent mechanism for semantic reasoning. Similarly, LLM \textsc{Warmup} trials are excluded from LLM-based reflection because the sparse early-stage history provide insufficient observations and lack a reference ``basic configuration" for meaningful comparison. To bridge this gap and maintain a consistent search history, we employ fixed programmatic rules to automatically summarize these trials.

\textbf{Reflection of BO \textsc{Local Search}}. 
The BO \textsc{Local Search} generates candidates by sampling from the neighborhood of a basic configuration via small perturbations. Since this process is inherently incremental, its reflection is designed to explicitly capture these fine-grained modifications by concatenating two programmatic components:

\begin{enumerate}[topsep=-1pt, partopsep=0pt, itemsep=3pt, parsep=0pt]
    \item \textbf{Parameter Change Description}: Lists all hyperparameters modified from the parent configuration, formatted as ``Changed parameters: \texttt{\{param\_name\}: \{old\} -> \{new\}}''.
    \item \textbf{Performance Change Description}: Quantifies the performance shift using the computed delta, formatted as ``Performance \{improved|declined\} from \texttt{\{parent\_perf\}} to \texttt{\{current\_perf\}}''.
\end{enumerate}

Example Case: Consider a \textsc{Local Search} trial on the \texttt{cmc} classification dataset. The optimization transition is defined by the following parent and child nodes:

\begin{itemize}[topsep=-1pt, partopsep=0pt, itemsep=3pt, parsep=0pt]
    \item Parent Node:
    \begin{itemize}[topsep=-1pt, partopsep=0pt, itemsep=3pt, parsep=0pt]
        \item Performance: \texttt{0.5846677}
        \item Configuration: 
        \texttt{\{"algorithm": "SAMME", "learning\_rate": 4.049957e-02, "max\_depth": 5, "n\_estimators": 217\}}
    \end{itemize}
    
    \item Child Node:
    \begin{itemize}[topsep=-1pt, partopsep=0pt, itemsep=3pt, parsep=0pt]
        \item Performance: \texttt{0.59120365}
        \item Configuration: 
        \texttt{\{"algorithm": "SAMME", "learning\_rate": 4.000687e-02, "max\_depth": 5, "n\_estimators": 231\}}
    \end{itemize}
\end{itemize}

Based on this transition, the system automatically generates the following textual reflection, which serves as part of the context for future LLM iterations:

\begin{ReflectionBox}[System Generated Reflection]
\texttt{"Bayesian Local search performed; Changed parameters: learning\_rate: 4.049957e-02 -> 4.000687e-02, n\_estimators: 217 -> 231. Performance improved from 0.5846677 to 0.59120365."}
\end{ReflectionBox}

\vspace{1em}
\textbf{Special Cases: BO \textsc{Random Search} and LLM \textsc{Warmup}:}

For \textsc{Warmup} trials and BO \textsc{Random Search} trials (where the parent node is an \texttt{HPORootNode}), the reflection string is set to ``Warmup configuration.''
Moreover, instead of a performance change description, the system appends ``Initial performance: \texttt{\{current\_perf\}}'' to indicate this is the first trial of an optimization trajectory.

Notably, if \textsc{Warmup} or BO \textsc{Random Search} trials achieve exceptionally high performance and are selected by the Pareto frontier, they may appear in the \texttt{Informative Historical Trials} section of the prompt. In such cases, only the Child configuration and Reflection are displayed (no Parent), since the parent is an \texttt{HPORootNode}.

\subsection{Prompt Example: Execution Trace}
\label{app:example}
This section presents a concrete execution trace from the \texttt{cmc} classification dataset experiment.It displays the raw input prompts and the corresponding model outputs during a \textbf{Hyperparameter Tuning} step and its subsequent \textbf{Reflection} step.

\begin{PromptBox}[taskblue]{User Prompt For Hyperparameter Tuning}
        {\large \textbf{Task Description}}: \\      
        Your ultimate goal is to solve the Combined Algorithm Selection and Hyperparameter optimization(CASH) problem: finding the best  possible algorithm and its optimal hyperparameters for a classification task on the provided dataset. \\
        \textbf{Algorithm ``adaboost'' has been selected}. Your immediate goal is to propose a hyperparameter configuration for it. \\
        \smallskip
        The proposed configuration must strictly adhere to the defined search space:
        \begin{itemize}[leftmargin=15pt, nosep, font=\ttfamily \scriptsize]
            \item algotirhm: Categorical, \{SAMME.R, SAMME\}
            \item learning\_rate: Float, Range=[1e-02, 2e+00] (Log-Scale)
            \item max\_depth: Integer, Range=[2, 8]
            \item n\_estimators: Integer, Range=[50,500]
        \end{itemize}
        \vspace{1em}
        \smallskip
        \textbf{The Basic Configuration to Build Upon}: \\
        This configuration is derived from previous trials and represents a promising starting point: \\
        {\ttfamily \{"algorithm": "SAMME", "learning\_rate": 4.5e-02, "max\_depth": 5, "n\_estimators": 350\}}
        
        \smallskip
        Your current strategic goal is \textbf{Exploitation}. Refine and optimize the basic configuration to achieve peak performance. Focus on making minor, incremental adjustments to the basic configuration, guiding it towards a more optimal setup using insights from successful trials. \\

    \vspace{1em} %

        {\large \textbf{Dataset Context}}: \\
        The target task is a \textbf{3-class classification} problem evaluated on the provided dataset. 
        The primary optimization metric is \textbf{balanced accuracy}, where a higher value indicates superior performance.   
        The dataset consists of \textbf{1,178 samples} and \textbf{9 total features}, partitioned into 9 numeric and 0 categorical features. 
        The target classes are distributed as \textbf{42.7\%} for Class 1, \textbf{34.7\%} for Class 3, \textbf{22.6\%} for Class 2.

    \vspace{1em}

        {\large \textbf{Selective Tuning Memory}}:  \\
         This component integrates historical trajectories and high-performing trials to inform the current optimization step.

        \vspace{0.5em} 

        \begin{adjustwidth}{1em}{0pt}
         \textbf{Optimization Trajectory Context}:  \\
        This sequence preserves the ancestral path from the algorithm root, providing trajectory awareness:
        \end{adjustwidth}
        
        \begin{adjustwidth}{2em}{0pt}
        \begin{description}[leftmargin=2em, nosep]
            \item[\quad \dots] 
            \item[Ancestral Node] {\ttfamily \{ "performance": 0.58031504, "configuration": \{"algorithm": "SAMME", "learning\_rate": 3.9e-02, "max\_depth": 5, "n\_estimators": 265\}, "reflection": "..."  \}}
            \item[Current Basic Node] {\ttfamily \{ "performance": 0.57471745, "configuration": \{"algorithm": "SAMME", "learning\_rate": 4.5e-02, "max\_depth": 5, "n\_estimators": 350\}, "reflection": "Performance degradation (-0.97\%) indicates overfitting caused by simultaneously increasing learning\_rate (0.039->0.045) and n\_estimators (265->350). Future tuning must constrain the learning rate  when scaling up model complexity."\}}
        \end{description}

        \end{adjustwidth}

        \vspace{0.5em} 
        \begin{adjustwidth}{1em}{0pt}
        \textbf{Informative Historical Trials}:  \\
         The following trials represent top-performing historical attempts that are most similar to the current configuration:
        \end{adjustwidth}
        
        \smallskip
        \begin{adjustwidth}{2em}{0pt}
            
            {\textit{\textbf{Historical Trial A(Proposed by BO Local Search):}}}
            \begin{itemize}[leftmargin=12pt, nosep]
                \raggedright    
                
                \item \textbf{Parent}: {\ttfamily \{ "performance": 0.5846677, "configuration": \{"algorithm": "SAMME", "learning\_rate": 4.049957e-02, "max\_depth": 5, "n\_estimators": 217\} \}}
                
                \item \textbf{Child}: {\ttfamily \{ "performance": 0.59120365, "configuration": \{"algorithm": "SAMME", "learning\_rate": 4.000687e-02, "max\_depth": 5, "n\_estimators": 231\} \}}

                \item \textbf{Reflection}: {\ttfamily "Decreasing learning\_rate (0.04049957->0.04000687) and increasing n\_estimators (217->231) improved performance by 0.0065 (+1.12\%)." }
        
            \end{itemize}

        {\textit{\textbf{Historical Trial B(Proposed by LLM):}}}
        \begin{itemize}[leftmargin=12pt, nosep]
            \raggedright    
            
            \item \textbf{Parent}: {\ttfamily \{ "performance": 0.58264467, "configuration": \{"algorithm": "SAMME", "learning\_rate": 4.1953e-02, "max\_depth": 5, "n\_estimators": 233\} \}}
            
            \item \textbf{Child}: {\ttfamily \{ "performance": 0.58918061, "configuration": \{"algorithm": "SAMME", "learning\_rate": 3.8e-02, "max\_depth": 5, "n\_estimators": 280\} \}}
    
            \item \textbf{Reflection}:  {\ttfamily "Reducing the learning rate to 0.038 successfully mitigated oscillation, creating the stability required to profitably scale n\_estimators from 233 to 280. This confirms that higher model capacity demands a stricter learning rate constraint." }
        \end{itemize}
        
        \end{adjustwidth}
        
        \smallskip
        \dots

    \vspace{1em}

        {\large \textbf{Required Output Format}}: \\
        You must provide your reasoning followed by the final configuration in the following structure:

        \smallskip
        \textbf{Thought}: \\
        (Provide a detailed rationale. Analyze the \textbf{Optimization Trajectory Context} and the \textbf{Informative Historical Attempts} and connect your insights with the \textbf{Dataset Summary} and the current \textbf{Directive} to justify your parameter adjustments.)

        \smallskip
        \textbf{Action}: \\
        Output the configuration as a single, valid JSON block.\\    
        \smallskip
        \texttt{\{ "parameter\_1": value\_1, "parameter\_2": value\_2, ... \}}
\end{PromptBox}

\begin{ResponseBox}[ASSISTANT Response (Tuning)]
    \textbf{Thought}: \\
    The current directive is \textbf{EXPLOITATION}, which requires refining the basic configuration based on historical insights.
    
    Analyzing the \textit{Optimization Trajectory Context}, the current basic node (LR 0.045, Est 350) suffered a performance drop due to overfitting. The reflection explicitly warns that the learning rate must be constrained when scaling up estimators.
    
    Looking at \textit{Informative Historical Trials}, specifically "Historical Trial B", we see a clear success pattern where a lower learning rate of 0.038 enabled the effective scaling of estimators to 280. This aligns perfectly with the trajectory's lesson.
    
    Therefore, to exploit this finding and optimize performance:
    1. I will revert the \texttt{learning\_rate} to the proven stable value of \textbf{0.038}.
    2. I will set \texttt{n\_estimators} to \textbf{300}. This is a precise adjustment that sits between the under-fitted 265 and the over-fitted 350, slightly pushing the boundary of the successful Trial B (280) to maximize model capacity within the safe learning rate regime.
    3. I will maintain the \texttt{algorithm} as "SAMME" and \texttt{max\_depth} at 5, as these are established optima.

    \smallskip
    \textbf{Action}:
    \begin{verbatim}
{   
    "algorithm": "SAMME",
    "learning_rate": 0.038,
    "max_depth": 5,
    "n_estimators": 300
}
    \end{verbatim}
\end{ResponseBox}

\vspace{2em}

\begin{PromptBox}[taskblue]{User Prompt For Reflection}

        {\large \textbf{Task Description}}: \\
        Your ultimate goal is to solve the Combined Algorithm Selection and Hyperparameter optimization(CASH) problem: finding the best possible algorithm and its optimal hyperparameters for a classification task on the provided dataset.\\
         A hyperparameter configuration for algorithm \textbf{``adaboost''} has just been evaluated. Your immediate objective is to Reflect on Trial Performance. You must analyze the results of the recent trial to synthesize quantitative, structured, and reusable lessons for future tuning.

        \vspace{1em}
        \smallskip
        {\large \textbf{The Trial to Analyze}}: \\
        This section details the configuration changes and the resulting performance shift relative to the basic configuration:

        \begin{description}[leftmargin=4em, nosep, labelindent=2em]
            \item[Basic Configuration] {\ttfamily \{"algorithm": "SAMME", "learning\_rate": 4.5e-02, "max\_depth": 5, "n\_estimators": 350\}}
            \item[Basic Performance] \texttt{0.57471745}
            \item[Original Rationale] ``The current directive is EXPLOITATION... [omitted detailed analysis] ... "
            \item[Resulting Configuration] {\ttfamily \{"algorithm": "SAMME", "learning\_rate": 3.8e-02, "max\_depth": 5, "n\_estimators": 300\}}
            \item[Resulting Performance] \texttt{0.59244859} (improved by 3.09\% relative to the basic configuration)
            \item[Global Ranking] 1 out of 219 trials of `adaboost'
        \end{description}

        \vspace{1em}
        \smallskip
        {\large \textbf{Analytical Reflection Requirements}}: \\
        Your analysis must be \textbf{``quantitative, structured, and globally useful"}:
        \begin{itemize}[leftmargin=12pt, nosep]
            \item \textbf{In Reflection}:
                \begin{enumerate}[leftmargin=10pt, nosep]
                    \item \textbf{Triage \& Compare}: Position this trial among historical data using numerical comparisons (absolute differences and percentages).
                    \item \textbf{Attribute \& Synthesize}: Explain which parameter deviation led to the outcome.
                        \begin{itemize}[leftmargin=8pt, nosep] 
                            \item Highlight \textbf{clearly important} parameters, \textbf{value ranges} that look promising/risky, and any \textbf{interactions} inferred from numbers.
                            \item Make reasoning \textbf{reusable} to guide future tuning (not just describing this single trial).
                            \item Only include conclusions you are \textbf{confident about} based on quantitative evidence.
                        \end{itemize}
                \end{enumerate}
            \item \textbf{In Reflection Summary}: Distill into a high-confidence summary of global lessons.
                \begin{itemize}[leftmargin=10pt, nosep]
                    \item Include only insights supported by multiple numerical observations.
                    \item Do \textbf{not} invent rules for noisy data; do \textbf{not} suggest explicit "next steps" or new configurations.
                \end{itemize}
        \end{itemize}

    \vspace{1em}

         {\large \textbf{Dataset Context}}: \\
        The target task is a \textbf{3-class classification} problem evaluated on the provided dataset. 
        The primary optimization metric is \textbf{balanced accuracy}, where a higher value indicates superior performance. 
        The dataset consists of \textbf{1,178 samples} and \textbf{9 total features}, partitioned into 9 numeric and 0 categorical features. 
        The target classes are distributed as \textbf{42.7\%} for Class `1', \textbf{34.7\%} for Class `3', \textbf{22.6\%} for Class `2'.

    \vspace{1em}

        {\large \textbf{Selective Tuning Memory}}:  \\
         \textit{... (For brevity, the Selective Tuning Memory content is omitted as it remains identical to the Tuning Phase input) ...}

    \vspace{1em}

        {\large \textbf{Required Output Format}}: \\
        \smallskip
        \textbf{Reflection}: \\
        (Your structured analysis addressing the two points above, focused on quantitative comparison and attribution of the outcome.)

        \smallskip
        \textbf{Reflection Summary}: \\
        (A compact, single paragraph summary of global, reusable and high-confidence conclusions derived from this analysis.)

\end{PromptBox}

\begin{ResponseBox}[ASSISTANT Response (Reflection)]
    \textbf{Reflection}: \\
    1. \textbf{Triage \& Compare}: The trial achieved a score of \textbf{0.5924}, improving by \textbf{3.09\%} over the basic configuration (0.5747). Ranking \textbf{1st out of 219}, this confirms the configuration is in the optimal region.
    
    2. \textbf{Attribute \& Synthesize}: The primary driver of success was reducing the \texttt{learning\_rate} from 0.045 to \textbf{0.038}. As hypothesized in the rationale, the previous failure at LR 0.045 was due to overfitting when combined with high estimators. By constraining the LR to 0.038, the model successfully accommodated the \texttt{n\_estimators} (300), validating the interaction pattern observed in Historical Trial B. The stability of \texttt{max\_depth=5} further supported this gain.
    
    \smallskip
    \textbf{Reflection Summary}: \\
    For 'adaboost' on the provided dataset, optimal performance is achieved by \textbf{constraining the learning rate to $\approx$ 0.038} when scaling \textbf{n\_estimators} (280-300). The interaction is critical: higher capacities require stricter learning rate constraints to prevent overfitting. \textbf{max\_depth} should be maintained at 5.
\end{ResponseBox}

\section{Results and Analysis}

\subsection{Search Space}
\label{app:search_space}

We evaluate \sys across both classification and regression tasks using a search space adapted from the VolcanoML AutoML system~\citeapp{mindware_li2023volcanoml}. Specifically, we refined the candidate pool by pruning less common models, ultimately retaining eight representative algorithms for each task type to ensure both diversity and efficiency (see Table~\ref{tab:search_space_for_algorithms}). These selections span major model families, including tree ensembles (e.g., Random Forest, Extra Trees), boosting frameworks (e.g., XGBoost, LightGBM), and linear models (e.g., Logistic Regression, Ridge).

For each algorithm, we construct a comprehensive hyperparameter search space involving a mix of continuous and categorical parameters. As shown in the table, the classification setting involves a total of 45 hyperparameters (37 continuous, 8 categorical) across the 8 algorithms, while the regression setting comprises 46 hyperparameters (38 continuous, 8 categorical). This formulation constitutes a high-dimensional, hierarchical CASH problem, where the optimizer must first select an algorithm and then navigate its specific conditional hyperparameter subspace.

\begin{table}[t!]
\caption{Search space for ML algorithms. We distinguish categorical (cat) hyperparameters from numerical (cont) ones.}
\label{tab:search_space_for_algorithms}

\centering
\begin{subtable}{0.45\linewidth}
    \centering
    \begin{tabular}{lcccc}
        \hline
        Type of Classifier & \#$\lambda$ & cat & cont \\
        \hline
        AdaBoost & 4 & 1  & 3  \\
        Random Forest & 5 & 2  & 3  \\
        Extra Trees & 5 & 2  & 3  \\
        Gradient Boosting & 7 & 1  & 6  \\
        Logistic Regression & 4 & 2  & 2  \\
        LightGBM & 7 & - & 7  \\
        Catboost & 4 & - & 4  \\
        Xgboost & 9 & - & 9  \\
        \hline
        Total (8 algos) & 45 & 8 & 37 \\
        \hline
    \end{tabular}
\end{subtable}
\begin{subtable}{0.45\linewidth}
    \centering
    \begin{tabular}{lcccc}
        \hline
        Type of Regressor & \#$\lambda$ & cat & cont  \\
        \hline
        AdaBoost & 4 & 1  & 3  \\
        Random Forest & 5 & 2  & 3  \\
        Extra Trees & 5 & 2  & 3  \\
        Gradient Boosting & 7 & 1  & 6  \\
        Ridge & 4 & 1  & 3  \\
        LightGBM & 7 & - & 7  \\
        Catboost & 5 & 1  & 4  \\
        Xgboost & 9 & - & 9  \\
        \hline
        Total (8 algos) & 46 & 8 & 38 \\
        \hline
    \end{tabular}
\end{subtable}
\end{table}

\begin{figure*}[tp]
    \centering
    \begin{minipage}[t]{0.4\linewidth}
        \vspace{0pt} %
        \centering
        \includegraphics[width=\linewidth]{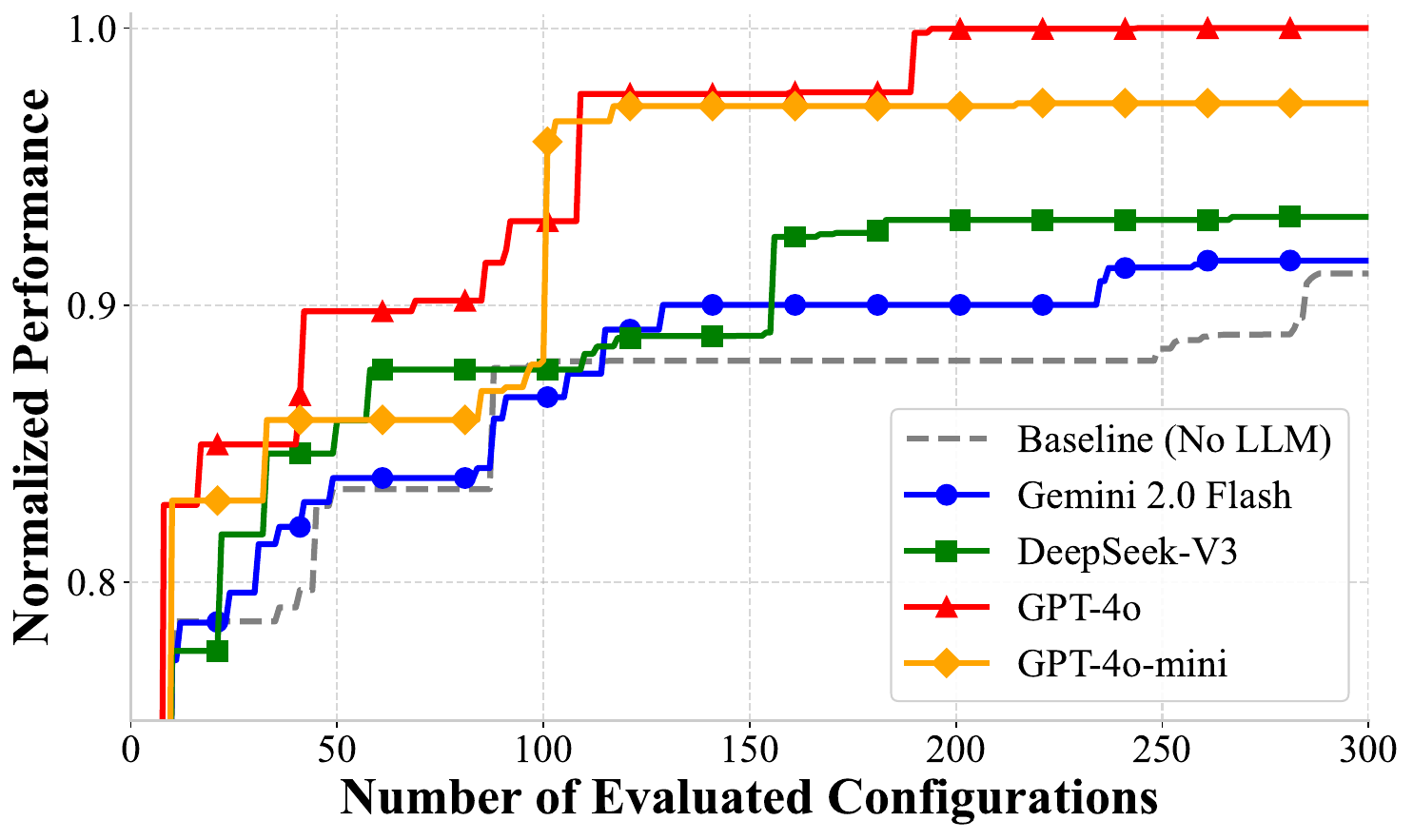}
        \caption{Average performance on 2 datasets.}
        \label{fig:model_ablation}
    \end{minipage}
    \hfill
    \begin{minipage}[t]{0.57\linewidth}
        \vspace{0pt} %
        \centering
        \captionof{table}{Ablation study of different LLM backbones on the \texttt{ada} (classification) and \texttt{quake} (regression) datasets. Cost is reported in USD. The best performance for each dataset is bolded, and the second-best results are underlined.}
        \label{tab:diff_llm}
        \resizebox{\linewidth}{!}{
            \begin{tabular}{l|cc|cc}
            \hline
            \multirow{2}{*}{\textbf{LLM Backbone}} & \multicolumn{2}{c|}{\textbf{\texttt{ada}}} & \multicolumn{2}{c}{\textbf{\texttt{quake}}} \\
             & \textbf{Bal Acc. $\uparrow$} & \textbf{Cost (\$)} & \textbf{MSE $\downarrow$} & \textbf{Cost (\$)} \\
            \hline
            No LLM (Pure BO) & 0.8086 & 0.00 & 0.03475 & 0.00 \\
            \hline
            Gemini 2.0 Flash & 0.8078 & 0.11 & 0.03462 & 0.06 \\
            GPT-4o-mini & \underline{0.8126} & 0.14 & \underline{0.03458} & 0.11 \\
            DeepSeek V3 & 0.8102 & 0.23 & 0.03472 & 0.14 \\
            GPT-4o & \textbf{0.8142} & 3.81 & \textbf{0.03449} & 2.85 \\
            \hline
            \end{tabular}
        }
    \end{minipage}
\end{figure*}

\subsection{Ablation on Different LLM Backbones}
\label{app:llm_backbone}

This appendix further studies how the choice of LLM backbone affects \sys.
We compare four LLMs (GPT‑4o~\citeapp{gpt-4o}, GPT‑4o‑mini~\cite{gpt-4o-mini}, DeepSeek‑V3~\citeapp{deepseekv3github}, and Gemini‑2.0‑Flash~\citeapp{gemini2flash}) and a No LLM variant that removes the LLM proposer and uses only the BO proposer. 
Due to the substantial API costs associated with premium models (for instance, GPT-4o can incur expenses of several USD per individual task), experiments are run on one classification task (ada) and one regression task (quake) to cover both metric types while maintaining a manageable budget.

\cref{fig:model_ablation} reports the normalized best‑so‑far validation performance versus the number of evaluations. The most capable (and most expensive) GPT‑4o consistently achieves the highest curve, with GPT‑4o‑mini close behind. Importantly, \emph{all} LLM‑based variants outperform the pure‑BO baseline, confirming that our integration framework can effectively exploit LLM guidance to complement Bayesian Optimization (BO) regardless of the specific backbone.

\cref{tab:diff_llm} summarizes the final performance and cost. 
GPT-4o represents the performance upper bound, achieving the best results on both datasets (0.8142 Balanced Accuracy and 0.03449 MSE). However, it is also the most expensive, costing significantly more than other backbones. Conversely, Gemini 2.0 Flash is the most affordable option but yields the lowest performance among the LLM variants.
GPT-4o-mini emerges as a highly cost-effective choice. It secures the second-best performance ranking while maintaining a low cost (e.g., $0.14$ for the ``ada'' task compared to GPT-4o’s $3.81$). Its performance-to-price ratio makes it a practical recommendation for budget-constrained large-scale CASH tasks.

To sum up, the results indicate that the ``ceiling'' of our proposed method is directly correlated with the reasoning capabilities of the underlying LLM. As the field of LLM research continues to evolve—leading to models with enhanced reasoning capabilities and reduced inference costs—the efficacy and accessibility of our framework are expected to scale proportionally, making it increasingly viable for a wider range of applications.

\subsection{Implementation  Details of Baselines}
\label{app:implementation_details}
We adhered to the open-source versions or the methodologies outlined in the original papers for all other baseline implementations.
For consistency, we implement the BO components of all baselines with Openbox 0.8.1~\cite{openbox_jiang2024openbox}, an open-source toolkit designed for black-box optimization.
We employ the same LLM backbone, GPT-4o-mini~\cite{gpt-4o-mini}, for all LLM-based optimizers to eliminate differences due to model strength.
Since prior LLM-based methods typically use only a few initial configurations (e.g., $\sim$5 in LLAMBO and BORA) on much simpler tasks, we standardize and strengthen warm-starting for our harder CASH setting.
Accordingly, to ensure fairness, all LLM-based baselines use the same number of warm-start configurations as our method (3 LLM-generated initial configurations per algorithm).
Meanwhile, pure BO-based baselines follow their standard initialization protocols, as they are natively designed for the CASH setting.
All the experiments are
conducted on a machine with 24 `AMD EPYC 7702P’ CPU cores and 80G memory.
\begin{itemize}[leftmargin=1.6em, topsep=0pt, partopsep=0pt, itemsep=3pt, parsep=0pt]
    \item \textbf{SMAC~\cite{smac_hutter2011sequential}.} We implement SMAC as a standard BO-based CASH optimizer using a random-forest surrogate. The optimizer treats CASH as a structured configuration space with conditional hyperparameters, and iteratively proposes one configuration per evaluation.
    \item \textbf{OptDivBO~\cite{optdivbo_poduval2024cash}.}
    OptDivBO enhances standard BO by encouraging diversity in the proposed models through ensemble selection. Following the original paper, we fix the diversity trade-off parameter $\tau$ at 0.2 and set the ensemble size for post-hoc selection to 25.
    \item \textbf{Rising Bandit(RB)~\cite{rb_li2020efficient}.}
    RB addresses the CASH problem by treating algorithm selection as a multi-armed bandit problem. 
    If the upper bound of the reward of an arm $k$ (algorithm $A^k$ ) is less than or equal to the lower bound of another arm in the candidate set, then the arm $k$ will be eliminated from the candidate set.
    The parameter $C$ for computing the smooth growth rate is set to 7.
    \item \textbf{MOSAIC~\cite{mosaic_rakotoarison2019automated}.}
    MOSAIC uses a tree policy for algorithm selection and BO for HPO. Algorithm selection is guided by a UCB-style criterion with exploration constant $C_{\text{ucb}}=1.3$, and tree expansion is controlled by progressive widening with coefficient $PW=0.6$. For the BO component, MOSAIC fits a \emph{single global} random-forest surrogate over the entire CASH space using all observed trials (rather than per-algorithm surrogates), and proposes configurations under the selected algorithm accordingly.
    \item \textbf{OPRO~\cite{opro_yang2023large}.}
    OPRO treats LLM as a standalone optimizer. In each iteration, we prompt the LLM with the task description and a truncated history containing only the top-20 best-performing trials found so far. The LLM is instructed to identify patterns from these high-quality examples and propose a single new configuration for evaluation.
    \item \textbf{LLAMBO~\cite{llambo_liularge}.}
    LLAMBO replaces traditional BO components with LLM. Following the official implementation, we include the entire optimization history in the prompt (subject to context limits). 
    In each round, the LLM is tasked to generate candidates that improve upon the current best score. 
    The method generates 20 candidate configurations per iteration. It then randomly shuffles the order of historical examples 10 times to aggregate the predictions.
    Consequently, each optimization round necessitates a total of 30 API calls.
    \item \textbf{BOPRO~\cite{bopro_agarwal2025searching}.}
    BOPRO uses BO to optimize the prompt instructions themselves. For the in-context learning component, the prompt includes a set of 8 historical examples. In each iteration, the BO-selected instruction template guides the LLM to reason over these examples and generate a new hyperparameter configuration.
    \item \textbf{BORA~\cite{bora_cisse2025language}.}
    BORA dynamically switches between an LLM and a GP-BO proposer based on posterior variance. When the LLM is active, we follow the original paper's setting by including the full optimization history in the prompt to maximize context awareness. 
\end{itemize}
\begin{figure*}[tb]
    \centering

    \begin{subfigure}[b]{0.34\linewidth} %
        \centering
        \includegraphics[width=\linewidth]{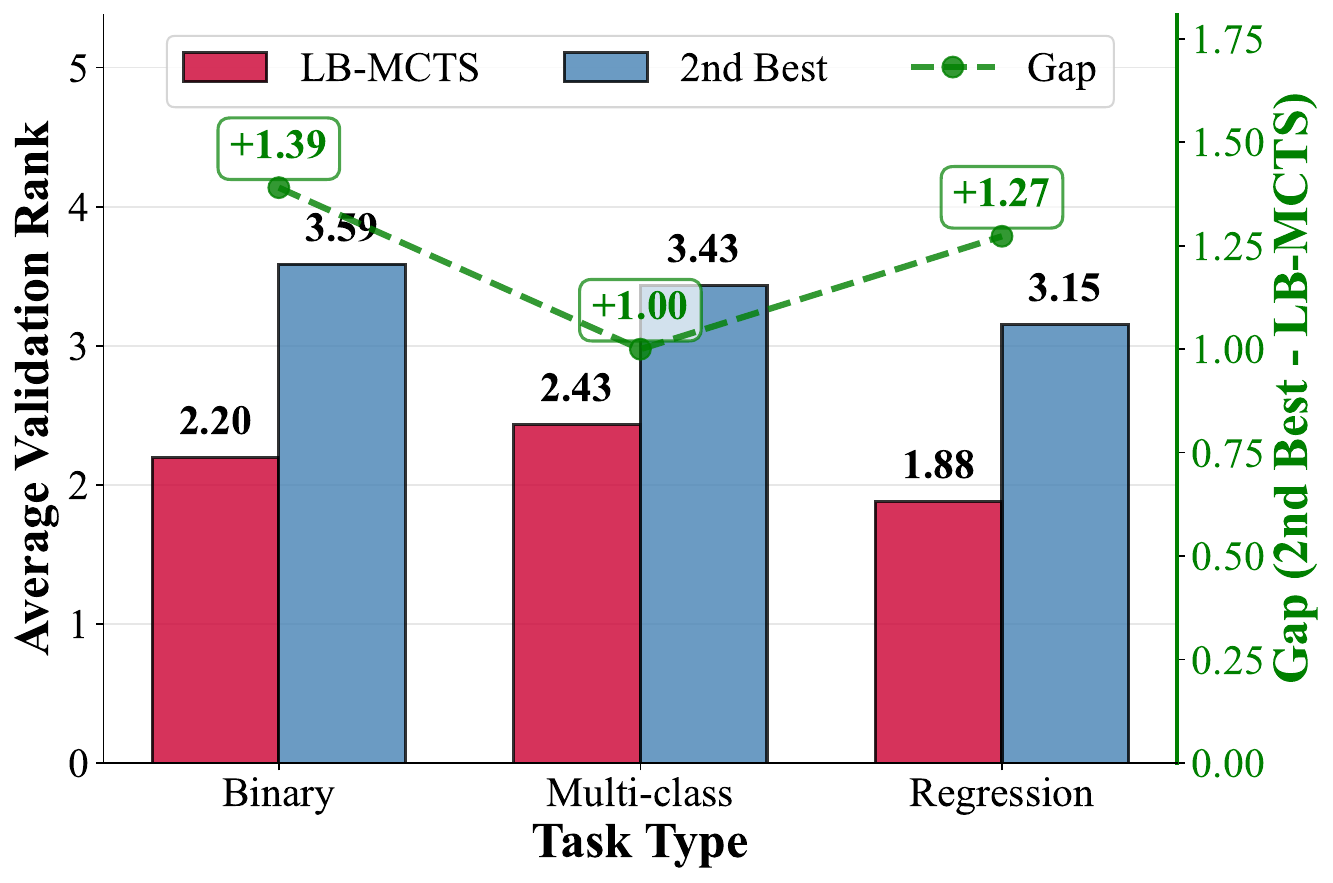}
        \caption{Performance by task type.}
        \label{fig:ana_task_type}
    \end{subfigure}
    \hfill %
    \begin{subfigure}[b]{0.31\linewidth} %
        \centering
        \includegraphics[width=\linewidth]{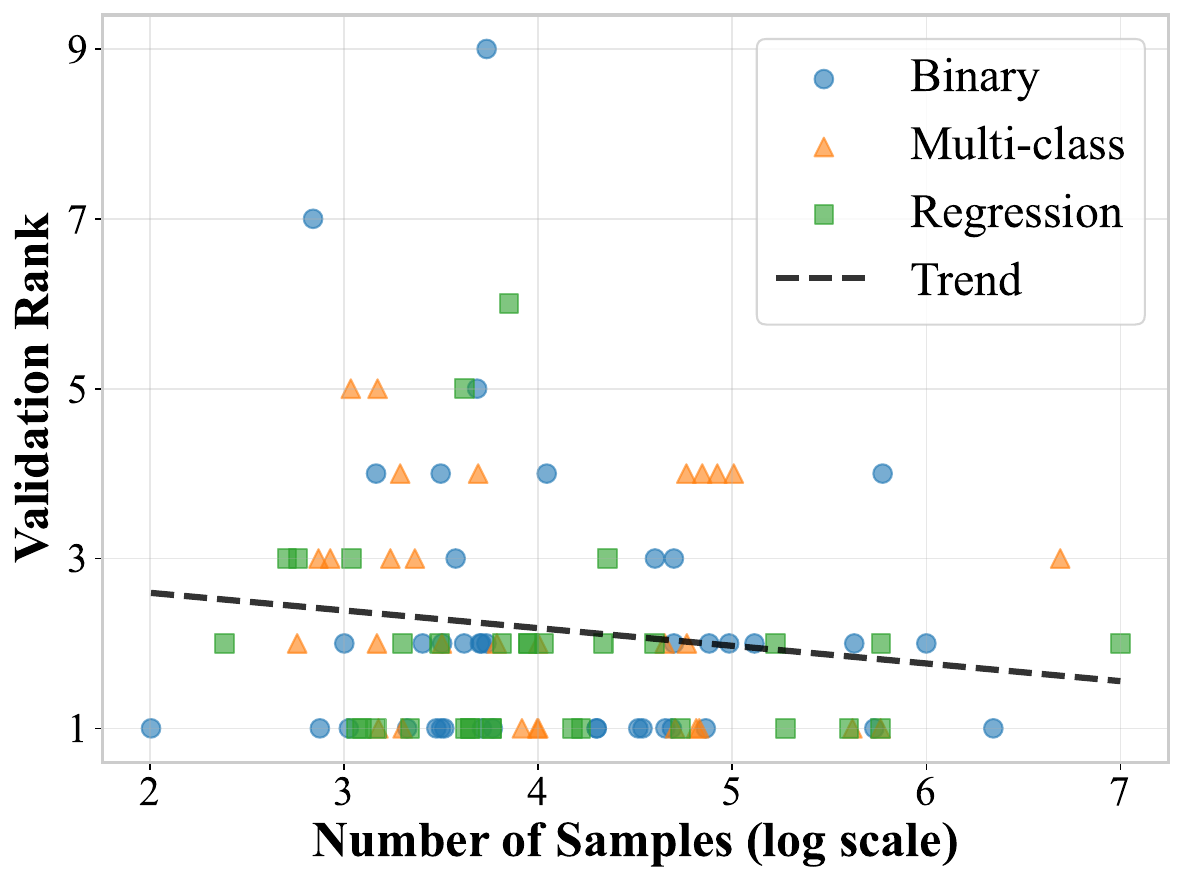}
        \caption{Performance by sample size.}
        \label{fig:ana_sample}
    \end{subfigure}
    \hfill %
    \begin{subfigure}[b]{0.31\linewidth} %
        \centering
        \includegraphics[width=\linewidth]{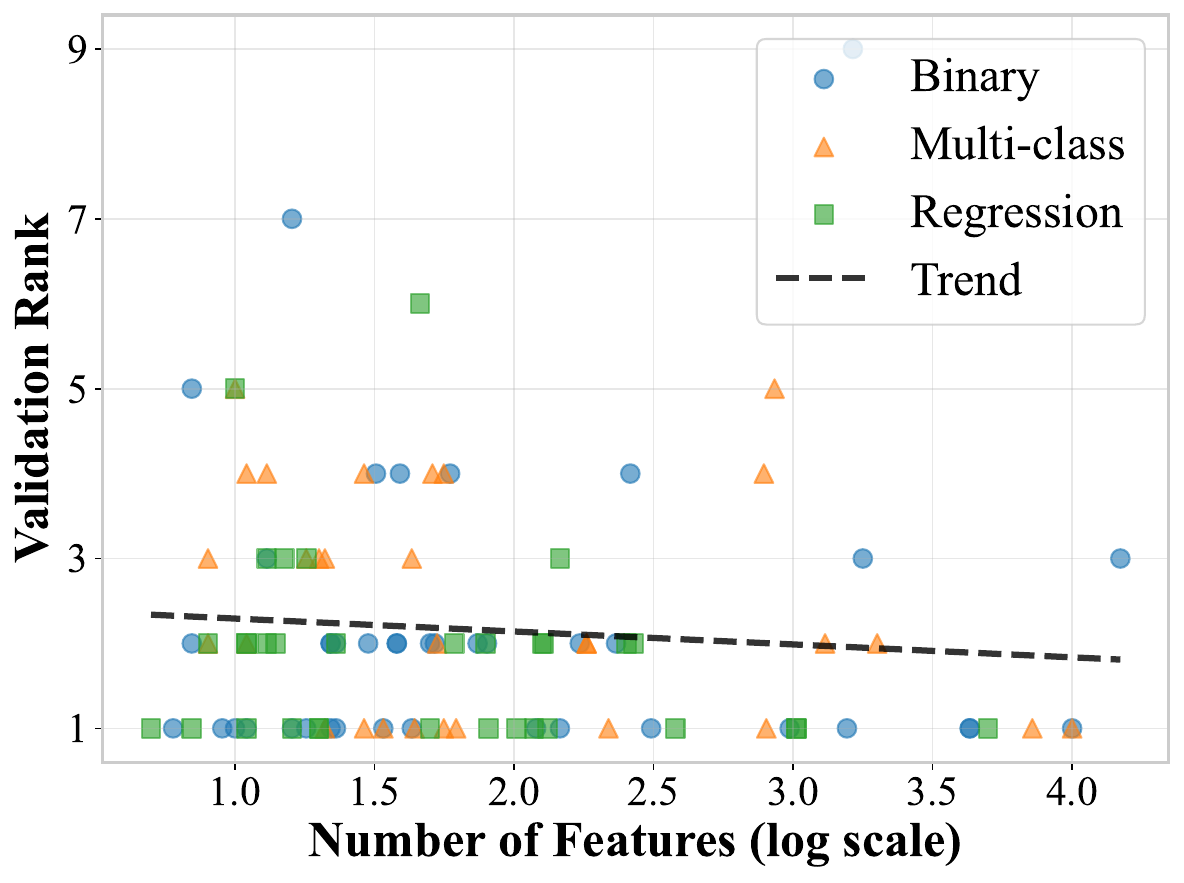}
        \caption{Performance by feature count.}
        \label{fig:ana_feature}
    \end{subfigure}
    
    \caption{Performance vs. task characteristics across 104 datasets.}
    \label{fig:analyze_rank} %
\end{figure*}

\subsection{Impact of Task Characteristics}
\label{app:task_characteristics}

We further analyze how the performance of \sys correlates with task characteristics by examining the final ranks after 300 iterations across all 104 datasets. \cref{fig:analyze_rank} visualizes the relationship between the validation rank and three key factors: task type, sample size, and feature dimensionality.

\textbf{Impact of Task Type.} \cref{fig:ana_task_type} compares the average rank of \sys against the second-best baseline across Binary Classification, Multi-class Classification, and Regression tasks. \sys consistently outperforms the runner-up with a significant rank advantage across Binary ($+1.39$), Multi-class ($+1.00$), and Regression ($+1.27$) tasks. Notably, it achieves the best performance on Regression tasks (Rank 1.88). This may be attributed to the continuous nature of the MSE metric used in regression, which provides smoother gradient-like feedback for the LLM's reflection mechanism compared to the discrete nature of Balanced Accuracy in classification.

\textbf{Impact of Problem Complexity.} \cref{fig:ana_sample} and \cref{fig:ana_feature} plot the validation rank against the $\text{log}_{10}$-scale sample size and feature count, respectively, with fitted trend lines. We observe a clear downward trend in rank (indicating better performance) as task complexity increases.
To quantify this, we partition the datasets into three equal-sized groups (Small/Medium/Large) based on sample size or feature count respectively.
\begin{itemize}[leftmargin=1.6em, topsep=0pt, partopsep=0pt, itemsep=3pt, parsep=0pt]
    \item \textbf{Sample Size:} The average rank progressively improves from 2.31 on small datasets ($n < 10^{3.54}$) to 2.24 on medium datasets ($10^{3.54} \le n < 10^{4.53}$), and reaches \textbf{1.94} on large datasets ($n \ge 10^{4.53}$).
    \item \textbf{Feature Count:} Similarly, the rank improves from 2.37 on low-dimensional tasks ($p < 10^{1.33}$) to 2.15 on medium-dimensional tasks ($10^{1.33} \le p < 10^{2.05}$), and achieves \textbf{1.97} on high-dimensional tasks ($p \ge 10^{2.05}$).
\end{itemize}
Overall, across all 104 datasets, \sys demonstrates robust dominance, achieving Rank 1 on 43 tasks and Rank $\le$ 3 on 87 tasks. These results suggest that \sys is particularly adept at handling complex, large-scale optimization problems where traditional methods often struggle to scale.

\subsection{Supplementary Experiments on Deep Learning and Feature Engineering}
\label{app:dl_fe_experiment}

Furthermore, we conducted supplementary experiments on a more diverse search space including both deep learning (DL) models and feature engineering (FE).
Specifically, we considered four DL models, MLP, ResNet-Tabular and FT-Transformer~\citeapp{gorishniy2021revisiting}, and TabNet~\citeapp{arik2021tabnet}, forming a 31-dimensional model hyperparameter space (Table~\ref{tab:dl_fe_search_space}), together with a 10-dimensional FE space.
In each model subtree, \sys jointly optimizes the model hyperparameters together with the 10-dimensional FE space.
This search space is shared by classification and regression tasks, and jointly optimizing FE with each model is natural since effective feature engineering is often model-dependent.

\begin{table}[t!]
\caption{Search space for DL models. The same space is used for classification and regression tasks.}
\label{tab:dl_fe_search_space}
\centering
\begin{tabular}{lccc}
    \toprule
    Type of Model & \#$\lambda$ & cat & cont \\
    \midrule
    MLP              & 5  & 3  & 2  \\
    ResNet-Tabular   & 7  & 4  & 3  \\
    FT-Transformer   & 9  & 6  & 3  \\
    TabNet           & 10 & 3  & 7  \\
    \midrule
    Total (4 algos)  & 31 & 16 & 15 \\
    \bottomrule
\end{tabular}
\end{table}

\begin{figure}[t!]
    \centering
    \includegraphics[width=0.75\linewidth]{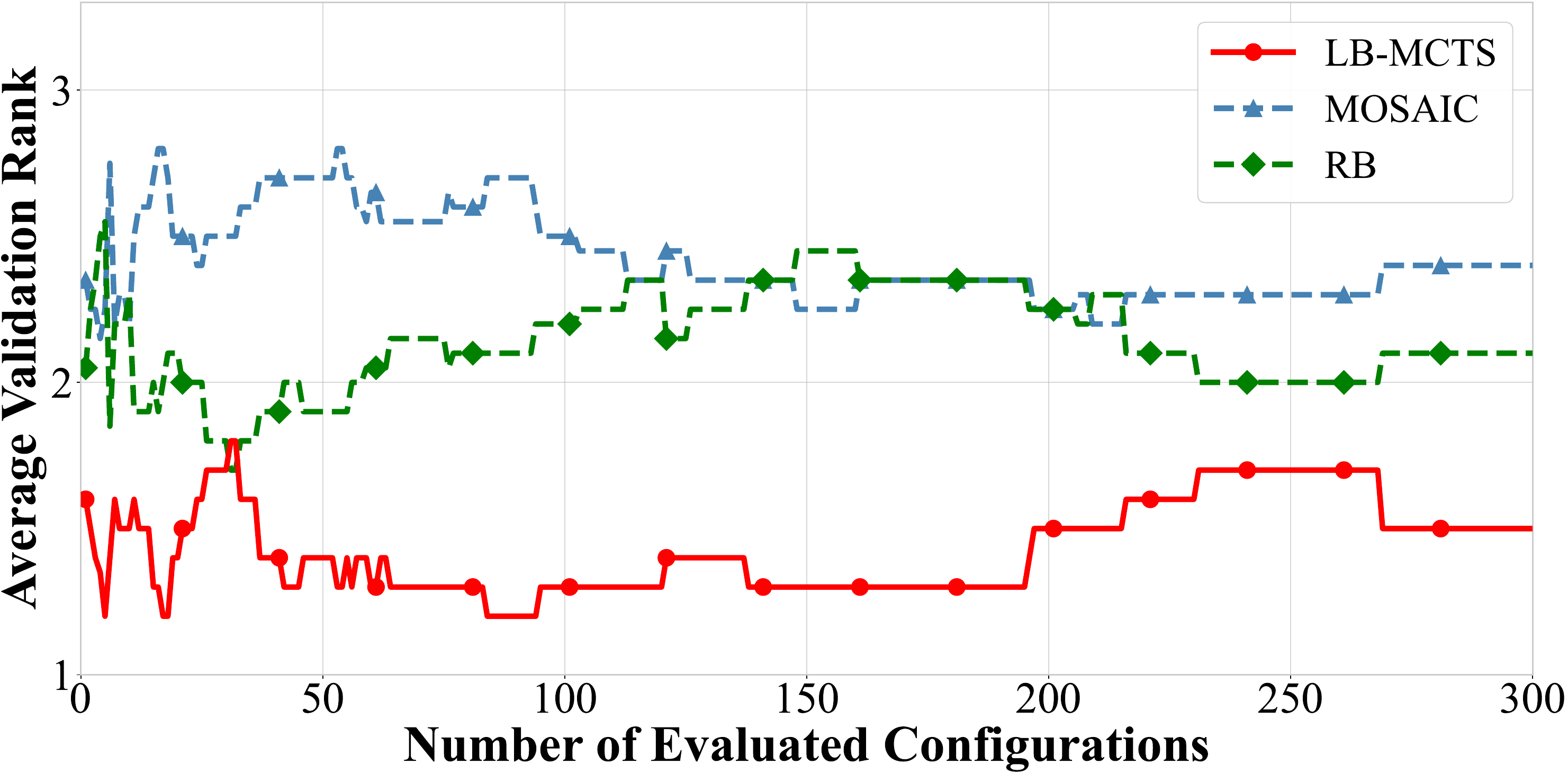}
    \caption{Average validation rank over 10 datasets on the deep learning pipeline scenario.}
    \label{fig:dl_fe_pipeline}
\vspace{-1em}
\end{figure}

Results across 10 representative datasets in \cref{fig:dl_fe_pipeline} show that \sys achieves the best average rank of 1.5, outperforming the top two baselines from our main experiments.
This confirms that our framework consistently excels at searching within structured spaces, regardless of the underlying model type.

\subsection{Post-hoc Ensemble Performance}
\label{app:ensemble}

In practical applications, an ensemble of promising configurations is sometimes preferred over a single best model, as ensembling can potentially enhance overall performance~\citeapp{auto-sklearn_feurer2015efficient,pseo_xu2025pseo}. To evaluate this auxiliary aspect, we perform post-hoc ensembling using the pool of 300 models identified during the optimization process (without retraining).
Among different ensemble strategies (e.g., Bagging, Boosting, Stacking), we adopt the ensemble selection~\citeapp{caruana2004ensemble}, which works empirically well with AutoML as shown in previous studies~\citeapp{mindware_li2023volcanoml} and~\cite{divbo_shen2022divbo,optdivbo_poduval2024cash}.
In short, ensemble selection starts from an empty ensemble and iteratively adds models from the pool with replacement to maximize the ensemble validation performance (with uniform weights).

\begin{figure*}[tb]
    \centering

    \begin{subfigure}[b]{0.49\linewidth} %
        \centering
        \includegraphics[width=\linewidth]{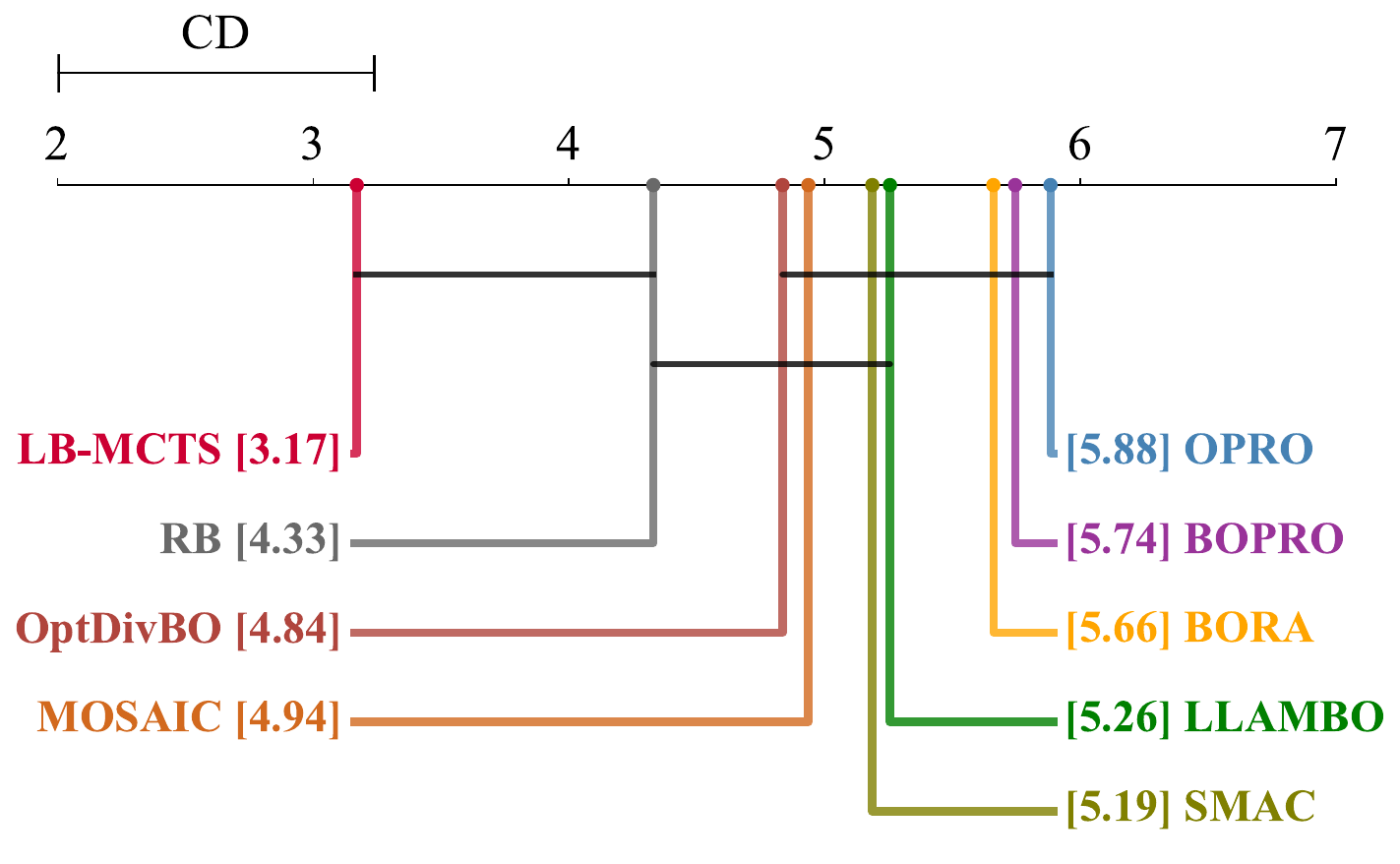}
        \caption{CD plot of ensemble test performance.}
        \label{fig:ens_cd_rank}
    \end{subfigure}
    \hfill %
    \begin{subfigure}[b]{0.47\linewidth} %
        \centering
        \includegraphics[width=\linewidth]{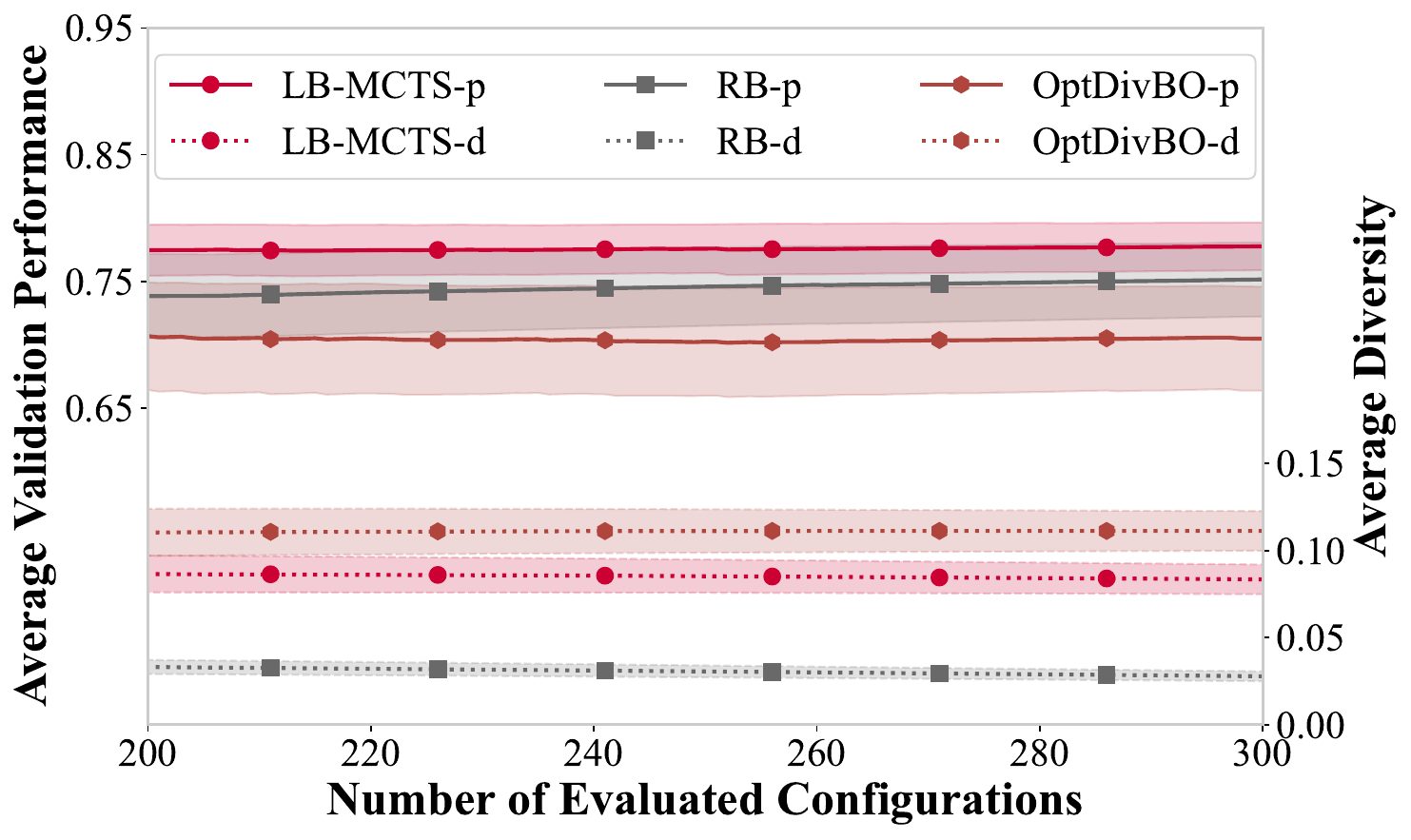}
        \caption{Average validation performance and diversity on ``ada''.}
        \label{fig:ens_diverdity}
    \end{subfigure}
    
    \caption{Post-hoc ensemble performance of 9 methods across 104 datasets.}
    \label{fig:main_figure_ens} %
\end{figure*}

Figure \ref{fig:ens_cd_rank} presents the \textit{test performance after ensemble selection} on 104 datasets. 
Two points emerge: (i) Notably, OptDivBO rises from the lowest-ranking BO baseline (in single-model settings) to the second-best position.
This validates that effective ensembling requires not just strong individual learners but also diversity among them~\citeapp{bian2021does,zhou2002ensembling}, which aligns precisely with OptDivBO’s motivation of considering model diversity.
(ii) \sys maintains its leading position, achieving \textbf{an average rank of 3.17}—a significant margin over the second-best performing method (4.33). This is attributed to its explicit balance of inter- and intra-algorithm exploration and exploitation; \sys not only identifies high-performing individual models but also ensures broad coverage, effectively achieving the trade-off between individual model performance and diversity among models.

To support this interpretation, Fig.~\ref{fig:ens_diverdity} reports, on the ``ada'' dataset, how (i) the \textit{average validation performance} (solid lines) of all evaluated models and (ii) the \textit{average inter-model diversity} (dashed lines) evolve over the course of optimization for the three methods with the best average ensemble performance. Following~\citeapp{zhang2020efficient} and \cite{divbo_shen2022divbo}, we quantify the diversity between two models \((\boldsymbol{\lambda}_i, \boldsymbol{\lambda}_j)\) as the average Euclidean distance between their predicted class-probability vectors on the validation set:
$$
\mathrm{Div}(x_i,x_j)=\frac{\sqrt{2}}{2}\cdot \frac{1}{|D_{\mathrm{val}}|}\sum_{s\in D_{\mathrm{val}}}\left\|T_{\boldsymbol{\lambda}_i}(s)-T_{\boldsymbol{\lambda}_j}(s)\right\|_2,
$$
where \(T_{x}(s)\) denotes the class-probability prediction of model \(x\) on validation example \(s\).
We can observe from the figure that OptDivBO yields the highest average diversity among the three models, but its average validation performance is substantially worse than the other two methods, resulting in an extreme “diverse-but-weak” pool. RB shows the opposite behavior: it achieves the second-best average validation performance but produces the least diverse model set. In contrast, \sys attains the best average validation performance while maintaining high diversity (second only to OptDivBO), providing a more balanced and thus more effective pool for post-hoc ensemble selection, which aligns with its strongest ensemble results. We note that this balance arises as a by-product of \sys’s exploration–exploitation design; explicitly optimizing for ensemble performance is left for future work.

\subsection{LLM Cost Comparison}
\label{app:cost}

\begin{figure}[t!]
    \centering
    \begin{subfigure}[b]{0.43\textwidth}
        \centering
        \includegraphics[width=\linewidth]{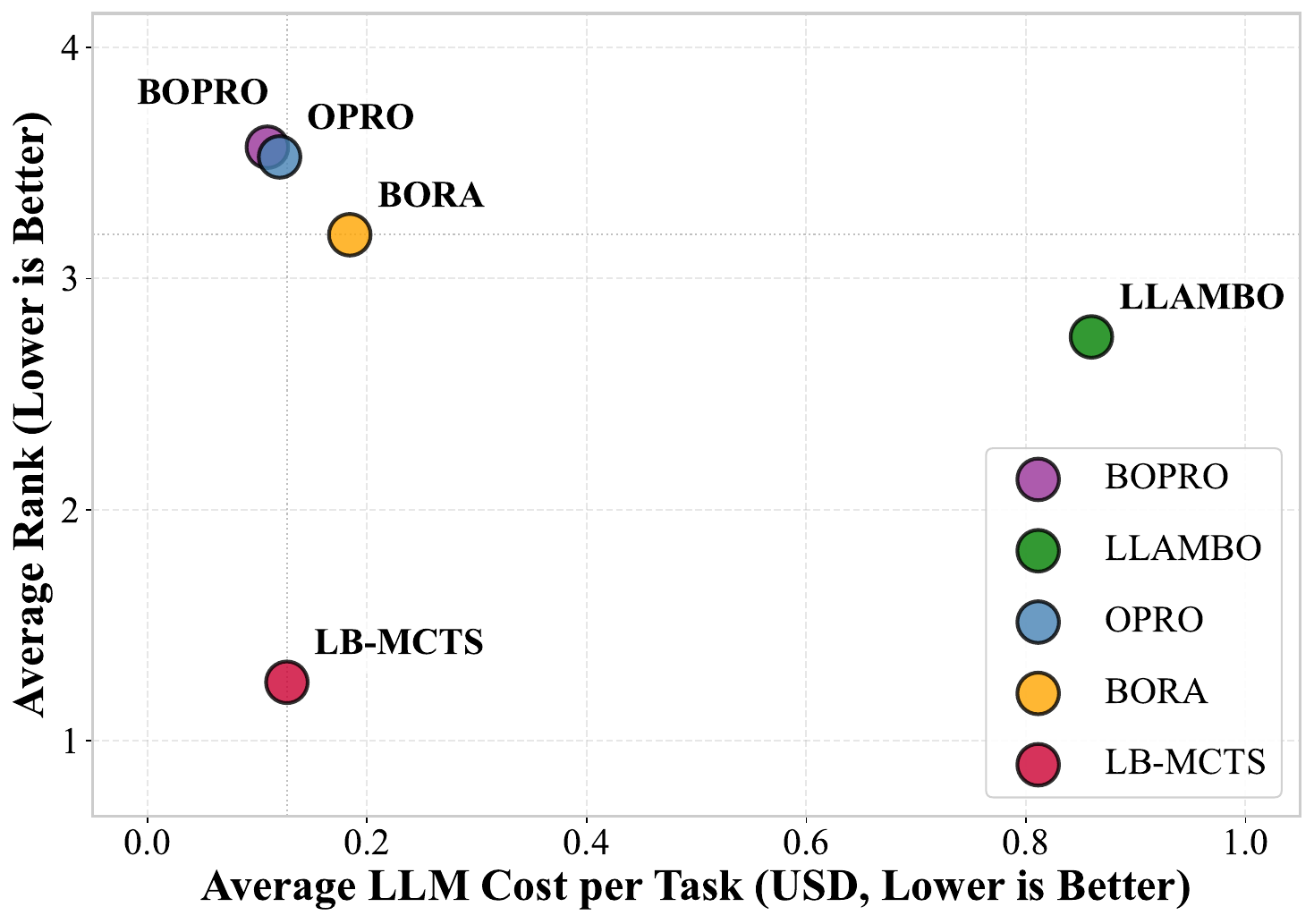}
        \caption{Cost-performance trade-off of LLM-based methods.}
        \label{fig:cost}
    \end{subfigure}
    \hspace{2em}
    \begin{subfigure}[b]{0.48\textwidth}
        \centering
        \includegraphics[width=\textwidth]{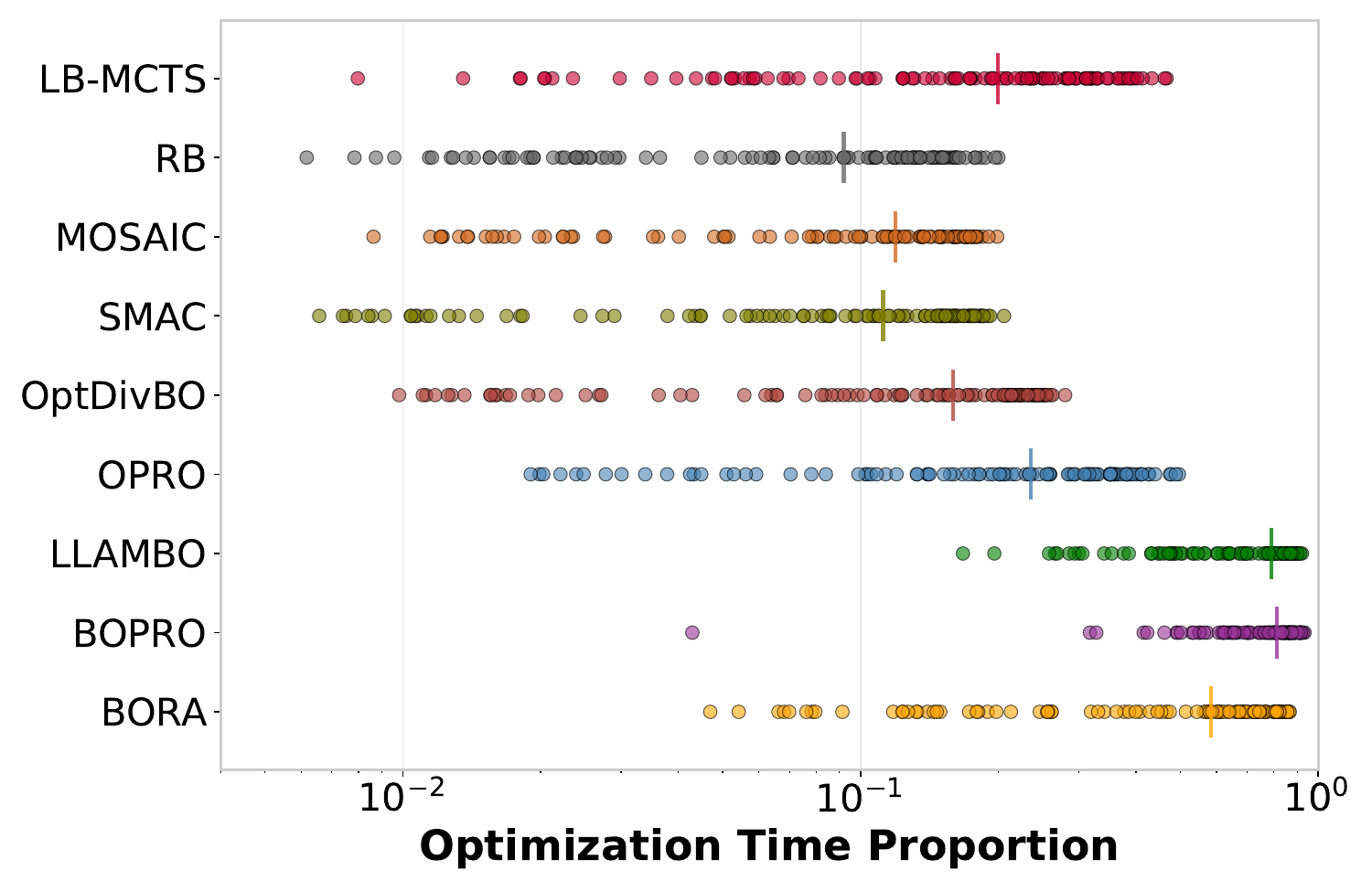}
        \caption{Proportion of time spent on optimization.}
        \label{fig:opt_time_ratio}
    \end{subfigure}
    \caption{Comprehensive Cost and Performance Analysis.}
    \label{fig:cost_analysis}
\end{figure}

To comprehensively evaluate the practicality of \sys, we analyze the cost from two perspectives: the monetary cost of LLM API usage and the temporal overhead of the optimizer. Figure~\ref{fig:cost_analysis} summarizes the results across 104 datasets.

\textbf{Monetary Cost vs. Performance.} 
We estimate API cost using model pricing at the time of experiments, accounting for input and output tokens.
\cref{fig:cost} illustrates the final average validation rank (across 104 datasets after 300 rounds) versus the average LLM API cost per task (in USD). The results highlight several key insights:
(i) \sys occupies the ideal bottom-left quadrant of the plot, achieving a state-of-the-art \textbf{average rank of 1.25} with an \textbf{economical cost of approximately \$0.127 per task}. Compared to other baselines, \sys provides a significant performance leap without a proportional increase in expenditure.
(ii) While LLAMBO attains the second-best rank (2.73), it is by far the most expensive method, costing \$0.860 per task—nearly 6.8$\times$ that of \sys. This high cost stems from its high-volume API calls per iteration, which do not translate into superior rankings compared to our tree-search approach.
(iii) Methods like BOPRO, OPRO, and BORA cluster in the low-cost region (\$0.11–\$0.18) but suffer from poor optimization efficacy, with average ranks exceeding 3.20. Although these methods are budget-friendly, their inability to effectively navigate complex search spaces limits their utility.

\textbf{Optimizer Time Overhead.} Figure~\ref{fig:opt_time_ratio} reports the ratio of time spent on optimization (proposal generation) to the total runtime (optimization + model evaluation) over 300 iterations. Ideally, the optimizer should be lightweight, directing the majority of the time budget toward model training and evaluation.
The results highlight several key insights:
(i) \sys maintains a low overhead with a median ratio of 0.20, meaning $\approx80\%$ of the time is productively spent on evaluation. This is significantly more efficient than other LLM-based approaches like OPRO (0.2356) and BORA (0.5841). The most extreme examples are LLAMBO (0.80), which suffers from excessive API calls, and BOPRO (0.81), which is slowed down by embedding model computations for all configurations.
(ii) Scalability to Complex Tasks: Intuitively, as the dataset scale increases, model evaluation time grows, naturally reducing the optimizer's relative overhead. Indeed, for large-scale tasks, \sys's overhead ratio drops as low as 0.008. Crucially, this aligns perfectly with our finding in Appendix~\ref{app:task_characteristics} that \sys's performance advantage is most pronounced on large, complex datasets, making it highly suitable for real-world, high-stakes optimization scenarios.
(iii) From a wall-clock perspective, in our GPT-4o-mini benchmark runs, the average per-iteration time costs are 13.78s for LLM inference and 2.56s for BO computation, while the black-box evaluation (model training and validation) averages 63.02s. The LLM overhead is therefore worthwhile because better proposals reduce the number of expensive evaluations needed: \sys achieves average end-to-end wall-clock speedups of 2.0$\times$ over RB and 2.2$\times$ over MOSAIC, the second- and third-ranked baselines, respectively.
(iv) Finally, we note that as LLM inference speeds continue to improve, the temporal overhead of \sys will further decrease, enhancing its applicability to time-sensitive tasks.

\begin{table}[t!]
    \centering
    \caption{Overview of the 10 AMLB datasets selected for ablation studies.}
    \label{tab:ablation_datasets}
    \resizebox{0.8\textwidth}{!}{%
    \begin{tabular}{l r r r c}
        \toprule
        \textbf{Task Type} & \textbf{Dataset Name (OpenML ID)} & \textbf{Samples ($n$)} & \textbf{Features ($p$)} & \textbf{Classes ($C$)} \\
        \midrule
        \multirow{4}{*}{\makecell[l]{Binary \\ Classification}} 
        & ada (190411) & 4,147 & 49 & 2 \\
        & blood-transfusion (359955) & 748 & 5 & 2 \\
        & sylvine (359972) & 5,124 & 21 & 2 \\
        & qsar-biodeg (359956) & 1,055 & 41 & 2 \\
        \midrule
        \multirow{3}{*}{\makecell[l]{Multi-class \\ Classification}} 
        & cmc (359959) & 1,473 & 9 & 3 \\
        & GesturePhaseSegmentation (359970) & 9,873 & 32 & 5 \\
        & connect-4 (359977) & 67,557 & 42 & 3 \\
        \midrule
        \multirow{3}{*}{Regression} 
        & house\_prices\_nominal (359951) & 1,460 & 79 & - \\
        & pol (359946) & 15,000 & 48 & - \\
        & quake (359930) & 2,178 & 3 & - \\
        \bottomrule
    \end{tabular}%
    }
\end{table}

\subsection{Representative Dataset Selection for Ablation Study}
\label{app:dataset_selection}

Conducting hyperparameter optimization (HPO) with Large Language Models (LLMs) incurs substantial computational overhead and API costs. To balance experimental rigor with resource constraints, we selected a representative subset of \textbf{10 datasets} from the AutoML Benchmark (AMLB) suite for our ablation studies.
As summarized in Table \ref{tab:ablation_datasets}, these datasets were carefully chosen with three criteria:
(i) Diverse task types: Including Binary Classification (4), Multi-class Classification (3), and Regression (3). This distribution is aligned with that of the original AMLB benchmark.
(ii) Varying data scale: Sample sizes ($n$) ranging from 748 to 67,557.
(iii) Varying feature dimensionality: Feature counts ($p$) ranging from 4 to 80.
As all the datasets are collected from
OpenML~\citeapp{openml_vanschoren2014openml}, we provide the OpenML ID as the identification of the dataset.

To quantify the representativeness of this subset, we compare the nine methods in the main experiment (\cref{fig:validation_rank}) at 100, 200, and 300 evaluations.
At each checkpoint, we rank the methods on each dataset by their best-so-far validation performance and average these ranks over the original 10-dataset subset and over the full 104-dataset benchmark, respectively.
We then compute the Spearman and Kendall correlations between the resulting two nine-method average-rank vectors.

\begin{table}[htbp]
    \centering
    \caption{Rank correlations between the nine-method average-rank vectors on the original 10-dataset subset and the full 104-dataset benchmark.}
    \label{tab:ablation_subset_correlation}
    \begin{tabular}{rcc}
        \toprule
        \textbf{Evaluations} & \textbf{Spearman $\rho$} & \textbf{Kendall $\tau$} \\
        \midrule
        100 & 0.783 & 0.611 \\
        200 & 0.850 & 0.722 \\
        300 & 0.971 & 0.930 \\
        \bottomrule
    \end{tabular}
\end{table}

Table~\ref{tab:ablation_subset_correlation} shows that the ranking agreement is already substantial at 100 evaluations and increases with the evaluation budget, becoming near-perfect at 300 evaluations.
These results indicate that the original 10 datasets capture the relative performance trends of the methods in the main comparison, supporting their use for the ablation studies.

\subsection{Ablation Study: Exploration and Exploitation Strategy}
\label{app:ablation_exploration_exploitation}

\begin{figure}[t!]
    \centering
    \begin{subfigure}[b]{0.42\textwidth}
        \centering
        \includegraphics[width=\textwidth]{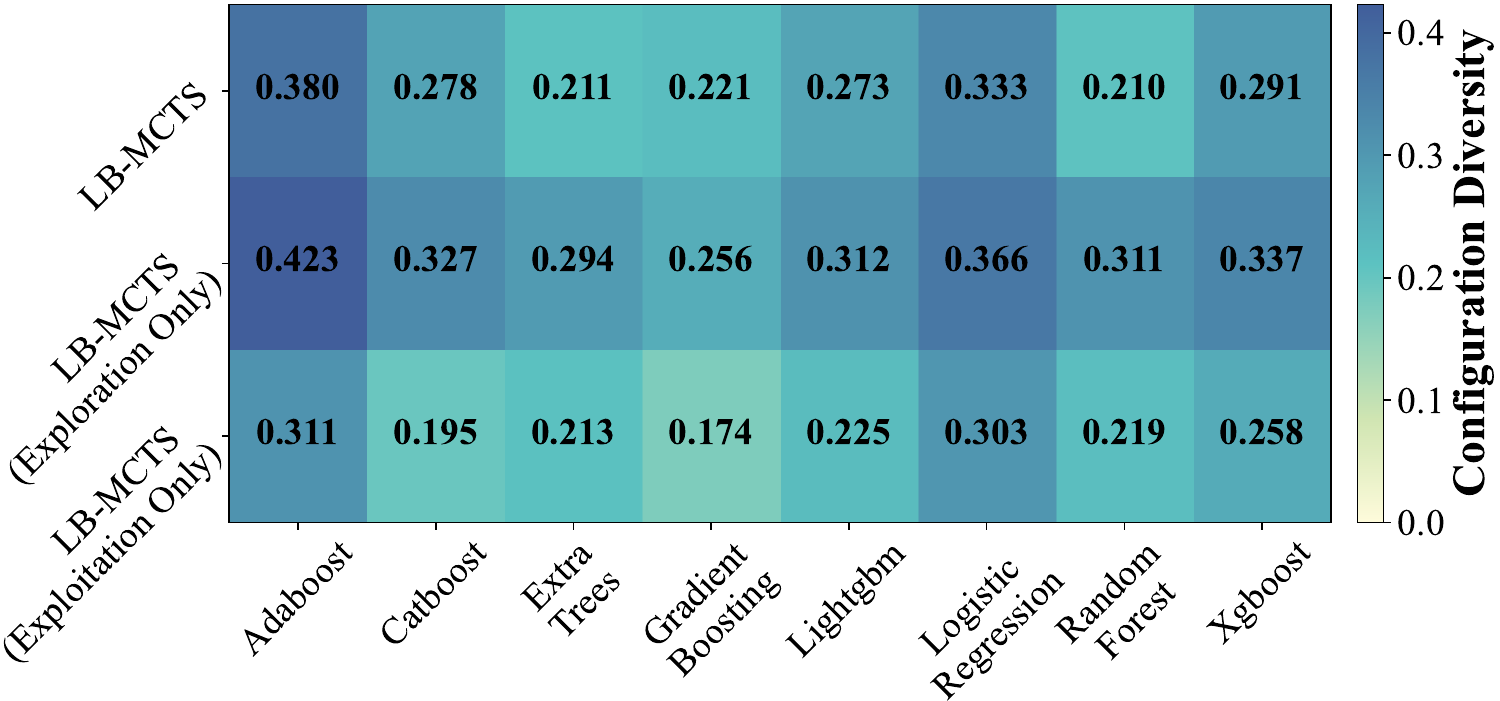}
        \caption{Average diversity heatmap across 10 datasets.}
        \label{fig:diversity_heatmap}
    \end{subfigure}
    \hfill
    \begin{subfigure}[b]{0.57\textwidth}
        \centering
        \includegraphics[width=\textwidth]{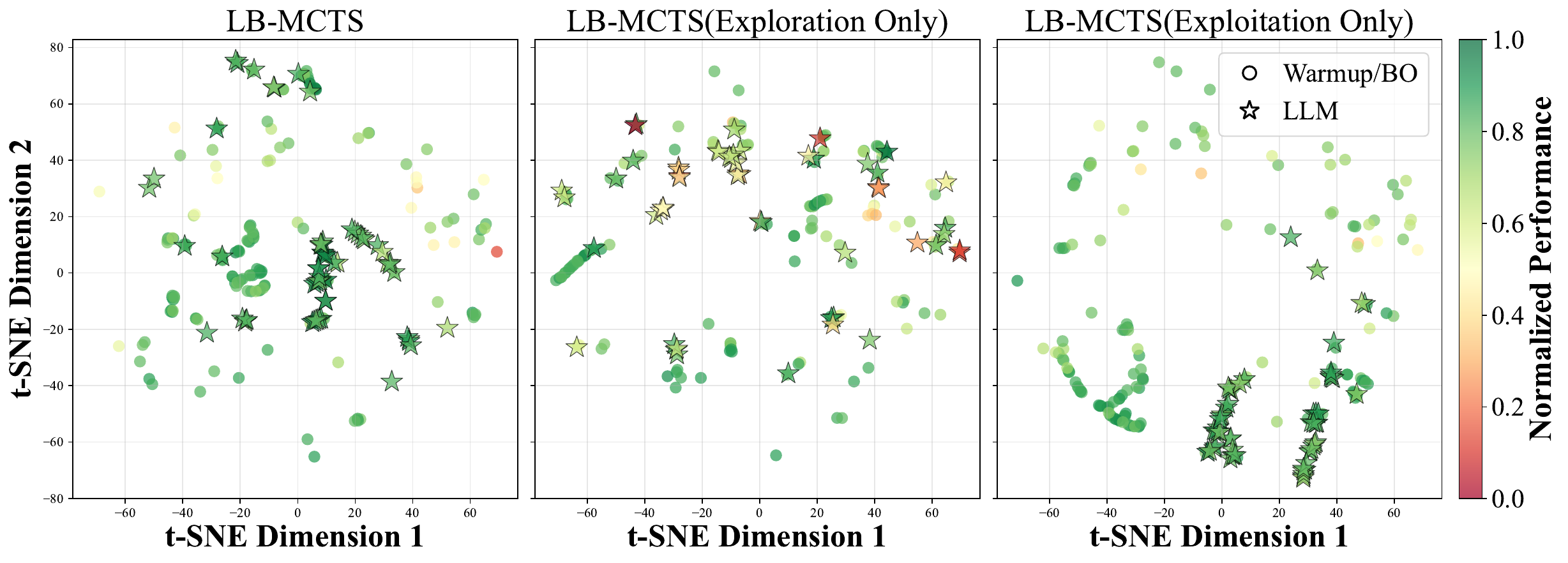}
        \caption{Distribution visualization of evaluated configurations with t-SNE.}
        \label{fig:dominant_algo_curve}
    \end{subfigure}
    \caption{Visualizations for the Exploration-Exploitation Ablation Study. (a) compares the hyperparameter diversity, showing LB-MCTS achieves a balance between the extremes. (b) visualization of dominant algorithm ``Adaboost'' on ``qsar-biodeg''}
    \label{fig:ablation_visuals}
\end{figure}

\textbf{Implementation details of LB-Flat}. 
To isolate the contribution of the hierarchical tree structure and trajectory-aware reasoning, we implement LB-Flat, a variant that collapses the MCTS into a two-level structure (Root and Algo Nodes).
It treats all evaluated configurations and performance within an algorithm's subspace as an unstructured history.
The BO proposer remains unaffected, as it still suggests candidates via random sampling and local refinement of historical observations. Similarly, the Proposer Selection mechanism is unchanged, continuing to calculate based on the BO surrogate model’s ability to fit the existing observations for each algorithm. 
In contrast, the LLM proposer is significantly modified: it no longer employs PUCT rules to select basic node configurations for refinement via exploration or exploitation directives, but prompts the LLM to directly recommend the next configuration. 
Due to the absence of optimization trajectories, Global and Local memories are discarded; instead, the top-20 historical observations are used as memory (following the OPRO approach). 
Finally, the reflection prompt is adjusted by removing all information related to basic configurations.

\textbf{Tuning behaviors of \textsc{Exploration} and \textsc{Exploitation}}. This appendix extends the ablation study on \sys's \textsc{Exploration} and \textsc{Exploitation} strategies by investigating their impact on the distribution of proposed configurations. 
We first analyze the \textit{diversity of proposed configurations}.
Unlike our previous analysis of ensemble diversity, which considered the predictive differences between models on the validation set, here we quantify diversity directly at the level of the proposed hyperparameter configurations themselves. This allows for a more direct behavioral analysis of how different strategies shape the search space.
Specifically, the diversity score is calculated independently for each algorithm. For a given algorithm $A^i$ with a hyperparameter space $\Lambda^i$, the overall diversity score $D(A^i)$ is defined as the mean of the normalized diversity scores of all its hyperparameters:
$D(A^i) = \frac{1}{|\Lambda^i|} \sum_{p \in \Lambda^i} S(p)$, 
where $S(p)$ denotes the normalized diversity score for a specific parameter $p$. The calculation of $S(p)$ depends on the parameter type (numeric or categorical).

\begin{itemize}[leftmargin=1.6em, topsep=0pt, partopsep=0pt, itemsep=3pt, parsep=0pt]
    \item 
    \textbf{Numeric Parameters.} For continuous or integer parameters (e.g., learning rate, max depth), we normalize sampled values to $x' \in [0, 1]$ using linear or log-domain min-max scaling as appropriate. The diversity score $S_{num}(p)$ is then defined as the standard deviation of these normalized samples:
$S_{num}(p) = \text{std}(\{x'_i\})$.

    \item 
    \textbf{Categorical Parameters.}
We measure categorical diversity using normalized Shannon entropy:
$S_{cat}(p) = \frac{-\sum_{k=1}^K q_k \ln(q_k)}{\ln(K)}$, 
where $q_k$ is the empirical probability of category $k$ among $K$ possibilities. This ensures $S_{cat}(p) \in [0, 1]$, with 1 representing a perfectly uniform distribution (maximum diversity).
\end{itemize}

\cref{fig:diversity_heatmap} presents the heatmap of diversity scores by algorithm across all 10 datasets.
Note that diversity is quantified per algorithm rather than across algorithms, preventing bias from different hyperparameter dimensionalities.
We can observe that
the \textsc{Exploration} Only strategy (middle row) exhibits the highest diversity across almost all algorithms, indicating a broad but potentially inefficient search that mimics random sampling.
The \textsc{Exploitation} Only strategy (bottom row) shows the lowest diversity, suggesting rapid convergence to a narrow region of the hyperparameter space, carrying the risk of getting trapped in local optima.
\sys (top row) maintains an intermediate diversity level. This confirms that our method effectively balances the trade-off, exploring sufficient configuration space before focusing resources on promising regions.
To further illustrate how these diversity patterns arise, Figure~\cref{fig:dominant_algo_curve} visualizes the evaluated configurations for the AdaBoost algorithm on the ``qsar-biodeg'' dataset using t-SNE, with LLM-proposed points marked as stars and warmup/BO proposals as circles. 
Under the \textsc{Exploration} Only strategy (middle), LLM proposals are widely scattered over the space, with many mediocre or poor configurations. 
Conversely, the \textsc{Exploitation} Only case (right) shows LLM proposals tightly clustered in a single area; while the performance is acceptable, the search fails to expand beyond this local neighborhood, leaving the majority of the space unexplored.
In contrast, \sys (left) generates a distribution that forms distinct, high-performing clusters (dark green stars) while retaining separation between them. This demonstrates that \sys successfully guides the search to exploit promising modes deeply while periodically exploring new regions to escape stagnation.
Collectively, these analyses provide a mechanistic explanation for the accelerated and superior convergence of \sys observed in \cref{fig:ee_hpo}.

\subsection{Ablation of PUCT-Based Tree Search}
\label{app:ablation_puct}

In this appendix, we conduct ablation studies on 10 representative datasets to first evaluate whether PUCT is an appropriate tree-search policy for \sys, and then analyze the sensitivity of its exploration weight $c_{puct}$.

\begin{figure}[t!]
    \centering
    \begin{subfigure}[t]{0.48\linewidth}
        \centering
        \includegraphics[width=\linewidth]{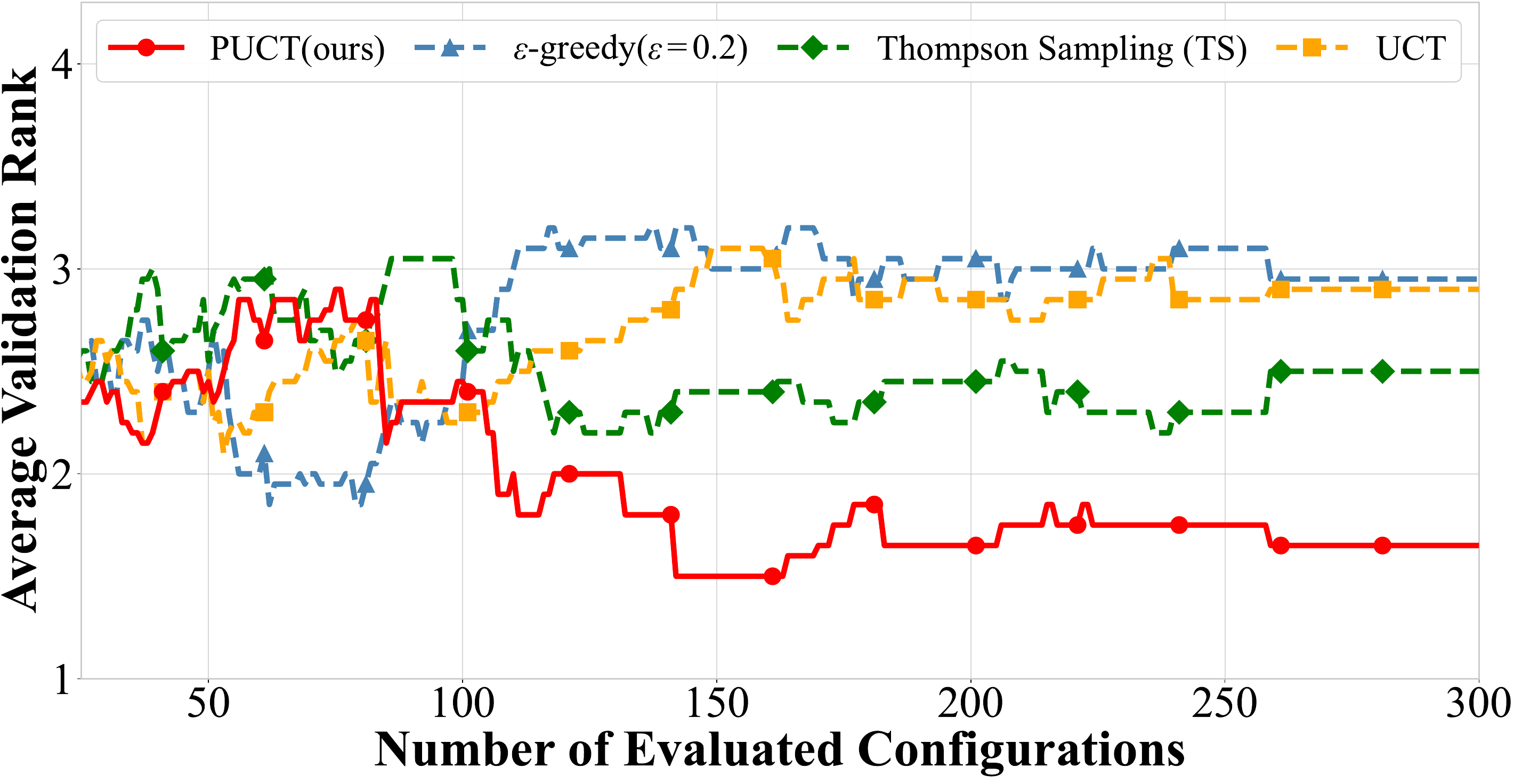}
        \caption{Tree-search strategy.}
        \label{fig:ablation_of_puct}
    \end{subfigure}
    \hfill
    \begin{subfigure}[t]{0.49\linewidth}
        \centering
        \includegraphics[width=\linewidth]{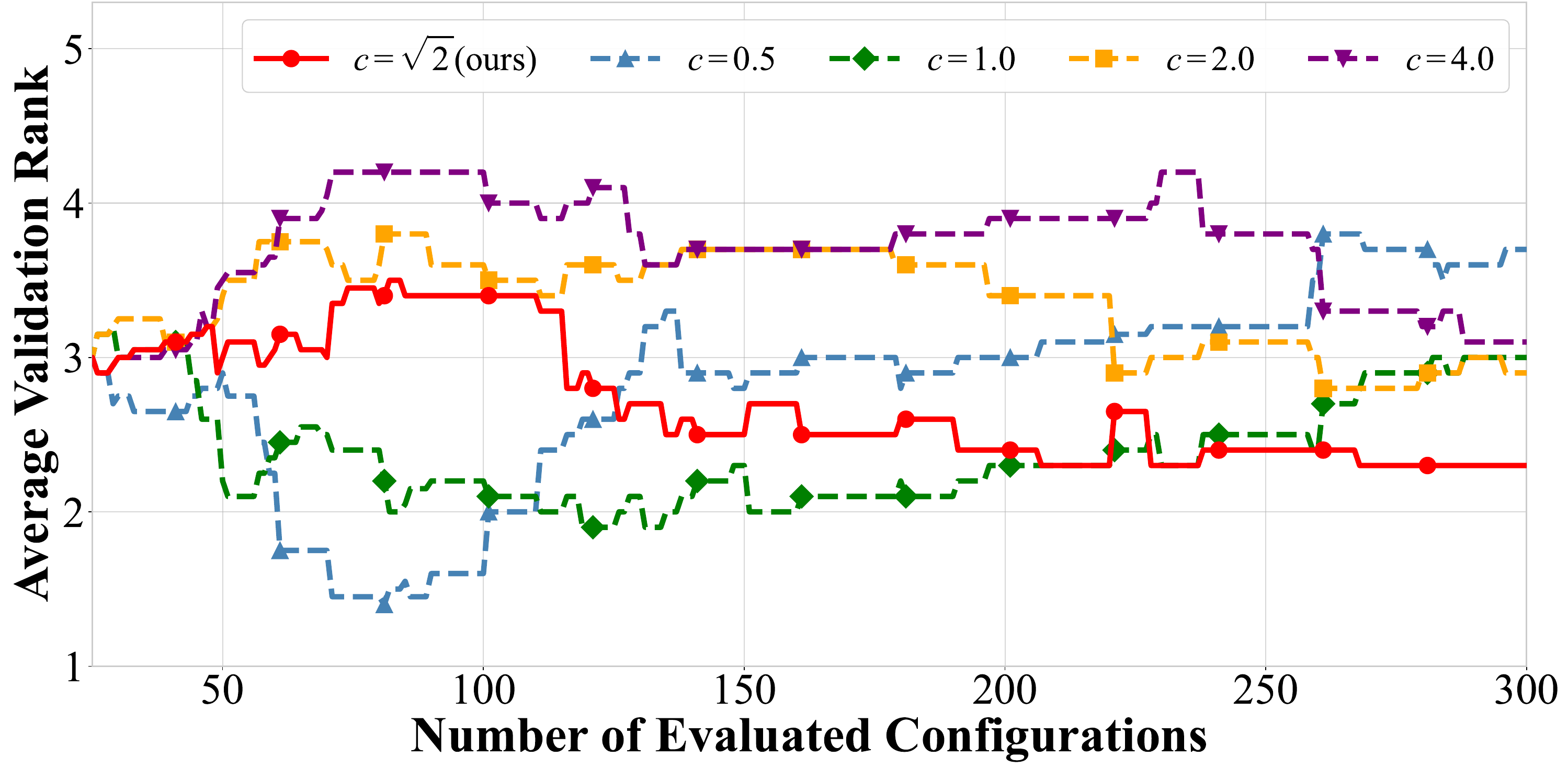}
        \caption{Exploration weight $c_{puct}$.}
        \label{fig:ablation_of_cpuct}
    \end{subfigure}
    \caption{Ablations of the PUCT-based tree search.}
    \label{fig:puct_ablation}
\vspace{-1em}
\end{figure}

\textbf{Choice of tree-search policy.}
We adopt PUCT because it combines empirical value estimates with prior-guided exploration, making it well suited for structured CASH search.
As shown in \cref{fig:ablation_of_puct}, PUCT achieves the best average rank (1.65), outperforming Thompson Sampling (2.50), UCT (2.90), and $\epsilon$-greedy (2.95).
The advantage comes from PUCT's ability to bias exploration toward globally promising branches while still preserving sufficient search diversity.
In contrast, UCT ignores prior information and may spend evaluations on clearly weak branches.
$\epsilon$-greedy tends to over-commit to the current best branch, with exploration occurring only through random perturbations, making it vulnerable to local optima.
Thompson Sampling provides stochastic exploration, but lacks the explicit prior-weighted control used by PUCT, leading to less stable search in our setting.

\textbf{Exploration-weight sensitivity.}
We further evaluate the sensitivity to the exploration weight $c_{puct}$.
As shown in \cref{fig:ablation_of_cpuct}, small values lead to over-exploitation and premature convergence, whereas large values over-emphasize exploration and slow down refinement.
The results show that intermediate values are generally stable, and our default choice $c_{puct}=\sqrt{2}$ maintains strong performance throughout the search process.

\subsection{Ablation Study: Reflection Mechanism}
\label{app:ablation_reflection}

\begin{figure}[t!]
    \centering
    \begin{minipage}[b]{0.48\textwidth}
        \centering
        \includegraphics[width=\linewidth]{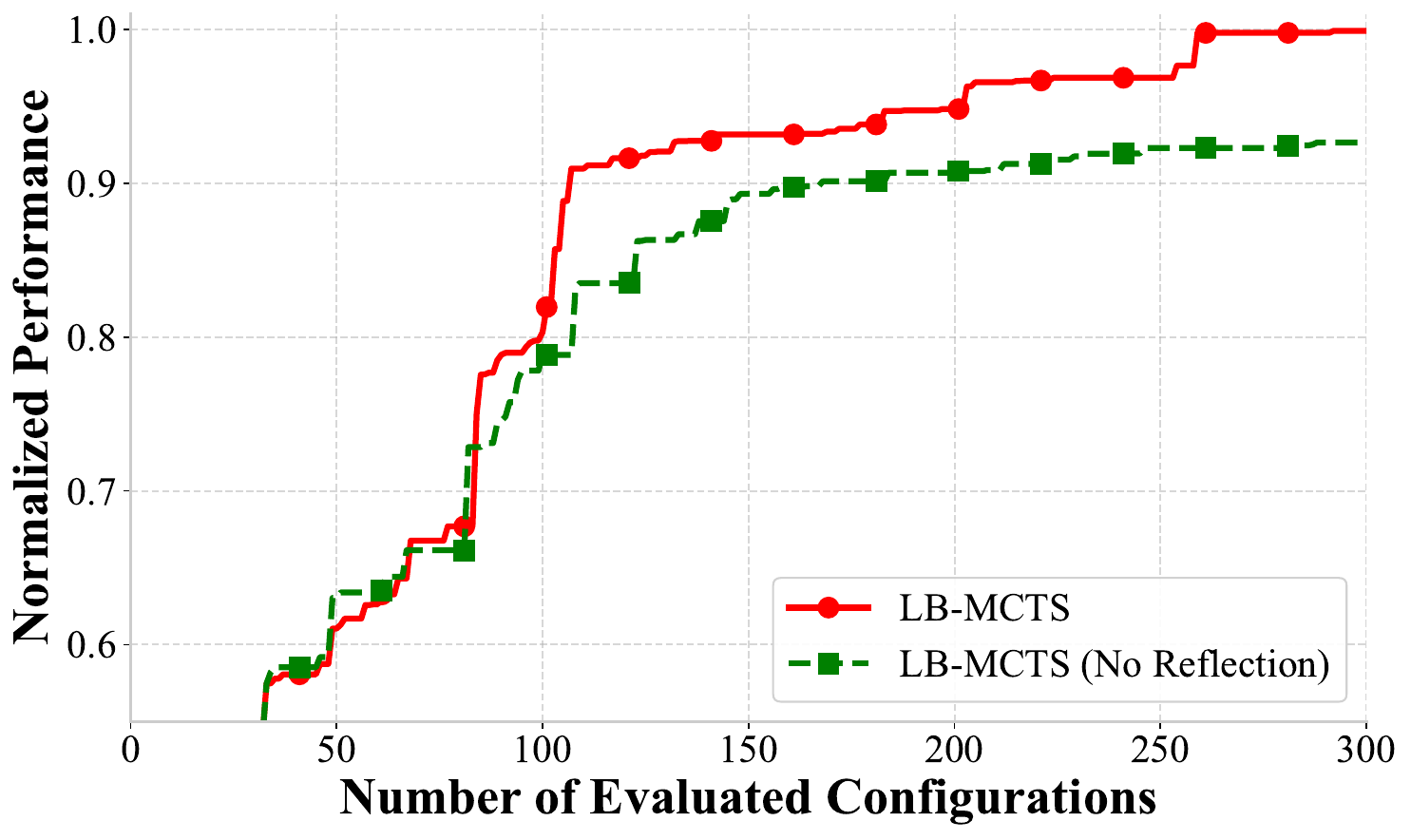}
        \caption{Ablation study on Reflection over 10 datasets.}
        \label{fig:reflection_ablation}
    \end{minipage}
    \hfill %
    \begin{minipage}[b]{0.48\textwidth}
        \centering
        \includegraphics[width=\linewidth]{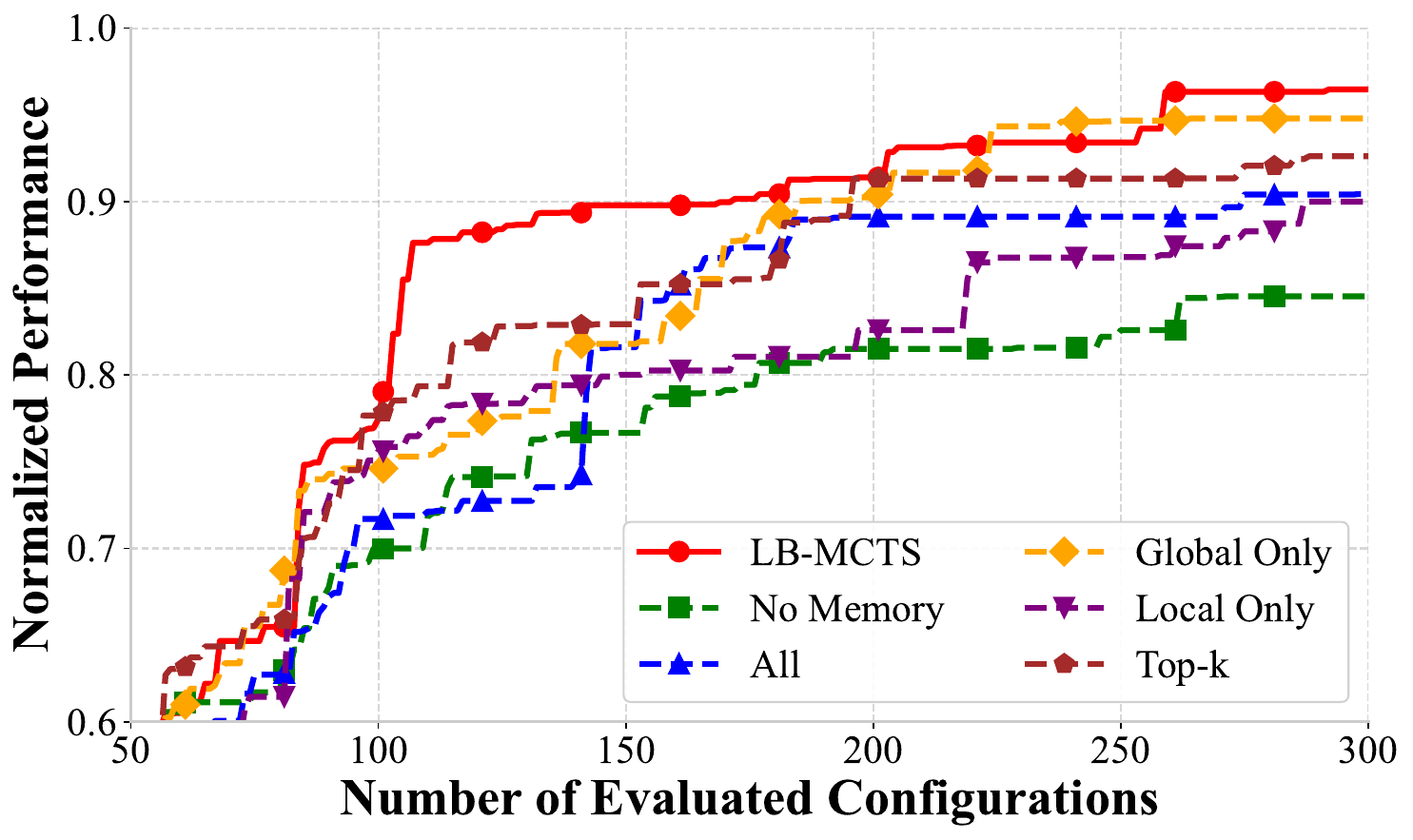}
        \caption{Ablation study on STM over 10 datasets.}
        \label{fig:memory_ablation} %
    \end{minipage}
\vspace{-1em}
\end{figure}

To assess the contribution of the LLM's self-reflection capability, we conducted an ablation study by deactivating the reflection module (denoted as No Reflection) while keeping all other components unchanged. We aggregated the \textit{normalized best-achieved validation performance} across the 10 selected datasets to analyze the global trend.
As observed in \cref{fig:reflection_ablation},
(i) \sys consistently maintains a higher normalized score throughout the optimization process.
(ii) The performance gap becomes more pronounced in the later stages (e.g., after 110 iterations). 
Finally, \sys achieves a superior \textbf{average validation rank of 1.1}, compared to 1.7 for the No Reflection baseline.
This indicates that while both methods start similarly, the reflection mechanism effectively enables the framework to learn from historical failures and successes.
It acts as a trajectory-level state summarizer to improve planning in the long run rather than a reactive proposal.

\subsection{Ablation Study: Selective Tuning Memory}
\label{app:ablation_memory}

In this appendix, we validate the effectiveness of our Selective Tuning Memory (STM) mechanism. Unlike prior methods that mix history from different algorithms, a key prerequisite of our framework is the isolation of optimization history by algorithm to prevent cross-algorithm noise. 
Building on this cleaner, algorithm-specific history, we adapt and thereby strengthen retrieval strategies from existing LLM-based optimizers for fair comparison:
(1) All: including the entire history of the current algorithm, mimicking the strategy of BORA and LLAMBO;
(2) Top-$k$: retrieving only the best-performing trials, mimicking OPRO (here we set $k=20$ to match the context length of OPRO).
Next, we ablate the components of our STM design:
(3) Global Only: retaining only the Pareto-selected global memory;
(4) Local Only: retaining only the ancestral trajectory.
Note that \textit{all variants operate on per-algorithm histories (no cross-algorithm mixing)}, so differences come purely from the retrieval strategy within each algorithm.
Finally, (5) we include a lower bound baseline, No Memory, which provides zero historical trials.

Figure~\ref{fig:memory_ablation} plots the normalized performance averaged over 10 datasets (detailed in Appendix~\ref{app:dataset_selection}). Five observations can be concluded:
(i) LB-MCTS (full STM) outperforms all variants, achieving the \textbf{best final average rank of 1.80}. 
(ii) We observe that strategies relying solely on high-performing samples—Global Only (Rank 2.30) and Top-$k$ (Rank 2.80)—perform reasonably well but are suboptimal. This indicates that while knowing ``where the best solutions are'' is useful, lacking the context of ``how we got here'' (trajectory) limits the LLM's reasoning ability.
In particular, trajectory information may help the LLM (a) avoid repeatedly moving back into regions that have already been explored and found unpromising, and (b) infer trends between parameter changes and performance changes—analogous to a coarse, history-based gradient signal—both of which are absent when only Global Memory are shown.
(iii) Global Only outperforms Top-$k$, confirming that retrieving trials based on the Pareto frontier of similarity and performance provides more relevant guidance to the current expansion task than simply selecting based on raw performance alone.
(iv) All (Rank 3.60) exposes the entire history and therefore includes the trajectory, but suffers from severe context pollution—many mediocre or redundant trials dilute the useful signal and consume the prompt budget.
(v) Finally, Only Local (Rank 4.30) and No Memory (Rank 5.40) yield the poorest results, underscoring that without a set of high-quality global references, the search struggles to efficiently locate optimal regions.

\bibliographystyleapp{plainnat}
\bibliographyapp{LB-MCTS_appendix}

\clearpage
\section*{NeurIPS Paper Checklist}

\begin{enumerate}

\item {\bf Claims}
    \item[] Question: Do the main claims made in the abstract and introduction accurately reflect the paper's contributions and scope?
    \item[] Answer: \answerYes{} %
    \item[] Justification: The abstract and \cref{sec:introduction} summarize the scope and contributions of \sys, including tree-structured LLM--BO collaboration, STM, dynamic proposer selection, and evaluation on 104 AMLB datasets.
  These claims are supported by \cref{sec:tree_structured_search_for_cash,sec:algorithm_selection,sec:dynamic_proposer_selection} and the experimental results in \cref{sec:experiment}.

\item {\bf Limitations}
    \item[] Question: Does the paper discuss the limitations of the work performed by the authors?
    \item[] Answer: \answerYes{} %
    \item[] Justification: We discuss limitations in Section~\ref{sec:limitation_and_futurework}, including noisy validation feedback and dependence on the underlying LLM, and analyze API cost and wall-clock overhead in Appendix~\ref{app:cost}.

\item {\bf Theory assumptions and proofs}
    \item[] Question: For each theoretical result, does the paper provide the full set of assumptions and a complete (and correct) proof?
    \item[] Answer: \answerYes{} %
    \item[] Justification: The global convergence theorem is stated in \cref{thm:convergence}, and Appendix~\ref{app:proof} provides the full assumptions, finite-time regret analysis, acceleration characterization, and proof.

    \item {\bf Experimental result reproducibility}
    \item[] Question: Does the paper fully disclose all the information needed to reproduce the main experimental results of the paper to the extent that it affects the main claims and/or conclusions of the paper (regardless of whether the code and data are provided or not)?
    \item[] Answer: \answerYes{} %
    \item[] Justification: The experimental setup in \cref{sec:experiment} specifies the baselines, datasets, metrics, data splits, evaluation budget, LLM backbone, and key hyperparameters. Further reproducibility details are provided in Appendix~\ref{app:search_space} (search space), Appendix~\ref{app:implementation_details} (baseline implementations and compute environment), and Appendix~\ref{app:prompt_tuning} (LLM prompt templates).

\item {\bf Open access to data and code}
    \item[] Question: Does the paper provide open access to the data and code, with sufficient instructions to faithfully reproduce the main experimental results, as described in supplemental material?
    \item[] Answer: \answerYes{} %
    \item[] Justification: We provide the code as supplementary material with instructions for reproducing the main experiments. The code is also available at \url{https://github.com/PKU-DAIR/open-box}. The datasets are publicly available through AMLB/OpenML.

\item {\bf Experimental setting/details}
    \item[] Question: Does the paper specify all the training and test details (e.g., data splits, hyperparameters, how they were chosen, type of optimizer) necessary to understand the results?
    \item[] Answer: \answerYes{} %
    \item[] Justification: The experimental setup is described in \cref{sec:experiment}, including datasets, metrics, data splits, baselines, budget, and key hyperparameters.

\item {\bf Experiment statistical significance}
    \item[] Question: Does the paper report error bars suitably and correctly defined or other appropriate information about the statistical significance of the experiments?
    \item[] Answer: \answerYes{} %
    \item[] Justification: We assess statistical significance for the main benchmark using a Critical Difference diagram with the Nemenyi post-hoc test over 104 AMLB datasets, as shown in \cref{fig:test_cd}. A similar CD analysis for post-hoc ensemble performance is provided in Appendix~\ref{app:ensemble}.

\item {\bf Experiments compute resources}
    \item[] Question: For each experiment, does the paper provide sufficient information on the computer resources (type of compute workers, memory, time of execution) needed to reproduce the experiments?
    \item[] Answer: \answerYes{} %
    \item[] Justification: We report the compute environment in Appendix~\ref{app:implementation_details} and the evaluation budget in \cref{sec:experiment}. Appendix~\ref{app:cost} further analyzes API cost and wall-clock overhead.
    
\item {\bf Code of ethics}
    \item[] Question: Does the research conducted in the paper conform, in every respect, with the NeurIPS Code of Ethics \url{https://neurips.cc/public/EthicsGuidelines}?
    \item[] Answer: \answerYes{} %
    \item[] Justification: The research uses public benchmark datasets and does not involve human subjects, private data collection, or deployment in high-risk settings. To the best of our knowledge, the work conforms to the NeurIPS Code of Ethics.

\item {\bf Broader impacts}
    \item[] Question: Does the paper discuss both potential positive societal impacts and negative societal impacts of the work performed?
    \item[] Answer: \answerYes{} %
    \item[] Justification: We discuss broader impacts in Section~\ref{sec:limitation_and_futurework}.
    
\item {\bf Safeguards}
    \item[] Question: Does the paper describe safeguards that have been put in place for responsible release of data or models that have a high risk for misuse (e.g., pre-trained language models, image generators, or scraped datasets)?
    \item[] Answer: \answerNA{} %
    \item[] Justification: The paper does not release pretrained generative models, scraped datasets, or other assets with high misuse risk.

\item {\bf Licenses for existing assets}
    \item[] Question: Are the creators or original owners of assets (e.g., code, data, models), used in the paper, properly credited and are the license and terms of use explicitly mentioned and properly respected?
    \item[] Answer: \answerYes{} %
    \item[] Justification: We properly cite the existing assets used in this work, including AMLB/OpenML datasets, prior AutoML systems and baselines, and LLM backbones. Public datasets are accessed through AMLB/OpenML under their original licenses and terms.

\item {\bf New assets}
    \item[] Question: Are new assets introduced in the paper well documented and is the documentation provided alongside the assets?
    \item[] Answer: \answerYes{} %
    \item[] Justification: We release the implementation of \sys as supplementary material.

\item {\bf Crowdsourcing and research with human subjects}
    \item[] Question: For crowdsourcing experiments and research with human subjects, does the paper include the full text of instructions given to participants and screenshots, if applicable, as well as details about compensation (if any)? 
    \item[] Answer: \answerNA{} %
    \item[] Justification: This work does not involve crowdsourcing, user studies, human-subject experiments, or human data collection.

\item {\bf Institutional review board (IRB) approvals or equivalent for research with human subjects}
    \item[] Question: Does the paper describe potential risks incurred by study participants, whether such risks were disclosed to the subjects, and whether Institutional Review Board (IRB) approvals (or an equivalent approval/review based on the requirements of your country or institution) were obtained?
    \item[] Answer: \answerNA{} %
    \item[] Justification: This work does not involve human-subject experiments, crowdsourcing, user studies, or human data collection.

\item {\bf Declaration of LLM usage}
    \item[] Question: Does the paper describe the usage of LLMs if it is an important, original, or non-standard component of the core methods in this research? Note that if the LLM is used only for writing, editing, or formatting purposes and does \emph{not} impact the core methodology, scientific rigor, or originality of the research, declaration is not required.
    \item[] Answer: \answerYes{} %
    \item[] Justification: LLM usage is a core component of the proposed method. We describe the LLM proposer in \cref{sec:llm_proposer}, and the LLM backbone used in experiments in \cref{sec:experiment}. Prompt designs for hyperparameter tuning and reflection are provided in Appendix~\ref{app:prompt_tuning} and Appendix~\ref{app:prompt_reflection}.

\end{enumerate}

\end{document}